# Beyond Words: Evaluating Large Language Models in Transportation Planning

Shaowei Ying[1,3], Zhenlong Li[1*], Manzhu Yu[2]

[1] Geoinformation and Big Data Research Lab, Department of Geography, The Pennsylvania State University, University Park, PA, United States

[2] Department of Geography, The Pennsylvania State University, University Park, PA, United States

[3] NCS Group, Singapore

[*] zhenlong@psu.edu

**Abstract.** The resurgence and rapid advancement of Generative Artificial Intelligence (GenAI) in 2023 has catalyzed transformative shifts across numerous industry sectors, including urban transportation and logistics. This study investigates the evaluation of Large Language Models (LLMs), specifically GPT-4 and Phi-3-mini, to enhance transportation planning. The study assesses the performance and spatial comprehension of these models through a transportation-informed evaluation framework that includes general geospatial skills, general transportation domain skills, and real-world transportation problem-solving. Utilizing a mixed-methods approach, the research encompasses an evaluation of the LLMs' general Geographic Information System (GIS) skills, general transportation domain knowledge as well as abilities to support human decision-making in the real-world transportation planning scenarios of congestion pricing. Results indicate that GPT-4 demonstrates superior accuracy and reliability across various GIS and transportation-specific tasks compared to Phi-3-mini, highlighting its potential as a robust tool for transportation planners. Nonetheless, Phi-3-mini exhibits competence in specific analytical scenarios, suggesting its utility in resource-constrained environments. The findings underscore the transformative potential of GenAI technologies in urban transportation planning. Future work could explore the application of newer LLMs and the impact of Retrieval-Augmented Generation (RAG) techniques, on a broader set of real-world transportation planning and operations challenges, to deepen the integration of advanced AI models in transportation management practices.

## 1. Introduction

Geospatial Artificial Intelligence (GeoAI) represents an innovative fusion of geospatial science with advanced artificial intelligence and machine learning technologies to extract valuable insights from spatial data. This interdisciplinary field leverages the power of artificial intelligence (AI) to analyze, interpret, and visualize geospatial information, enabling more efficient and accurate decision-making in areas ranging from urban planning and environmental management to transportation and logistics. In recent years, one key driver for the active development of GeoAI has been the resurgence of Generative AI (GenAI), a branch of AI technology focused on creating new, original data that mimics the characteristics of its training data. These models are trained on large datasets to recognize and replicate complex patterns, allowing for the generation of content that is both new and remarkably similar to the original data. Through this process, GenAI systems gain the capability to produce nuanced outputs that can emulate human creativity and intelligence, facilitating groundbreaking applications in numerous fields. Since the release of ChatGPT in November 2022, there has been a surge of research interest in studying the impact of GenAI on geospatial tasks. While many of the recent publications address general geospatial tasks, such as accurately extracting location descriptions from various types of unstructured texts (Mooney, Cui, & Juhász, 2023; Hu et al., 2023; Haris, Cohn & Stell, 2024), performing geospatial analyses (Frez & Baloian, 2023) or automating GIS workflows (Li & Ning, 2023), there is also a growing body of research work focused on domain-specific applications such as public health (Jo et al., 2023; Lim & Schmälzle, 2023) and disaster management (Otal & Canbaz, 2024; Akinboyewa et al., 2024).

Among the different GenAI foundation models, text-based Large Language Models (LLMs) arguably will play a pivotal role in the transformation of the transportation sector. The ability of LLMs to process, understand, and generate human-like text makes them especially suitable for managing the complex textual datasets commonly found in transportation logistics, such as regulatory documents, customer communications, operational directives, and global trade documentation. Several notable transportation-focused LLM prototypes have been developed, including TrafficGPT (Zhang et al., 2023), an LLM-powered system that combines ChatGPT and traffic foundation models. TrafficGPT uses its reasoning ability to deconstruct the user-input task and calls the respective Traffic Foundation Models to provide answers. Another example is TrafficSafetyGPT (Zheng et al., 2023), a LLaMA-2 model fine-tuned with transportation safety knowledge to provide responses that are aligned with ratified traffic safety standards and guidelines. Additionally, Open-TI (Da et al., 2023) aims to "harness external traffic analysis packages based on existing conversation".

As more applications of LLM are being explored, there has also been a corresponding growth in research to evaluate LLMs. Most of the evaluation research focuses on "horizontal" tasks of natural language processing (understanding, reasoning, generation, question answering, multi-lingual), factuality, robustness, ethics, biases, and trustworthiness, with fewer works investigating the feasibility of using LLMs in GIS contexts and even fewer on evaluating LLMs in specific domains such as transportation planning and management (Chang et al., 2023). In urban transportation management, addressing challenges such as congestion pricing, traffic safety, and route optimization requires sophisticated analysis and decision-making skills typically found in human experts. With the increasing complexity of urban transportation systems and the growing need for efficient, data-driven solutions, it becomes imperative to explore advanced technological tools that can assist or even enhance human capabilities in this domain. Traditional AI has been used in five major areas of transportation management, but the potential of GenAI models in transportation planning and management remains underexplored (Table 1). Understanding how effectively LLMs can comprehend and operate within geospatial and transportation-specific contexts could lead to significant advancements in how urban transportation systems are analysed and managed. If successful, LLMs could provide substantial support in decision-making processes, enhance the efficiency of transportation planning, and offer innovative solutions to persistent challenges.

**Table 1** Summary of how GenAI could potentially promote exciting new use cases in five major areas in the transportation domain.

| Functional aspect | | What "traditional" AI can do | What GenAI has the potential to do incrementally |
|---|---|---|---|
| Network Analysis and Optimization | Determining the most efficient routes for vehicles, taking into account current traffic conditions, road closures, and other factors to minimize travel delays and improve overall transportation efficiency. | Analyze real-time data for route optimization, predict traffic patterns and congestion for dynamic rerouting. ( Abduljabbar et al., 2019) | Generate optimized routing strategies and simulate complex networks under varying conditions. (Zhang et al., 2024) |
| Asset Management | Systematic monitoring and maintenance of transportation infrastructure, such as roads, bridges, and tunnels, to ensure they remain in good condition and function effectively over time | Predict maintenance needs, automate condition monitoring using imagery and sensor data. (Venugopal, 2022; Wu et al., 2022; Gunderia et al., 2024) | Create predictive models for optimal maintenance schedules, simulate effects of maintenance strategies. (Ćelić et al., 2024) |
| Safety Analysis | Identifying and predicting areas where accidents are likely to occur, enabling targeted interventions to enhance road safety and improve emergency response times. | Predict accident hotspots from historical data, automate incident detection for faster emergency response. (Bhattacharya et al., 2022; Olugbade et al., 2022) | Predict and simulate accident scenarios, create training simulations for improved preparedness. (Cai et al., 2020; Lin et al., 2023) |
| Environmental Impact Assessment | Predicting and assessing the environmental impacts of transportation projects, including emissions, noise, and effects on wildlife, to support the development of sustainable and eco-friendly transportation solutions. | Model environmental impacts of projects, support sustainable route planning. (Saleh et al., 2024) | Generate predictive models simulating environmental consequences of projects. (Mohanty et al., 2024) |
| Public Engagement and Information Dissemination | Providing travelers with essential information about travel times, transit status, weather conditions, and other factors to help them make informed decisions about their travel plans. | Deliver personalized travel information, enhance user interaction with chatbots and virtual assistants. (Dash & Sharma, 2022) | Create personalized, interactive content and responses for enhanced public engagement (Zeng, 2023). |

This study aims to evaluate the performance and spatial comprehension of LLMs by employing a transportation-informed evaluation framework designed to probe the depth of LLMs' understanding of geospatial concepts as they apply to typical urban transportation management scenarios. Specifically, this study investigates the following three research questions: 1) Do out-of-the-box LLMs exhibit general geospatial skills and basic knowledge of transportation planning? 2) Do out-of-the-box LLMs have sufficient reasoning capabilities to design and execute an analysis workflow analogous to that undertaken by a human expert transportation planner in addressing real-world transportation planning challenges such as congestion pricing? 3) How does a highly advanced, large-scale LLM compare to a compact, mobile-optimized model in terms of understanding of geospatial and transportation-specific knowledge when applied to real-world transportation planning scenarios? By investigating these questions, the study seeks to uncover the potential and limitations of LLMs in the domain of transportation planning, providing insights that could inform future research and development efforts.

## 2.  Methodology

This research starts with the selection of the LLMs.  These "Test LLMs" are then deployed on Azure cloud and subjected to three levels of evaluation, assessing the Test LLMs' (1) general geospatial skills, (2) general transportation domain knowledge, and (3) real-world transportation problem-solving abilities (Figure 1).

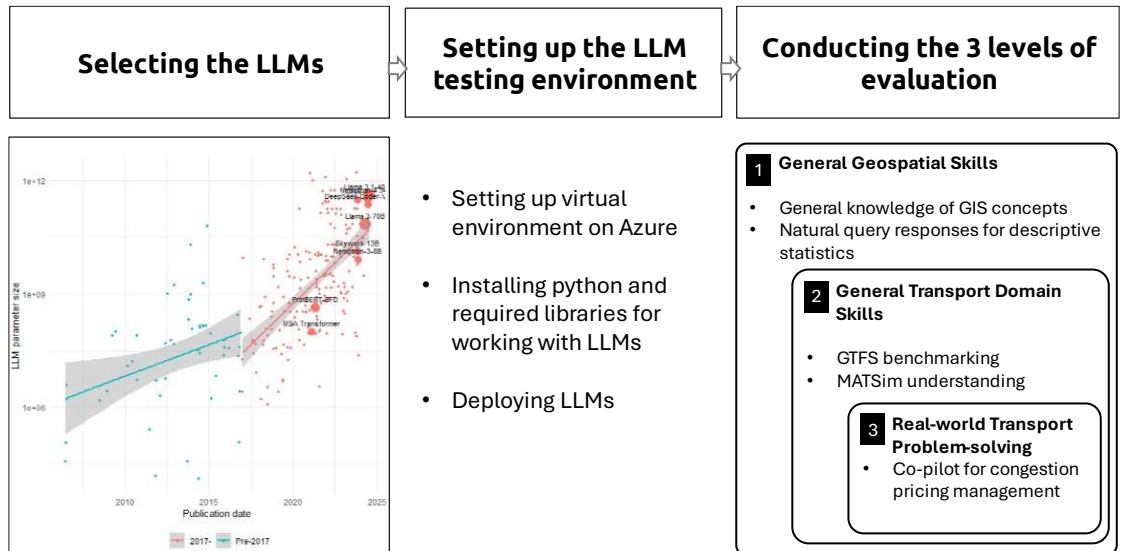

**Figure 1.**  The three steps of the research

### 2.1 Selection of the Test LLMs

A comprehensive database of notable and highly cited models is maintained by Epoch AI (Epoch AI, n.d.).  From this database, two distinct models representing contrasting development paradigms were chosen: Azure OpenAI GPT-4 and Microsoft Phi-3-mini. The selection of these two models as the Test LLMs reflects a deliberate choice to explore and compare the performance of a highly advanced, large-scale LLM with a more compact, mobile-optimized LLM.

GPT-4 stands out as a general-purpose, closed LLM with an impressive 170 trillion parameters, making it one of the most sophisticated language models available. GPT-4 is designed to generate human-like text based on the input it receives, positioning it as a state-of-the-art model for a wide range of tasks, including those in the domain of transportation planning. While GPT-4 is directly provided by OpenAI, it has also been integrated into Microsoft's cloud platform Azure as Azure OpenAI GPT-4, allowing users to leverage GPT-4's advanced language processing capabilities directly from the Azure cloud.

In contrast, Microsoft Phi-3-mini is a lightweight, open-source LLM with 3.8 billion parameters, specifically designed to operate on mobile platforms. Developed by Anthropic in collaboration with Microsoft, Phi-3-mini aims to provide LLM capabilities in a more accessible and versatile format. It is also accessible through APIs, with potential integrations within Microsoft's ecosystem, making it suitable for applications that require more mobile and adaptable solutions. Despite its smaller parameter size, Phi-3-mini offers significant utility, particularly in scenarios where resource constraints or the need for on-device processing are critical considerations. Achiam et al. (2023) and Abdin et al. (2024) provided a comparison of these two models using common academic benchmarks.

## 2.2 Setting up the LLM testing environment

A Python-based LLM evaluation platform named "Trafficgpt" was developed on Azure Cloud. This platform was designed to structure the interaction between users and LLMs. Trafficgpt features the following evaluation frameworks: (1) a chat-only interface, facilitating simple text-based interactions with the LLM; (2) a Prompt File Template, which structures text for AI interpretation by defining specific tasks ranging from queries to feedback statements; and (3) a Tabular Framework, which enables users to query loaded tabular data. This framework integrates with the REPL (Read-Eval-Print Loop) tool, allowing for the execution of Python code lines for in-depth data analysis.

## 2.3 Three levels of evaluation

The Test LLMs were subjected to three levels of testing, each level designed to assess the models' capabilities in different aspects of geospatial and transportation domain skills (Figure 2).

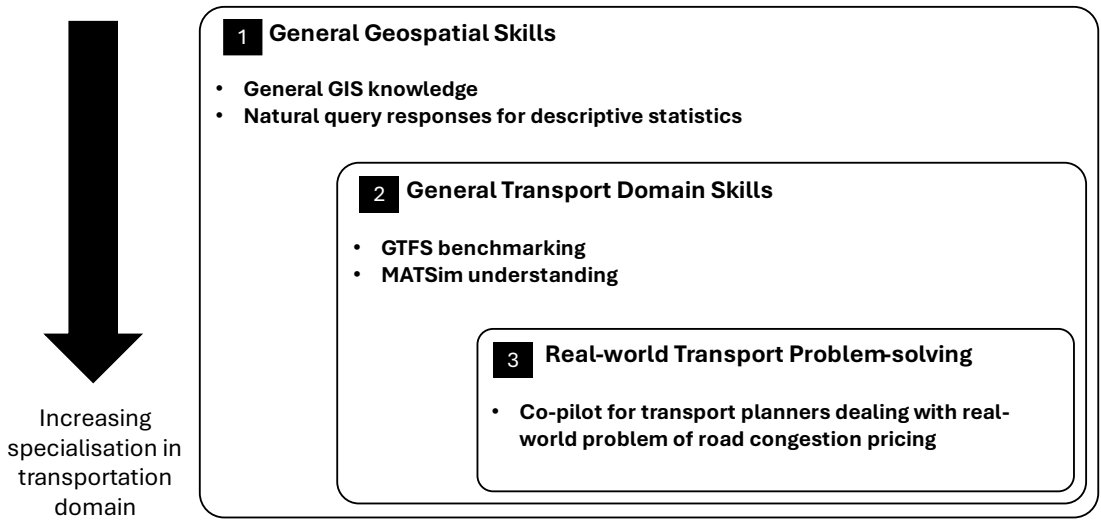

**Figure 2.** The evaluation framework comprises three levels of assessments representing increasing specialization in the transportation domain.

### 2.3.1 General Geospatial Skills

The first level of testing focuses on evaluating the general geospatial skills of the Test LLMs "out-of-the-box". This evaluation includes two types of examinations: GIS concepts and the understanding of transportation-related tabular data.

### GIS concepts

The GIS concept examination was designed by Mooney (2023), which was based on instructor resources from a popular GIS textbook called GIS Fundamentals *(Bolstad & Manson, 2022)*. A balanced set of 60 questions was selected manually to simulate a real exam that covers fundamental concepts of mapping and GIS, data sources and tabular data, spatial analysis, spatial statistics and interpolation, and applied GIS workflow. The Test LLMs were prompted to pick a response from a set of multiple-choice or True/ False options without requiring explanations.

### Understanding of transportation-related tabular data

The second type of examination is to assess the Test LLMs' ability to handle and interpret transportation-related tabular data, which is a common format for representing transportation information. The tasks include data interpretation of schedules, performance metrics, and operational statistics; analysis of patterns, trends, and anomalies for better resource allocation and planning; and generating accurate reports and summaries to enhance transparency and communication with stakeholders.

The Test LLMs are given a typical traffic volume and travel time dataset from MATSim, a transportation software simulation tool that contains 40,000 road links of Singapore's road transportation network, and the simulated traffic volume and travel speed along each of the road links (Figure 3). A list of manually curated 15 questions is used to test the LLM's ability to understand the intent of the query, and to search and retrieve the right values from the dataset.

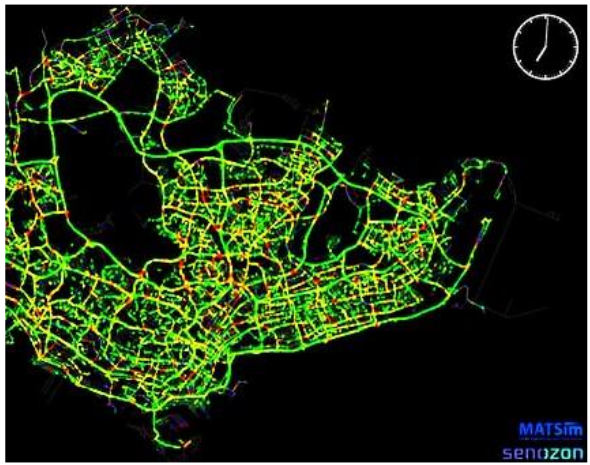

**Tabular Data fields**
- LINK : Unique link identifier for the road link
- FROM : ID of start location of link
- TO : ID of end location of link
- LENGTH : Distance of the link in metres
- FREESPEED : Speed limit of vehicles in metres / second
- CAPACITY : Capacity of road link in terms of number of vehicles in an hour
- HRS0-1avg to HRS23-24avg: Number of vehicles using the road link in each hour of day
- HRS0-24avg: Total number of vehicles using the road link for the day
- TRAVELTIME0 -1avg : to TRAVELTIME23 -24avg : Travel time of vehicles transversing the road link (in seconds).

**Figure 3.** MATSim dataset for assessing the Test LLMs' ability to handle tabular data information, contains simulated traffic information about the road links of Singapore

The LLM responses are evaluated by the two human transportation practitioners and scored according to their ability to define the right analytical steps and to generate accurate Python code. The rubric used for scoring LLM responses in this benchmarking involves three key dimensions: (1) understanding of context, (2) accuracy and completeness, and (3) clarity and presentation.

### 2.3.2 General Transport Domain Skills

The second level of testing delves into the LLMs' understanding of transportation domain-specific datasets and concepts. This involves testing the LLMs' knowledge about the General Transit Feed Specification (GTFS), a standard for publishing public transit data, and MATSim, an open-source transportation simulation software. The LLMs' responses are assessed to understand the depth and breadth of their domain knowledge.

### GTFS understanding

GTFS is an open data format that allows public transportation agencies to publish their transit data in a format that can be consumed by a variety of software applications. It standardizes information about routes, schedules, stop locations, and other transit features. While other data formats exist for land transportation, GTFS is preferred for evaluating and benchmarking Test LLMs due to its widespread global adoption (Antrim & Barbeau, 2013) and its extensive data coverage, including both static and real-time information (GTFS Realtime Overview, n.d.). This allows for a comprehensive assessment of LLMs on both stable data structures and dynamic real-time streams, reflecting real-world applications. Additionally, GTFS data's relevance extends beyond transit applications, serving as a critical resource in transportation research, urban planning, and mobility studies (Ho, 2021).

The benchmarking approach in the context of GTFS involves a structured set of assessments designed to probe the models' understanding and handling of GTFS-related tasks. This evaluation is based on the set of 195 multiple-choice questions (MCQs) designed by Devunuri, Qiam, & Lehe (2024), to test the LLMs' grasp of specific terminology, basic knowledge of GTFS usage, and the reasoning behind its data organization. It also assesses their ability to recognize GTFS file structures, correctly map data attributes to files, handle various data types within the GTFS framework, and accurately categorize and interpret GTFS data.

Responses are analyzed to determine how well the Test LLMs can select the correct answer based on their pre-existing knowledge of GTFS. The testing approach utilizes zero-shot (ZS) prompting techniques, where the Test LLMs are tested without prior exposure to similar tasks, relying solely on their pre-trained knowledge to respond to the questions.

### MATSim understanding

MATSim (Multi-Agent Transport Simulation) is an open-source framework primarily used to simulate the mobility of individuals as separate agents with distinct travel plans within a road network. As one of the more popular transportation simulation software, MATSim allows for detailed analysis of traffic patterns and policy impacts on individual behavior. MATSim has been the preferred tool used by traffic researchers in Singapore, which is the study area for this research. An example of MATSim research done in Singapore is Oh et al. (2021). LLMs have the potential to significantly enhance the accessibility and usability of MATSim and similar agent-based modeling software, which often require specialized training to use. Via natural language interfaces, LLMs can assist in generating initial models based on user descriptions, reducing the learning curve and time required to start modelling. LLMs can also provide real-time code assistance and debugging help.

The benchmarking approach for evaluating the Test LLMs involves a structured set of assessments designed to probe the models' understanding and handling of MATSim-related tasks. A set of 120 multiple-choice questions (MCQs) was developed based on the MATSim user guide (Horni, Nagel, & Axhausen, 2016). Each question has four options, with only one correct answer. The question list was initially generated by ChatGPT and then manually curated to ensure relevance and accuracy. The questions are categorized into six main sections of the guide, with 20 questions each, covering key areas of MATSim: basic usage and functionalities, visualization tools, input data generation, supplementary features, output analysis, and the ability to extend and customize MATSim for specific simulations. Responses are analyzed to determine how well the Test LLMs can select the correct answer, without prior exposure to similar tasks.

### 2.3.3 Real-world Transport Problem-solving

The third and most specialized level of evaluation challenges the LLMs with real-world transportation problem-solving tasks. The Test LLMs are tasked with guiding a transportation planner through the process of studying the effects of raising or lowering tolls on traffic congestion, a typical real-world transportation challenge. Congestion pricing is used in many countries as a strategy to improve transportation system performance. Singapore is one of the pioneers of congestion pricing, introducing a manual Area Licensing Scheme in 1975 and moving to electronic road pricing from 1998 (Phang & Toh, 1997).

### *Real world traffic scenario*

The evaluation is contextualized around a significant tolled highway in Singapore called the Kallang–Paya Lebar Expressway (KPE), which is among 11 expressways and other major arterial roads that crisscross the Singapore urban landscape (Figure 4). The KPE has the longest subterranean road tunnel in Southeast Asia and was the world's sixth-longest underground road project at the time of its construction, built at a cost of approximately $1.74 billion (National Library Board of Singapore, n.d.). It provides a seamless route between the southeastern and northeastern parts of Singapore (Figure 5), connecting major expressways such as the East Coast Parkway (ECP) and the Tampines Expressway (TPE). The real-world question a transportation planner seeks to answer is the extent the toll price changes will have on travel volume and speed along the KPE, as travel patterns are re-distributed across the Singapore road network.

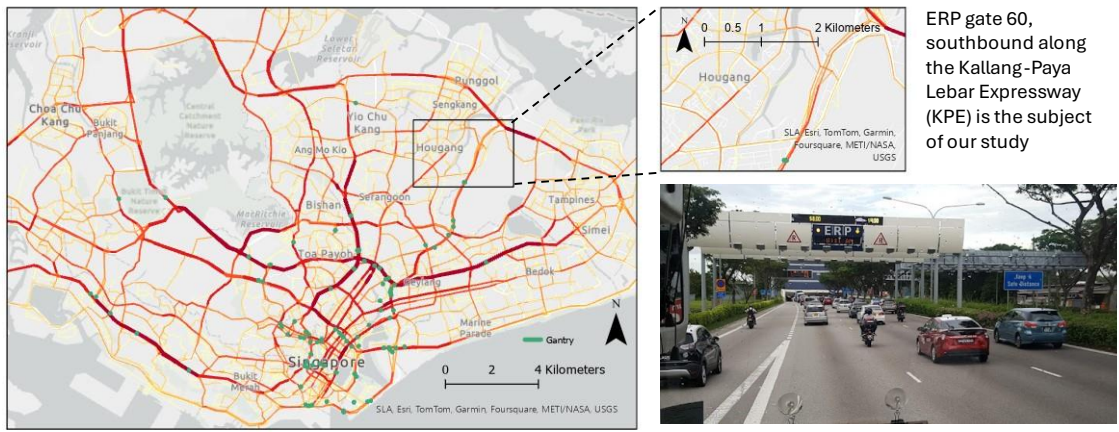

**Figure 4.** Electronic road pricing (ERP) gantries are placed along major expressways and arterial roadways in Singapore. There are 106 gantries in Singapore. The pricing at two specific ERP gantries is being reviewed in an effort to improve the traffic flow across the Marina Coastal Expressway (MCE).

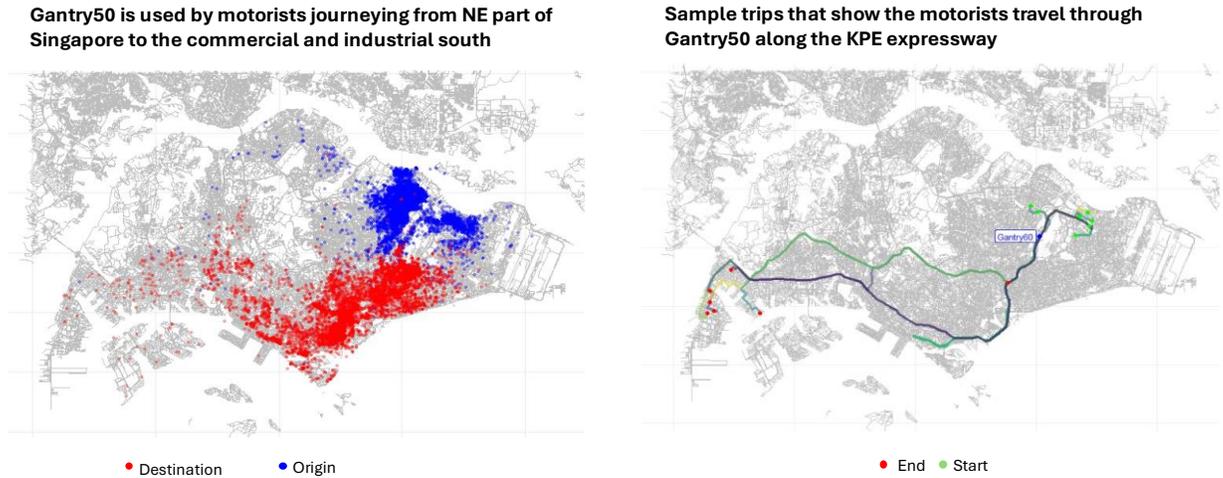

**Gantry50 is used by motorists journeying from NE part of Singapore to the commercial and industrial south**

**Sample trips that show the motorists travel through Gantry50 along the KPE expressway**

• Destination   • Origin

• End   • Start

**Figure 5.** Change in ERP rates along the KPE can affect how commuters will travel across the different regions of Singapore. For planning purposes, there are 55 planning areas, further broken down into 332 subzones. Specifically, Gantry 50 affects the daily movement of commuters from the northeast of Singapore towards the south commercial and industrial districts.

### *Evaluation approach*

Although a range of analytical methods can be applied to set road congestion pricing, two specific methods are selected for evaluating the Test LLMs, namely the Price Elasticity of Demand approach and the Agent-Based Modeling approach. These methods were chosen based on their balance of complexity and practical applicability (Figure 6). Price Elasticity of Demand measures how the quantity demanded of a good or service changes in response to a change in price. In the context of congestion pricing, it helps estimate how changes in toll rates might affect traffic volume. Agent-Based Modeling involves simulating the actions and interactions of autonomous agents to assess their effects on the system as a whole. In transportation planning, it models individual drivers' behavior and their responses to changes in toll rates.

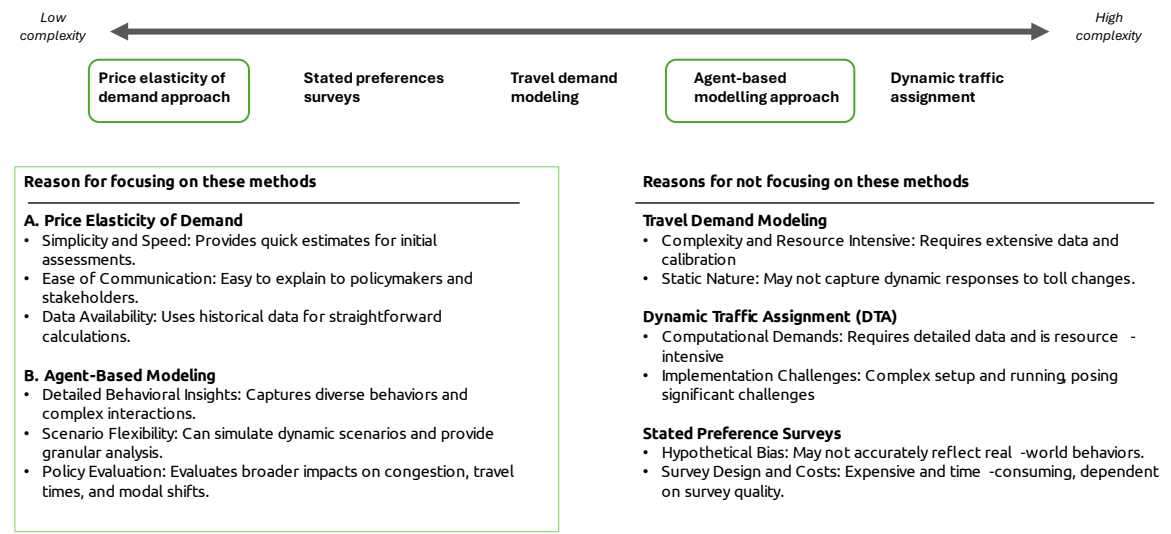

**Figure 6:** Spectrum of analytical methods of different levels of complexity that can be used to set congestion pricing.

The two corresponding analysis pathways are shown in Figure 7. The prompts for the two evaluation pathways used to check on the LLMs' ability to generate the workflow and the corresponding executable R or python scripts. The first pathway, the Price Elasticity of Demand approach, involves several steps. It begins with sense-checking datasets to ensure data accuracy and relevance. Following this, the elasticity coefficients are computed to understand how sensitive the demand for road usage is to changes in price. These coefficients are then validated against known benchmarks or through literature reviews to ensure their accuracy. Once validated, the next step is to simulate pricing changes based on these coefficients to observe potential impacts on traffic patterns. Finally, policy recommendations are formulated based on the simulation outcomes, providing actionable insights for managing congestion pricing effectively.

The second pathway, the Agent-Based Modeling approach, is slightly more complex and involves using a MATSim simulation output dataset. This approach also starts with sense-checking the dataset for common issues and verifying traffic patterns. The next step is to filter relevant origin-destination (OD) trips that pass through the specific gantry being studied. Travel costs for these trips are then computed, considering factors such as travel time, distance, and toll costs. This computation helps identify a cost threshold beyond which motorists might avoid using the gantry. After calculating these costs, simulations of pricing changes are conducted to observe how traffic might redistribute in response to different pricing strategies. The results of these simulations help make informed policy recommendations for optimizing congestion pricing.

Both these analysis pathways are what a human transportation planning expert would be expected to take. The responses of the LLM will be considered in terms of its usefulness in assisting the human expert to complete his or her job, namely:

- Understanding of User Intent: This involves evaluating the LLM's context awareness and its natural language understanding capabilities. The LLM should accurately comprehend and respond to user queries, maintaining context across multiple interactions.
- Interpretation of Data: The LLM's ability to perform data processing and analytical tasks should be as close to that of a human expert. This includes accurately interpreting data, identifying trends, and performing relevant analyses.
- Quality of Recommendations: The suggestions provided by the LLM should be relevant, actionable, and innovative. The LLM's recommendations need to be practical, implementable, and offer fresh perspectives or solutions that may not have been considered by human analysts.
- Automation of Report Generation: The LLM should be capable of generating comprehensive and accurate reports. These reports should be well-structured, clear, and easy to read, covering all necessary details and aspects of the analysis.

*Dataset*

The primary dataset used is a simulation dataset prepared by Anda C., as part of his PhD work detailed in references Anda, Medina & Fourie (2018) and Anda, Medina, & Axhausen (2021). This dataset is a realistic representation of individual motoring trips across the entire Singapore road network within a single day. It serves multiple purposes in transportation and land-use policy evaluation, travel demand forecasting, and transportation infrastructure testing.

The dataset was synthetically generated using the MATSim simulation software, with input data being DataSpark's proprietary mobile cellular network datasets (Dang et al., 2017; Li & Luo, 2018). The DataSpark API is regularly used by researchers and land transportation or urban planning agencies in Singapore to augment official statistics collection efforts such as household travel surveys. For instance, Marakkalage et al. (2018) used the DataSpark data to understand the daily lifestyle of the older population in Singapore, while Chiam, Dang, & Li (2018) compared the daily activity patterns of two populous metropolitans, namely Singapore and Sydney.

       Key data fields in the dataset include agent ID (a unique identifier for each vehicle), departure time, trip duration, origin and destination, and the route taken (represented as a sequence of road links).  The validity of this dataset has been established through various methods. As described in Anda, Medina, & Axhausen (2021), the simulation results were compared against actual traffic count data, showing accurate patterns for private car usage. Additionally, public transportation validation was conducted against smart card transaction data, revealing some discrepancies, especially an overestimation of morning peak demand. A supervised learning framework was employed to infer travel modes, enhancing the dataset's reliability. Temporal disaggregation methods ensured accurate representation of departure times, and MATSim's evolutionary algorithm dynamically adjusted and optimized routes, further validating the dataset's utility in reflecting realistic travel behaviors.

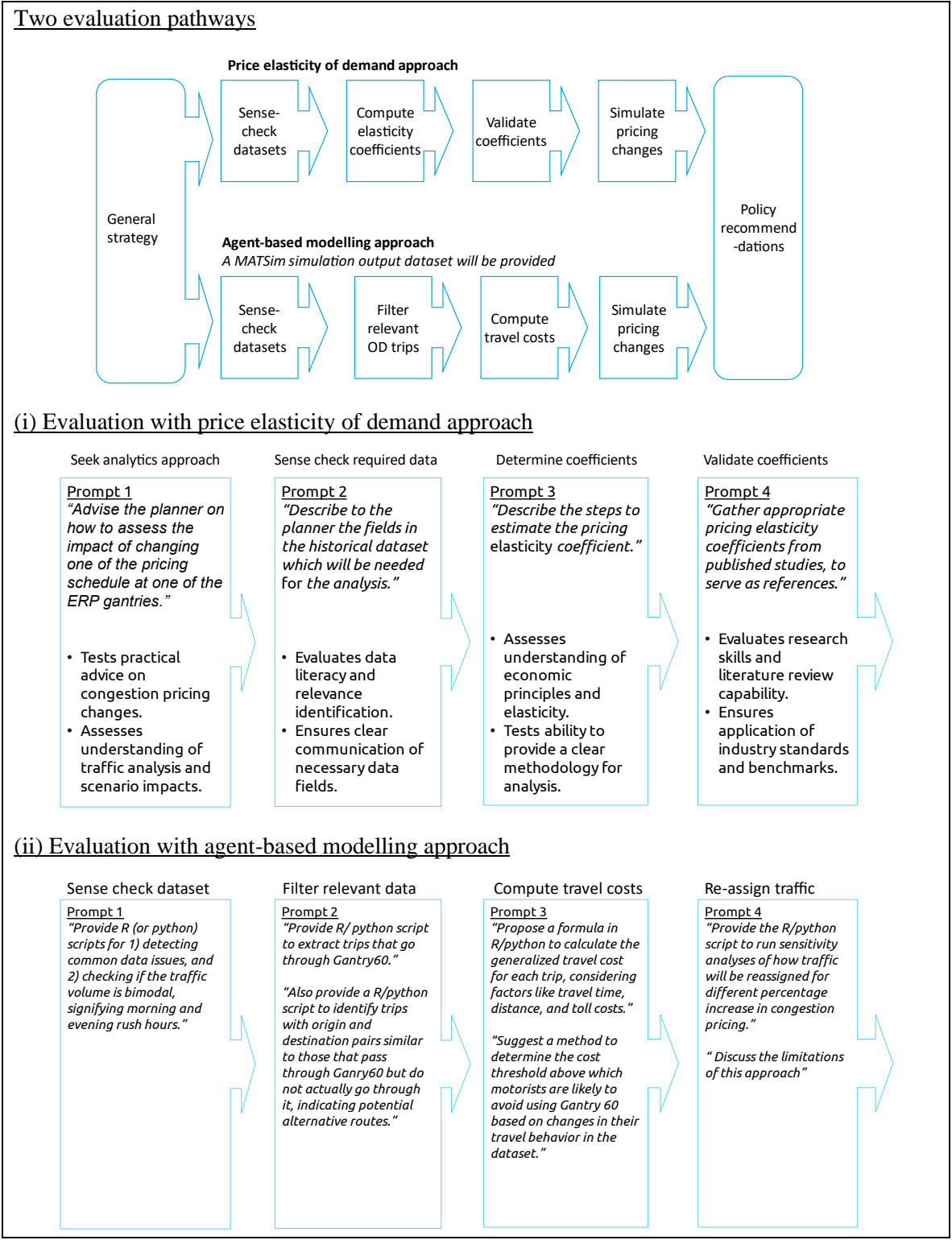

**Figure 7.** The prompts for the two evaluation pathways used to check on the LLMs' ability to generate the workflow and the corresponding executable R or python scripts.

## 3. Results and Discussion

### 3.1 Level 1: General Geospatial Skills of the Test LLMs

#### 3.1.1 Understanding of GIS concepts

Figure 8 summarizes the performance of the Test LLMs on the GIS exams. The responses of the two LLMs to each of the 60 exam questions can be found in Appendix A. Overall, GPT-4 outperformed Phi-3-mini, achieving an accuracy of 86% by correctly answering 52 out of 60 questions, while Phi-3-mini achieved an accuracy of 70% with 42 correct answers. These results are consistent with previous findings by Mooney (2023).

A z-test for two independent proportions showed that the outperformance of GPT-4 over Phi-3 was statistically significant, yielding a two-tailed p-value of 0.034. Across all GIS topics, GPT-4 demonstrated superior performance, particularly with more complex questions such as applied GIS workflows. Nevertheless, Phi-3-mini still performed competently in many GIS-related tasks and may be suitable for specific applications.

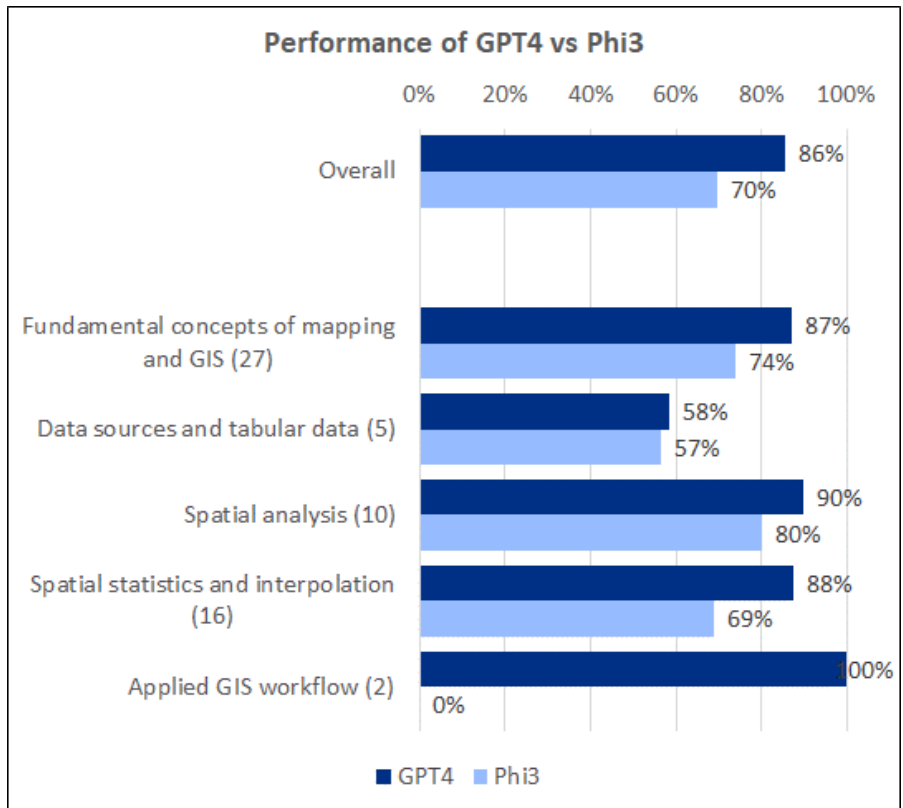

**Figure 8.** Results of GIS exams comprising 60 multi-choice and true-false questions for Phi-3 and GPT4. The actual responses of the Test LLMs to each of the 60 questions can be found in Appendix A.

#### 3.1.2 Descriptive Statistics on Tabular Data

Although GPT-4 and Phi-3-mini appeared to show similar performance in processing tabular data on the GIS examinations (Figure 8), these results were based on a small set of 5 questions. A more detailed evaluation using a MATSim-generated dataset revealed significant differences in performance between the Test LLMs. Overall, GPT-4 showed a substantially higher capability, with an accuracy of 89%, compared to Phi-3-mini's 53% (Figure 9). This difference is statistically significant, indicating that GPT-4 is more adept at handling tabular data. The responses of the two LLMs to each of the 150 questions can be found in Appendix B.

Table 2 provides an example of how Phi-3-mini incorrectly used data fields to compute the implied average travel speed for each link. It calculated the travel speed on a road link by dividing the traffic volume (the number of vehicles traversing the link) by the length of the road link. In contrast, GPT-4 correctly computed the travel speed by dividing the road link length by the hourly travel time and provided better explanations.

**Table 2.** Responses to the prompt question "What is the implied average speed of travel for each link?" by the Test LLMs, highlighted the superior performance of GPT-4 over Phi-3-mini.

| GPT-4 | Phi-3-mini |
|---|---|
| To find the implied average speed of travel for each link, we will perform the following analytical steps:<br><br>1. Calculate the average travel time for each link by averaging the hourly travel times (TRAVELTIME0-1avg, TRAVELTIME1-2avg, ..., TRAVELTIME23-24avg).<br>2. Divide the length of the link (LENGTH) by the average travel time to get the average speed in meters per second.<br>3. Convert the average speed to a more common unit, such as kilometers per hour (km/h), by multiplying the result by 3.6.<br><br>Here's the Python code for these steps:<br><br>```python<br>import pandas as pd<br><br># Assuming the dataset is in a CSV file called 'traffic_data.csv'<br>traffic_data = pd.read_csv('traffic_data.csv')<br><br># Step 1: Calculate the average travel time for each link<br>travel_time_columns = [f'TRAVELTIME{i}-{i+1}avg' for i in range(24)]<br>traffic_data['avg_travel_time'] = traffic_data[travel_time_columns].mean(axis=1)<br><br># Step 2: Calculate the average speed in meters per second<br>traffic_data['avg_speed_mps'] = traffic_data['LENGTH'] / traffic_data['avg_travel_time']<br><br># Step 3: Convert the average speed to kilometers per hour<br>traffic_data['avg_speed_kph'] = traffic_data['avg_speed_mps'] * 3.6<br><br># The result is stored in the 'avg_speed_kph' column of the traffic_data DataFrame<br>``` | ```python<br># First, we create a DataFrame to hold our data:<br>import pandas as pd<br><br>data = {<br>    "LINK": ["Link1", "Link2", "Link3"],  # Assuming some example links; in real scenario replace these with actual link identifiers from the dataset<br>    "ORIG_ID": [0, 1, 2],<br>    "FROM": [10, 15, 20],<br>    "TO": [20, 30, 40],<br>    # ... (other fields are assumed to be populated similarly)<br>}<br>df = pd.DataFrame(data)<br><br># Then we calculate the average speed for each link based on traffic volume and travel time:<br>def avg_speed(volume, travel_time):<br><br>"""Calculates average speed given volume and travel time."""<br>if travel_time == 0:<br><br>return float('inf')<br><br># Assuming all units are consistent (meters for distance and seconds for time)<br>average_distance = (travel_time * 1000) / 3600  # convert from km/hr to m/s<br><br>speed = volume / average_distance<br>    return speed<br><br># Assuming we have a function 'get_traffic_data' that extracts the necessary data from our dataset:<br>def get_traffic_data(df):<br>    for index, row in df.iterrows():<br>        travel_time = #... some logic to calculate/fetch average hourly travel time based on volume and speed limits etc |
| This code first reads the traffic dataset from a CSV file and calculates the average travel time for each link. Then, it calculates the average speed in meters per second by dividing the link length by the average travel time. Finally, the average speed is converted to kilometers per hour by multiplying the result by 3.6. The average speed for each link can be found in the 'avg_speed_kph' | `avg_speed_value = avg_speed(row['CAPACITY'], travel_time)`<br><br>`# Append the new data back to DataFrame` |

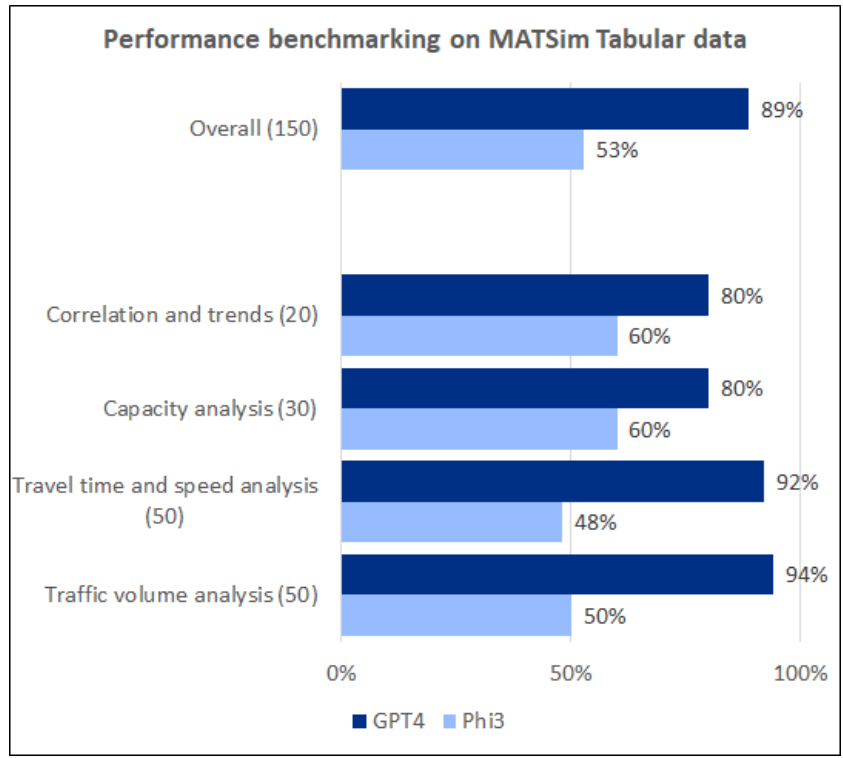

**Figure 9.** GPT-4 and Phi-3-mini relative performance in interpreting and processing tabular data. Appendix B contains the detailed results.

### 3.2 Level 2: General Transport Domain Skills of the Test LLMs

#### 3.2.1 GTFS Understanding

Figure 10 shows the results of the benchmarking analysis between GPT-4 and Phi-3-mini using GTFS (General Transit Feed Specification) across six key categories: Term Definition, File Structure, Data Structure, Common Reasoning, Categorical Mapping, and Attribute Mapping. The responses of the two LLMs to each of the 195 questions can be found in Appendix C. Overall, GPT-4 outperformed Phi-3-mini, achieving an accuracy of 66% compared to Phi-3-mini's 43%. This is consistent in most categories, with particularly high performance in File Structure (88%) and Attribute Mapping (83%). In contrast, Phi-3-mini's performance was considerably lower across all categories, with its best result in Data Structure at 48%. The smallest performance gap between the two models was observed in the Categorical Mapping category, where GPT-4 scored 42% and Phi-3-mini scored 40%, and the difference is not statistically significant using two-proportion z-tests (p = 0.499). While GPT-4 performs significantly better than Phi-3-mini, the overall accuracy of 66% indicates limitations of the state-of-the-art LLM in understanding more specialized domain knowledge in transportation.

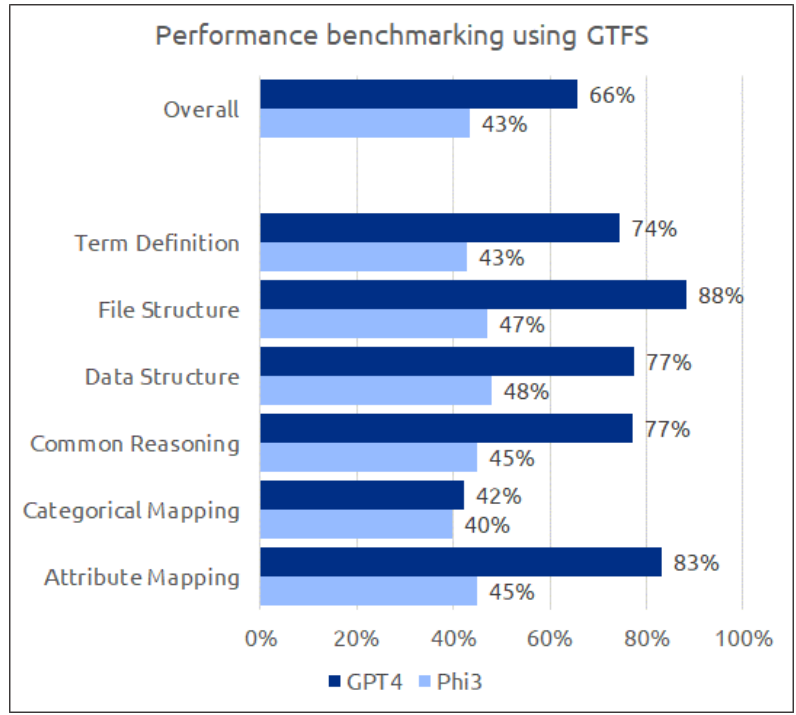

**Figure 10.** GPT-4 vs Phi-3 relative abilities to interpret GTFS dataset. The responses of the two LLMs to each of the 195 questions can be found in Appendix C.

### 3.2.2 MATSim Understanding

Figure 11 shows the performance benchmarking analysis between GPT-4 and Phi-3-mini using MATSim across six key categories with each category including 20 questions: Extending MATSim, Analysis, Additional Core Features, Generation of MATSim Input, Visualizers, and Introducing MATSim. The responses of the two LLMs to each of the 120 questions can be found in Appendix D. Overall, GPT-4 outperformed Phi-3-mini with an accuracy of 81% compared to Phi-3-mini's 72%. In several categories, GPT-4 demonstrated notable advantages. For instance, in the "Additional Core Features" category, GPT-4 achieved a perfect score of 100%, surpassing Phi-3's 95%. Both models performed equally well in Extending MatSIM (90%) and Generation of MATSim Input (75%). Both models show the worst performance in the Analysis and Visualizers with GPT-4 archiving 65% in both categories and 55% and 45% respectively for Phi-3-mini.

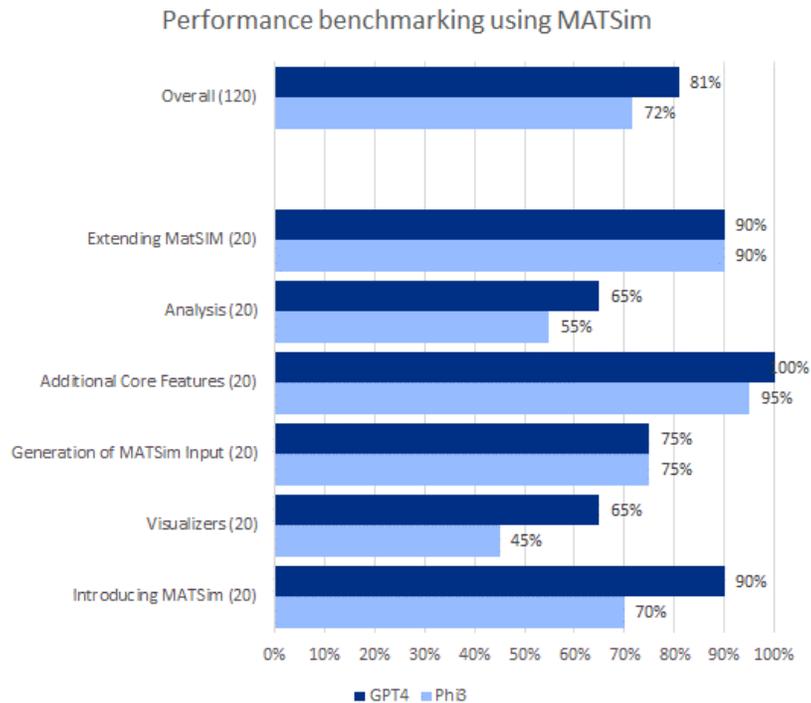

**Figure 11.** Results of exams comprising 120 multi-choice and true-false questions related to MATSim for GPT4 and Phi-3-mini. The response of the two LLMs to each of these questions can be found in Appendix D.

### 3.3 Level 3: Real-world Transportation Problem-Solving by the Test LLMs

In the context of real-world transportation problem-solving, the evaluation of the Test LLMs as potential co-pilots to the human transportation analysts plays a crucial role in determining their practical utility. The following sub-sections present the qualitative assessment by two human transportation experts on how well the Test LLMs can serve as a co-pilot to complete a congestion pricing analysis task.

### 3.3.1 Price Elasticity of Demand Approach

Table shows how the two LLMs were rated by the human analyst. GPT-4's responses were consistently more detailed, structured, and practical, making it a more reliable choice for transportation planning tasks. Its ability to understand complex user queries, accurately interpret data, provide high-quality recommendations, and automate comprehensive reports highlights its robustness and suitability for real-world applications. Phi-3-mini, while generally competent and occasionally providing more detailed statistical techniques, tended to fall short in overall clarity and comprehensiveness. Its responses, although good, were often less detailed and structured compared to GPT-4. Nonetheless, Phi-3-mini's methodological rigor in certain tasks, such as estimating pricing elasticity coefficients, showcases its potential value in specific analytical scenarios. For transportation planners seeking a reliable and effective LLM, GPT-4 is the recommended choice due to its superior performance across all evaluated criteria. Phi-3-mini, while a solid performer, may serve as a complementary tool.

### 3.3.2 Agent-based Modeling Approach

In the agent-based modelling approach tests, the test LLMs were required to provide codes in response to instructions that were more directive. Here, GPT-4 outperformed Phi-3-mini on all questions, as shown in Table 4. GPT-4's superior performance stems from its model size and training data. The same reasons why GPT-4 outperformed GPT-3.5 would likely apply here (Koubaa, 2023). Compared to Phi-3-mini, GPT-4 is a larger model with access to a more extensive and diverse dataset during training, allowing it to learn more nuanced patterns and provide more accurate and comprehensive responses. GPT-4 may also have been trained on a wider variety of complex tasks, making it better suited for handling sophisticated analytical queries and generating detailed reports, as well as understanding and maintaining context across multiple interactions as it responds to complex queries. GPT-4's enhanced structuring and organization capabilities result in more comprehensive and well-structured reports based on the testing results, making the information clearer and easier to implement for users.

**Table 3.** The responses of GPT-4 and Phi-3-mini were graded against four criteria on a score of 0-4 represented by the Harvey balls. The black and blue Harvey balls are the GPT-4 and Phi-3-mini scores respectively. A set of responses can be found in Appendix E.

| Question | Understanding of user intent | Interpretation of data | Quality of recommendations | Which LLM gave a better response? | Explanation |
|---|---|---|---|---|---|
| 1. Advise the planner on how to assess the impact of changing one of the pricing schedules at one of the ERP gantries. | ● (black full) / ◐ (blue) | ◐ (black) / ◐ (blue) | ◐ (black) / ◐ (blue) | GPT-4 | GPT-4 response is superior due to its clear, comprehensive, and actionable structure, ensuring all necessary steps are covered and easily understood. It provides a holistic approach, from planning and data collection to stakeholder engagement and reporting, making it highly practical for the user's needs. |
| 2. Describe to the planner the fields in the historical dataset that will be needed for the analysis. | ◔ (black) / ◔ (blue) | ◐ (black) / ◔ (blue) | ◕ (black) / ◔ (blue) | GPT-4 | GPT-4 is marginally better, listing more relevant, concise list of dataset fields, accurately interpreting data requirements and explaining their relevance, though including traffic direction flow and penalty rates from Phi-3's response would be beneficial. |
| 3. Describe the steps to estimate the pricing elasticity coefficient | ◔ (black) / ◔ (blue) | ◔ (black) / ◔ (blue) | ◔ (black) / ◐ (blue) | Phi-3-mini | Phi-3-mini's response is marginally better as it includes more comprehensive steps (including providing a formula) and advanced statistical techniques, making it suitable for thorough analysis |

| 4. Gather appropriate pricing elasticity coefficients from published studies, to serve as references | ◖ ◔ | ◕ ◔ | ◕ ○ | | GPT-4 | GPT-4's response is better as it provides a concise list of relevant studies, making it easier to use as references. Phi-3 gave irrelevant references, such as Shakespeare's Romeo and Juliet! |

**Table 4.** The responses of GPT-4 and Phi-3-mini were graded against four criteria on a score of 0-4 represented by the Harvey balls. The black and blue Harvey balls are the GPT-4 and Phi-3-mini scores respectively. A set of responses can be found in Appendix F.

| Question | Understanding of user intent | Interpretation of data | Quality of recommendations | Automation of report generation | Which LLM gave a better response? | Explanation |
|---|---|---|---|---|---|---|
| 1. Provide R script to detect common data issues (e.g., NA, outliers, duplicates) | ● ● | ● ◔ | ◕ ◔ | ● ◔ | GPT-4 | GPT-4 excelled in understanding the user's intent and provided a comprehensive R script that accurately detected NA values, outliers, and duplicates with detailed explanations, showcasing expert-level interpretation and analysis. In contrast, while Phi-3-mini also understood the intent and provided a correct script, it lacked the depth and clarity found in GPT-4's response, and its recommendations were less innovative, leading to slightly lower scores in interpretation and report generation. |
| 2. Write an R script to analyze the distribution of travel times and identify peak hours | 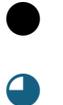 | 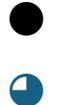 | 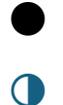 | 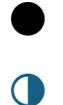 | GPT-4 | GPT-4 demonstrated a superior understanding of the user's intent by providing a detailed and insightful analysis script that effectively identified peak hours and analyzed travel time distributions. It offered relevant, actionable, and innovative recommendations, with an excellent structure for report generation. Phi-3-mini, though accurate, had minor issues with context and provided less innovative recommendations, and its report generation had noticeable gaps, making GPT-4 the better choice for comprehensive analysis. |
| 3. Provide an R script to extract trips that go through Gantry 50. Also Provide a R script to identify trips with origin and destination pairs similar to those that pass through Gantry 50 but do not actually go through it, indicating potential alternative routes. | 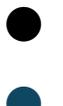 | 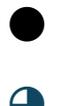 | 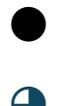 | 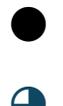 | GPT-4 | Both LLMs showed excellent comprehension and provided accurate scripts to extract trips passing through Gantry 50. However, GPT-4's response was slightly better in terms of depth and clarity, offering more detailed explanations and a structured approach. Phi-3-mini, while effective, had minor analytical issues and its recommendations were less detailed, leading to slightly lower scores in interpretation and report generation. |
| 4. Propose a formula in R to calculate the generalized travel costs | 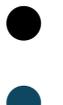 | 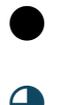 | 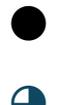 | 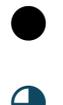 | GPT-4 | GPT-4 stood out by providing a detailed and accurate formula to calculate generalized travel costs, demonstrating expert-level interpretation and insightful analysis. Its recommendations were highly relevant and innovative, and the report generation was comprehensive and well-structured. Phi-3-mini, though accurate in its response, had minor analytical issues and its recommendations were less innovative, resulting in slightly lower scores in interpretation and report generation, making GPT-4 the superior choice for this task. |

| 5. Provide the R script to run sensitivity analyses of how traffic will be reassigned for different percentage increases in congestion pricing. Discuss the limitations of this approach. | 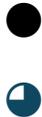 | 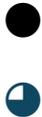 | 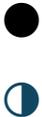 | 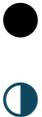 | GPT-4 | GPT-4 provided a more comprehensive and detailed analysis of how changes in congestion pricing would affect traffic reassignment. It demonstrated an excellent understanding of user intent, offered insightful interpretations of the data, and made highly relevant and actionable recommendations. Additionally, its report generation was thorough and well-structured. Phi-3-mini, while effective, provided less depth in analysis and fewer innovative recommendations, and had minor issues in report structuring, leading to lower scores in interpretation and report generation. |

## 4. Future Work

Several areas for future research are suggested.  First is the assessment of other and newer general-purpose LLMs, as well as those that have been fine-tuned with transportation-specific corpora. Since the launch of Phi-3 in April 2024, new frontier models such as llama3 by Meta and GPT-4 Omni by OpenAI have been released.  An example of a transportation-related fine-tuned candidate is TrafficSafetyGPT, a novel LLAMA-based model, which has undergone supervised fine-tuning using a transportation specific dataset consisting of human labels from government-produced guiding books and ChatGPT-generated instruction-output pairs (Zheng et al., 2023).

 Second would be to investigate the impact of Retrieval-Augmented Generation (RAG) techniques on LLM performance in the transportation domain. Ding et al. (2024) comprehensively review existing research studies in retrieval-augmented large language models (RA-LLMs), illustrating their practical significance by application areas in natural language processing and downstream tasks in AI for science, finance, software engineering, and recommendations. There is an opportunity to conduct research evaluating RAG-LLM in the transportation domain.

A third area of future work would be benchmarking the LLMs against other transportation-specific data and transportation-related software. Examples include data formats such as TransXML (Zheng E. , 2007), which is specifically designed for freight transportation data, and transportation planning software like PTV Vissim for microscopic traffic simulation (PTV Group, 2024).  Additionally, it is essential to test the practical applications of LLMs in various real-world transportation scenarios beyond congestion pricing, such as traffic incident management which can enhance public safety and decision support for urban planning and new infrastructure development.

Lastly, future research could explore the synergistic use of task-specific LLMs (such as a fine-tuned Phi-3 mini model) alongside the larger frontier models (exemplified by GPT-4) to solve complex real-world transportation problems. This hybrid approach leverages the specialized capabilities of task-specific models for handling domain-specific tasks while using frontier models to integrate and contextualize outputs, manage broader goals, and provide refinement. Research can examine how this combination can enhance scalability, resource efficiency, and overall problem-solving effectiveness, particularly in domains where specialized knowledge and broad contextual understanding are both crucial.

## 5. Conclusion

This study detailed an exploration of the transformative potential of Generative AI (GenAI) for urban transportation and logistics. The objectives were to evaluate the performance and spatial comprehension of LLMs representing different development paradigms, specifically GPT-4 and Phi-3-mini, within a transportation-informed evaluation framework. The methodology involved subjecting the two LLMs to three levels of testing with increasing knowledge specialization to yield a nuanced understanding of their strengths and weaknesses in grasping geospatial concepts and their applicability in real-world urban transportation management. Results indicate that GPT-4 demonstrates superior accuracy and reliability across various GIS and transportation-specific tasks compared to Phi-3-mini, highlighting its potential as a robust tool for transportation planners. Nonetheless, Phi-3-mini exhibits competence in specific analytical scenarios, suggesting its utility in resource-constrained environments. The findings underscore the transformative potential of GenAI technologies in urban transportation planning, paving the way for more efficient, sustainable, and informed decision-making processes. Future work could explore the application of newer LLMs and the impact of Retrieval-Augmented Generation (RAG) techniques on a broader set of real-world transportation planning and operations challenges, to deepen the integration of advanced AI models in transportation management practices.

# Appendix A
# GIS Examination Questions

The following 60 questions were designed by Mooney (2023) and used to evaluate the Test LLMs' basic understanding of geospatial knowledge.

| S/N | Question | Ground Truth | Topic | Type | GPT-4 response | Phi-3 response |
|---|---|---|---|---|---|---|
| 1 | A degree of longitude spans approximately 110,574 meters at the Equator. How many meters are spanned by a second of longitude at the Equator? {'a': 30.7, 'b': 22.2, 'c': 123} | ['a'] | basic_map | Simple choice | b | a |
| 2 | Three seconds of longitude span approximately 16.1 meters at a latitude of 80 degrees. How many meters are spanned by three minutes of longitude at 80 degrees? {'a': 30.7, 'b': 966.0, 'c': 1024.6} | ['b'] | basic_map | Simple choice | b | b |
| 3 | Convert the Longitude -88 04' 04.08" from DMS to DD (enter degrees, the decimal, and at least the 4 digits to the right of the decimal, e.g., 174.1234) {'a': -88.0678, 'b': 88.0678, 'c': 174.1234} | ['a'] | basic_map | Simple choice | a | a |
| 4 | Is the following statement true or false? A given arc distance (e.g., one degree) is almost the same surface distance anywhere on the globe in a north-south direction, but a degree in an east-west direction varies quite substantially, depending on where you are on the globe. Said differently, a degree of arc spans a more uniform Earth surface distance (e.g., in meters, kilometers, etc.) at different parts of the globe when measuring in a north-south direction (across latitudes) than in an east-west direction (across longitudes). {'a': True, 'b': False} | ['a'] | basic_map | True/False | a | a |
| 5 | When longitude/latitude earth coordinates are plotted on a Cartesian plane, shape distortion is greatest in what direction? {'a': 'north-south', 'b': 'east-west', 'c': 'at oblique angles', 'd': "they're equal in all directions"} | ['b'] | basic_map | Simple choice | a | b |
| 6 | Assume an Earth radius of 6,278 kilometers. What is the ground surface distance, expressed in meters, of an arc the spans 14.2 seconds of | ['c'] | basic_map | Simple choice | c | b |

| S/N | Question | Ground Truth | Topic | Type | GPT-4 response | Phi-3 response |
|---|---|---|---|---|---|---|
| | arc? Please enter your answer to the nearest 0.1 meter {'a': 864.4, 'b': 492.5, 'c': 432.2} | | | | | |
| 7 | Which will hold the largest number? {'a': '4 byte signed binary', 'b': 'two byte signed binary', 'c': '5 byte ASCII', 'd': '4 byte unsigned binary', 'e': 'text', 'f': '1 byte unsigned integer'} | ['d'] | basic_ map | Simple choice | d | d |
| 8 | What type of attribute is political party of the current governor for a states data layer for the U.S.? {'a': 'interval/ratio', 'b': 'Nominal', 'c': 'Ordinal'} | ['b'] | basic_ map | Simple choice | b | b |
| 9 | What type of attribute is human population (the number of people) in a U.S. county data layer? {'a': 'interval/ratio', 'b': 'Nominal', 'c': 'Ordinal'} | ['a'] | basic_ map | Simple choice | a | a |
| 10 | Is the following statement true or false? Vector data models are typically simpler than raster data models. {'a': True, 'b': False} | ['b'] | basic_ map | True/ False | a | b |
| 11 | Is the following statement true or false? Vector data models are often better for describing discrete themes such as counties, voting areas, or zip codes? {'a': True, 'b': False} | ['a'] | basic_ map | True/ False | a | a |
| 12 | Is the following statement true or false? Line thinning is often required when converting from a digital vector layer to a raster layer. {'a': True, 'b': False} | ['a'] | basic_ map | True/ False | a | a |
| 13 | Is the following statement true or false? It is best practice to convert all data used in a GIS project to the same datum and coordinate system {'a': True, 'b': False} | ['a'] | basic_ map | True/ False | a | a |
| 14 | Is the following statement true or false? Grid north is in the direction of the north pole {'a': True, 'b': False} | ['b'] | basic_ map | True/ False | b | a) Tr ue |
| 15 | Is the following statement true or false? Projections unavoidably distort distance, area, or direction measurements on maps, often all three. {'a': True, 'b': False} | ['a'] | basic_ map | True/ False | a | a |
| 16 | Is the following statement true or false? Each projection has both a forward and an inverse, from latitude/longitude to projected coordinates, and from projected coordinates to latitude/longitude. {'a': True, 'b': False} | ['a'] | basic_ map | True/ False | a | a |

| S/N | Question | Ground Truth | Topic | Type | GPT-4 response | Phi-3 response |
|---|---|---|---|---|---|---|
| 17 | Is the following statement true or false? A larger scale map covers less ground than a smaller scale map of the same physical size {'a': True, 'b': False} | ['a'] | basic_gis | True/False | a | b) False |
| 18 | Is the following statement true or false? Metadata are information about the structure, extent, content, and other important characteristics of a data set. {'a': True, 'b': False} | ['a'] | basic_gis | True/False | a | a |
| 19 | Identify the transformation type for the following equation set: E = 510,007 + 0.523*X + 0.11*Y N = 4,957,223 + 0.203*X + 0.793*Y {'a': 'Affine', 'b': '2nd order polynomial', 'c': 'Conformal', 'd': 'Transverse Mercator', 'e': '3rd order polynomial'} | ['a'] | basic_gis | Simple choice | a | b |
| 20 | Select the highest positional error source, on average, for GPS data, when collecting in dense forest with an inexpensive receiver and no external antenna, but using WAAS differential correction. {'a': 'Multipath', 'b': 'Ionospheric and athmospheric delays', 'c': 'Projection errors', 'd': 'Antenna miscalibration'} | ['a'] | basic_gis | Simple choice | a | b |
| 21 | Indicate the GNSS collection setup that is most appropriate, supporting timely and cost-efficient data collection and processing, while meeting accuracy requirements under the given conditions. What is (are) the best GNSS system for collecting 100s of kilometers of forested trail location data, with 2 meters average accuracy, much of it in forested conditions: {'a': 'Hand-held, consumer-grade GNSS system, autonomous ', 'b': 'Precise Point Positioning', 'c': 'Carrier-phase, dual frequency, post-processed GNSS', 'd': 'Consumer-grade WAAS GNSS', 'e': 'Carrier-phase, dual frequency RTK system'} | ['a', 'd'] | basic_gis | Multiple choice | d | e |
| 22 | You are building a GIS database, and are considering data collected via GPS by another department. The department collected point | ['c'] | basic_gis | Simple choice | c | c |

| S/N | Question | Ground Truth | Topic | Type | GPT-4 response | Phi-3 response |
|---|---|---|---|---|---|---|
| | data, averaging 100 fixes per point. You are mapping streams, rivers, and drainage ditches, and require data accurate to the nearest 1 meter. Your target data are in NAD83(NSRS2007) coordinates. Select the data you can accept without further processing (you don't need to project, transform the datum, or differentially correct). C/A code receivers are consumer-grade, e.g., phones or small handhelds, typically without a ground-plane antenna, while carrier phase receivers are professional grade, with enhanced antennas and multi-frequency measurements. {'a': 'collected in NAD27, C/A code, no correction', 'b': 'collected in NAD83(86), C/A code, differentially corrected', 'c': 'collected in NAD83(CORS96), carrier phase, differentially corrected', 'd': 'collected in WGS84(G1150), C/A code, differentially corrected'} | | | | | |
| 23 | Select the characteristics that are true for infrared radiation {'a': 'vegetation reflectances are higher than for visible wavelenghts', 'b': 'wavelengths are shorter than for visible wavelengths', 'c': 'Atmospheric absorption is slightly higher than for visible wavelengths', 'd': 'shows a larger difference in reflection between water and vegetation than visible wavelengths', 'e': 'is typically less expensive to collect than visible wavelengths'} | ['a', 'c', 'd'] | basic_gis | Multiple choice | a, c, d | a c d |
| 24 | Is the following statement true or false? LiDARs are useful for measuring building heights {'a': True, 'b': False} | ['a'] | basic_gis | True/False | a | a |
| 25 | Is the following statement true or false? Radar can be used at night with no loss of information {'a': True, 'b': False} | ['a'] | basic_gis | True/False | a | a |
| 26 | Is the following statement true or false? Image scale is generally the same across most of the frame of most large-scale images {'a': True, 'b': False} | ['b'] | basic_gis | True/False | b | a |
| 27 | Is the following statement true or false? All other things being equal, distortion is higher in | ['a'] | basic_gis | True/False | a | a |

| S/N | Question | Ground Truth | Topic | Type | GPT-4 response | Phi-3 response |
|---|---|---|---|---|---|---|
| | images over areas with greater terrain relief than in images over areas with less terrain relief {'a': True, 'b': False} | | | | | |
| 28 | Select the advantages of open sourced (crowd sourced) data over other types of data: ? {'a': 'Many developers, so wider coverage', 'b': 'Utilizes local knowledge', 'c': 'Is standardized across many regions', 'd': 'The data are uniform in detail', 'e': 'Data are oftem available in many widely-used formats', 'f': 'Data are uniform in quality'} | ['a', 'b', 'c'] | vector _relati onal | Multipl e choice | a, b, e | a --- --- - |
| 29 | Select the advantages of map service data over local data: {'a': 'They save disk space', 'b': 'A single, standard copy may be maintained across a wide range of users', 'c': 'They are always accessible in an acceptable time frame', 'd': 'You can manipulate the data with very few restrictions', 'e': 'Data are uniform in quality across layers'} | ['a', 'b'] | vector _relati onal | Multipl e choice | a | b = = = = = = = = |
| 30 | Is the following statement true or false? A client in a DBMS is the person using the system. {'a': True, 'b': False} | ['b'] | vector _relati onal | True/ False | a | a |
| 31 | Select the ONE answer below that BEST and MOST COMPLETELY describes why multiâ€• tier database architectures are better than singleâ€• tier, non-database file systems (e.g. flat, text or spreadsheet files) {'a': 'insulate the user from the DBMS', 'b': 'better jamdle multiple users', 'c': 'allow simultaneous, shared access', 'd': 'all of the above', 'e': 'none of the above'} | ['d'] | vector _relati onal | Simple choice | d | d |
| 32 | Select ALL of the following that are always true for relational tables in 3rd normal form (may be more than 1). {'a': 'Every non-key attribute is dependent on the primary key', 'b': 'The rows are unique', 'c': 'There are no transitive functional dependencies', 'd': 'There are no repeate columns'} | ['a', 'b', 'c', 'd'] | vector _relati onal | Multipl e choice | a, b, c | ab cd d --- |

| S/N | Question | Ground Truth | Topic | Type | GPT-4 response | Phi-3 response |
|-----|----------|--------------|-------|------|----------------|----------------|
| 33 | Is the following statement true or false? Dissolve operations generally result in more records in an output data layer, relative to the input data layer. {'a': True, 'b': False} | ['b'] | spatial _analy sis | True/ False | b | b |
| 34 | Is the following statement true or false? Proximity functions may be applied to both raster and vector data {'a': True, 'b': False} | ['a'] | spatial _analy sis | True/ False | a | a |
| 35 | Is the following statement true or false? Spatial selection operations may be based on a zoning effect, a size effect, or a modified area effect {'a': True, 'b': False} | ['a'] | spatial _analy sis | True/ False | a | a |
| 36 | Is the following statement true or false? Map algebra expressions may contain no more than two layers {'a': True, 'b': False} | ['b'] | spatial _analy sis | True/ False | b | b |
| 37 | Is the following statement true or false? Cell sizes must only be the same size and have parallel grid direction when combining layers in map algebra {'a': True, 'b': False} | ['b'] | spatial _analy sis | True/ False | b | b |
| 38 | Is the following statement true or false? Map algebra operations may be local, neighborhood, or global {'a': True, 'b': False} | ['a'] | spatial _analy sis | True/ False | a | a |
| 39 | Select the correct answer from the list. Contour lines usually show: {'a': 'lines of equal aspect', 'b': 'lines of equal elevation', 'c': 'lines of equal slope', 'd': 'lines of steepest uphill direction', 'e': 'flowlines', 'f': 'none of the other choices are correct'} | ['b'] | spatial _analy sis | Simple choice | b | b |
| 40 | Is the following statement true or false? Profile curvature is measured in a direction at right angles to the local downslope direction {'a': True, 'b': False} | ['b'] | spatial _analy sis | True/ False | b | a) Tr ue |
| 41 | Is the following statement true or false? Shaded relief plots quantitatively depict elevation heights. {'a': True, 'b': False} | ['b'] | spatial _analy sis | True/ False | b | a |
| 42 | Is the following statement true or false? Specific catchment area is the area drained divided by the area of an individual cell. {'a': True, 'b': False} | ['b'] | spatial _analy sis | True/ False | a | b) Fa ls e |
| 43 | The increasing the exponent in inverse-distance weighting, say, from 2 to 3, has what affect on the influence of a point in the region | ['a'] | intro_s patstat | Simple choice | a | a |

| S/N | Question | Ground Truth | Topic | Type | GPT-4 response | Phi-3 response |
|---|---|---|---|---|---|---|
| | near that point: {'a': 'A point becomes more influential locally, with interpolated values more like that point', 'b': 'The point becomes less influential locally, with interpolated values more affected by the entire set of points', 'c': 'It has no affect on local influence.'} | | _interpol | | | |
| 44 | Is the following statement true or false? Moran's I is a measure of spatial autocorrelation {'a': True, 'b': False} | ['a'] | intro_spatstat_interpol | True/False | a | a) True |
| 45 | Is the following statement true or false? Interpolated surfaces almost always contain some error. {'a': True, 'b': False} | ['a'] | intro_spatstat_interpol | True/False | a | a) True |
| 46 | Choose the interpolation method(s) that often create 'abrupt' or sharp boundaries in the interpolated surface, even when the underlying true data are continuous. {'a': 'nearest neighbor', 'b': 'fixed radius', 'c': 'inverse distance', 'd': 'splines', 'e': 'trend surface', 'f': 'kriging'} | ['a'] | intro_spatstat_interpol | Simple choice | a | bdf |
| 47 | Is the following statement true or false? Systematic sampling is the method most prone to bias due to repeating patterns in the landscape {'a': True, 'b': False} | ['a'] | intro_spatstat_interpol | True/False | a | b |
| 48 | Is the following statement true or false? In sampling we control either the number of points or the distribution of points, or both. {'a': True, 'b': False} | ['b'] | intro_spatstat_interpol | True/False | a | a |
| 49 | Is the following statement true or false? Spatial regression is an exact interpolator. {'a': True, 'b': False} | ['b'] | intro_spatstat_interpol | True/False | b | b) False |
| 50 | Is the following statement true or false? Spatial regression may include variables in addition to the target variable and location when estimating values. {'a': True, 'b': False} | ['a'] | intro_spatstat_interpol | True/False | a | a |
| 51 | Is the following statement true or false? Kernel mapping requires user-specified parameters {'a': True, 'b': False} | ['a'] | intro_spatstat_interpol | True/False | a | b |

| S/N | Question | Ground Truth | Topic | Type | GPT-4 response | Phi-3 response |
|---|---|---|---|---|---|---|
| 52 | Is the following statement true or false? Optimum bandwidth to select in kernel mapping is usually clearly defined. {'a': True, 'b': False} | ['b'] | intro_s patstat _interp ol | True/ False | b | b |
| 53 | Is the following statement true or false? The Thiessen polygon method is an exact interpolator {'a': True, 'b': False} | ['a'] | intro_s patstat _interp ol | True/ False | b | a |
| 54 | Is the following statement true or false? Fix radius sampling estimates depend on a search distance {'a': True, 'b': False} | ['a'] | intro_s patstat _interp ol | True/ False | a | a |
| 55 | Is the following statement true or false? Interpolated values in an inverse-distance weighted interpolator depend on parameters set by the analyst. {'a': True, 'b': False} | ['a'] | intro_s patstat _interp ol | True/ False | a | a |
| 56 | Is the following statement true or false? Kriging uses equal weights for sample points when estimating at an unknown location. {'a': True, 'b': False} | ['b'] | intro_s patstat _interp ol | True/ False | b | b |
| 57 | Is the following statement true or false? Convex hulls may have concave external boundaries {'a': True, 'b': False} | ['b'] | intro_s patstat _interp ol | True/ False | b | b (N ot e: |
| 58 | Is the following statement true or false? Convex hulls require user-specified parameters. {'a': True, 'b': False} | ['b'] | intro_s patstat _interp ol | True/ False | b | a |
| 59 | You are asked to do a spatial analysis that may include some of the following data layers: 1) BOUND - Study area boundary (vector polygon); 2) FEMA - 100-year floodplain map (vector poly); 3) CENSUS population block data (vector poly); 4) ROAD and rail data (vector line); 5) NASS landcover data (raster, 25 m res.) 6) DEM - USGS 10m DEM (raster, 10 m); 7) WETLAND - USFWS data (vector polys) | ['a'] | applie d_anal ysis_fl ow | Simple choice | a | b |

| S/N | Question | Ground Truth | Topic | Type | GPT-4 response | Phi-3 response |
|---|---|---|---|---|---|---|
| | Select the sequence of steps that comes closest to describing how to complete your task:<br><br>Identify flat building sites, outside of the floodplain, within 1/4 mile of a road.<br><br>Note that intermediate layers in the answers are written in ALL CAPS, and --> means output  {'a': 'Select roads from ROAD, simple buffer -> NEAR_RD; \nDEM to SLOPE, then reclass to FLAT; \nUnion BOUND/FEMA, discard flood areas --> DRY; union NEAR_RD, FLAT, DRY, select area matching criteria from output', 'b': 'Select roads from ROAD, simple buffer -> NEAR_RD; \nSelect FEMA areas in 100 year floodplain--> DRY; Select flat areas on DEM --> FLAT; \nUnion FLAT, DRY, NEAR_RD, and select appropriate areas;', 'c': 'DEM to SLOPE, then reclass to FLAT; \nUnion BOUND/FEMA, discard flood areas --> DRY; \nSimple buffer ROAD --> INDIST; Union INDIST, FLAT, & DRY, and select areas in all.', 'd': 'Select non-wetland area from NASS for DRY; \nSelect roads from ROAD, simple buffer -> NEAR_RD; \nDEM to SLOPE, then reclass to FLAT; \nunion NEAR_RD, FLAT, DRY, select area matching criteria from output '} | | | | | |
| 60 | You are asked to do a spatial analysis that may include some of the following data layers:<br>1) BOUND - Study area boundary (vector polygon);<br>2) FEMA - 100-year floodplain map (vector poly);<br>3) CENSUS population block data (vector poly);<br>4) ROAD and rail data (vector line);<br>5) NASS landcover data (raster, 25 m res.)<br>6) DEM - USGS 10m DEM (raster, 10 m);<br>7) WETLAND - USFWS data (vector polys) | ['a'] | applied_analysis_flow | Simple choice | a | b |

| S/N | Question | Ground Truth | Topic | Type | GPT-4 response | Phi-3 response |
|---|---|---|---|---|---|---|
| | Select the sequence of steps that comes closest to describing how to complete your task: IIdentify the wetlands that are each larger than 10 hectares that are a) within 200 meters of a city or other densely populated place, and b) have at least 50% of the surrounding 9 square kilometers in agriculture. Note that intermediate layers in the answers are written in ALL CAPS, and --> means output. Note that you might want to draw a sketch of the analysis description, starting a new branch each time a new data layer is introduced, and creating a branch from existing data layers that have been created or used in the analysis. {'a': 'Select all wetlands from WETLAND, dissolve, calculate area, and select those > 10 hectares --> 10WTL; \nselect city by high population density from CENSUS, reclass, dissolve, buffer at 200m --> NEARCT; \nbuffer each wetland at approx. 1.6 km (give 9sq km area) , without dissolving output across separate wetlands --> WTBUFF; reclassify NASS to Corn/noncorn --> CRN_RC; \nselect 10WTL by location, against NEARCT --> CTWET; \nintersect CTWET with CRN_RC, once for each individual wetland --> CRN_WET; \nSummarize area for CRN_WET, select those that have greater than 50% area in corn nearby --> FINAL LAYER.', 'b': 'Select city NLCD, reclass, dissolve, buffer at 200m --> CTBUFF; \nselect all wetlands from WETLAND, calculate area, and select those > 10 hectares --> 10WTL;buffer each wetland, dissolving output across separate wetlands --> WTBUFF; \nreclassify NLCD to crop/noncrop --> CRN_RC; \nselect 10WTL by location, against CTBUF --> CTWET; \nintersect CTWET with CRN_RC --> | | | | | |

| S/N | Question | Ground Truth | Topic | Type | GPT-4 response | Phi-3 response |
|---|---|---|---|---|---|---|
| | CRN_WET; \summarize area for CRN_WET, select those that have greater than 50% area in crop nearby --> FINAL LAYER. ', 'c': 'Select all wetlands from WETLAND, dissolve, calculate area, and select those > 10 hectares --> 10WTL; \buffer each wetland in 10WTL, without dissolving output across separate wetlands --> WTBUFF; \select city by high population density from CENSUS, reclass, dissolve --> NEARCT; \select 10WTL by location, against NEARCT --> CTWET; \intersect CTWET with NASS summarize area for CT_WET, select those that have greater than 50% area in corn nearby --> FINAL LAYER. ', 'd': 'Select city NLCD, reclass, dissolve, buffer at 200m --> CTBUFF; \select all wetlands from WETLAND, calculate area, and select those > 10 hectares --> 10WTL; \buffer each wetland, dissolving output across separate wetlands --> WTBUFF; \reclassify NLCD to crop/noncrop --> CRN_RC; \select 10WTL by location, against CTBUF --> CTWET; \intersect CTWET with CRN_RC --> CRN_WET; \summarize area for CRN_WET, select those that have greater than 50% area in crop nearby --> FINAL LAYER.'} | | | | | |

# Appendix B
## Understanding Tabular Dataset Questions

A system prompt is first provided to the Test LLMs to set up the context, including explanation of the different data fields in the MATSim-generated tabular dataset:

You are a helpful assistant to respond to questions using the information in attached dataset which has the fields shown below. You will be posed questions that will require you to retrieve and process information from this file.

You are to return a quantitative response, usually a number or a vector of numbers. If the length of the vector of numbers is more than 10, print out only the first 10 entries and highlight that these are only the first 10 elements of the vector.

The field names and descriptions are as follows.
LINK : Unique link identifier; ORIG_ID : Not used; FROM : ID of start location of link; TO : ID of end location of link; LENGTH : Distance of the link in metres; FREESPEED : Speed limit of vehicles in metres / second; CAPACITY : Capacity of road link in terms of number of vehicles in an hour; HRS0-1avg : Hourly number of vehicles using the road link between 0 and 1hrs; HRS1-2avg : Hourly number of vehicles using the road link between 1 and 2hrs; HRS2-3avg : Hourly number of vehicles using the road link between 2 and 3hrs; HRS3-4avg : Hourly number of vehicles using the road link between 3 and 4hrs; HRS4-5avg : Hourly number of vehicles using the road link between 4 and 5hrs; HRS5-6avg : Hourly number of vehicles using the road link between 5 and 6hrs; HRS6-7avg : Hourly number of vehicles using the road link between 6 and 7hrs; HRS7-8avg : Hourly number of vehicles using the road link between 7 and 8hrs; HRS8-9avg : Hourly number of vehicles using the road link between 8 and 9hrs; HRS9-10avg : Hourly number of vehicles using the road link between 9 and 10hrs; HRS10-11avg : Hourly number of vehicles using the road link between 10 and 11hrs; HRS11-12avg : Hourly number of vehicles using the road link between 11 and 12hrs; HRS12-13avg : Hourly number of vehicles using the road link between 12 and 13hrs; HRS13-14avg : Hourly number of vehicles using the road link between 13 and 14hrs; HRS14-15avg : Hourly number of vehicles using the road link between 14 and 15hrs; HRS15-16avg : Hourly number of vehicles using the road link between 15 and 16hrs; HRS16-17avg : Hourly number of vehicles using the road link between 16 and 17hrs; HRS17-18avg : Hourly number of vehicles using the road link between 17 and 18hrs; HRS18-19avg : Hourly number of vehicles using the road link between 18 and 19hrs; HRS19-20avg : Hourly number of vehicles using the road link between 19 and 20hrs;  HRS20-21avg : Hourly number of vehicles using the road link between 20 and 21hrs;  HRS21-22avg : Hourly number of vehicles using the road link between 21 and 22hrs;  HRS22-23avg : Hourly number of vehicles using the road link between 22 and 23hrs;  HRS23-24avg : Hourly number of vehicles using the road link between 23 and 24hrs;  HRS0-24avg : Daily number of vehicles using the road link between 0 and 1hrs;  TRAVELTIME0-1avg : Hourly travel time of vehicles using the road link between 0 and 1hrs;  TRAVELTIME1-2avg : Hourly travel time of vehicles using the road link between 1 and 2hrs;  TRAVELTIME2-3avg : Hourly travel time of vehicles using the road link between 2 and 3hrs;  TRAVELTIME3-4avg : Hourly travel time of vehicles using the road link between 3 and 4hrs;  TRAVELTIME4-5avg : Hourly travel time of vehicles using the road link between 4 and 5hrs;  TRAVELTIME5-6avg : Hourly travel time of vehicles using the road link between 5 and 6hrs;  TRAVELTIME6-7avg : Hourly travel time of vehicles using the road link between 6 and 7hrs;  TRAVELTIME7-8avg : Hourly travel time of vehicles using the road link between 7 and 8hrs;  TRAVELTIME8-9avg : Hourly travel time of vehicles using the road link between 8 and 9hrs;  TRAVELTIME9-10avg : Hourly travel time of vehicles using the road link between 9 and 10hrs;  TRAVELTIME10-11avg : Hourly travel time of vehicles using the road link between 10 and 11hrs;  TRAVELTIME11-12avg : Hourly travel time of vehicles using the road link between 11 and 12hrs;  TRAVELTIME12-13avg : Hourly travel time of vehicles using the road link between 12 and 13hrs;  TRAVELTIME13-14avg : Hourly travel time of vehicles using the road link between 13 and 14hrs;  TRAVELTIME14-15avg : Hourly travel time of vehicles using the road link between 14 and 15hrs;  TRAVELTIME15-16avg : Hourly travel time of vehicles using the road link between 15 and 16hrs;  TRAVELTIME16-17avg : Hourly travel time of vehicles using the road link between 16 and 17hrs;  TRAVELTIME17-18avg : Hourly travel time of vehicles using the road link between 17 and 18hrs;  TRAVELTIME18-19avg : Hourly travel time of vehicles using the road link between 18 and 19hrs;  TRAVELTIME19-20avg : Hourly travel time of vehicles using the road link between 19 and 20hrs;  TRAVELTIME20-21avg : Hourly travel time of vehicles using the road link between 20 and 21hrs;  TRAVELTIME21-22avg : Hourly travel time of vehicles using the road link between 21 and 22hrs;  TRAVELTIME22-23avg : Hourly travel time of vehicles using the road link between 22 and 23hrs;  TRAVELTIME23-24avg : Hourly travel time of vehicles using the road link between 23 and 24hrs.

All the travel time are measured in seconds.

The following list of 15 questions are then posed.

| | Question | Type | GPT-4 response | Phi-3 response | GPT-4 score | Phi-3 score |
|---|---|---|---|---|---|---|
| 1 | What is the average hourly traffic volume for the day for each link? | Traffic volume analysis | To calculate the average hourly traffic volume for each link, we need to follow these steps:<br><br>1. Load the dataset into a DataFrame.<br>2. Select the columns with hourly traffic volume data (HRS0-1avg, HRS1-2avg, ..., HRS23-24avg).<br>3. Compute the mean of these columns for each row (link) to get the average hourly traffic volume for each link.<br><br>Here's a Python code that demonstrates these steps using the pandas library:<br><br>```python<br>import pandas as pd<br><br># Load the dataset (assuming it's a CSV file)<br>dataset = pd.read_csv('traffic_flow_data.csv')<br><br># Select the columns with hourly traffic volume data<br>hourly_volume_columns = [f'HRS{i}-{i+1}avg' for i in range(24)]<br><br># Calculate the mean of the selected columns for each row (link)<br> | To find the daily average hourly traffic volume, we first need to calculate the sum of all hourly vehicle counts and then divide by the number of hours in a day (24). Here's an example Python code snippet:<br><br>```python<br># Sample dataset with a single link for demonstration purposes<br>data = {<br>    "ORIGIN": "A",<br>    "TO": "B",<br>    "LENGTH": 100, # Distance in metres<br>    "FREESPEED": 60, # Speed limit in metres per second<br>    "CAPACITY": 4000, # Capacity of link in vehicles per hour<br>    # Assuming data for HRS0-24avg is the daily total number of vehicles using the road link between 0 and 24hrs<br>    "HRS0-24avg": 7500<br>}<br><br># Calculate daily average hourly traffic volume<br>daily_average_hourly_volume = data["HRS0-24avg"] / 24 | 10 | 9 |

| | Question | Type | GPT-4 response | Phi-3 response |  |  |
|---|---|---|---|---|---|---|
| | | | average_hourly_volume = dataset[hourly_volume_columns].mean(axis=1)<br><br># Add the result to the dataset as a new column dataset['AverageHourlyVolume'] = average_hourly_volume<br>```<br><br>After running this code, the dataset DataFrame will have a new column called "AverageHourlyVolume" that contains the average hourly traffic volume for each link. | print("Daily average hourly traffic volume:", daily_average_hourly_volume, "vehicles per hour.")<br>```<br><br>Replace the sample dataset with your real dataset to find the daily average for each link in your actual project. | | |
| 2 | Which links have the highest and lowest traffic volumes? | Traffic volume analysis | To find the links with the highest and lowest traffic volumes, we need to perform the following analytical steps:<br><br>1. Load the dataset.<br>2. Extract the "LINK" and "HRS0-24avg" columns.<br>3. Sort the extracted data by "HRS0-24avg" in descending order.<br>4. Find the link with the highest traffic volume (the first entry) and the link with the lowest traffic volume (the last entry).<br><br>Here's a Python code snippet to perform these steps: | To find the links with the highest and lowest traffic volumes, you can follow these steps using the provided dataset:<br><br>Step 1: Read in the traffic flow dataset. This will give access to all the relevant data points about each link' FFLEASETIME0-24avg' for a comprehensive view of road link usage and conditions throughout the day.<br><br>```python<br>import pandas as pd<br><br># Assuming 'traffic_flow_data.csv' contains our traffic dataset | 10 | 5 |

| | Question | Type | GPT-4 response | Phi-3 response | GPT-4 score | Phi-3 score |
|---|---|---|---|---|---|---|
| | | | ` ```python `<br>`import pandas as pd`<br><br>`# Load the dataset`<br>`data = pd.read_csv("traffic_flow_dataset.csv")`<br><br>`# Extract the "LINK" and "HRS0-24avg"`<br>`columns`<br>`link_traffic = data[["LINK", "HRS0-24avg"]]`<br><br>`# Sort the extracted data by "HRS0-24avg" in`<br>`descending order`<br>`sorted_link_traffic =`<br>`link_traffic.sort_values(by="HRS0-24avg",`<br>`ascending=False)`<br><br>`# Find the link with the highest traffic volume`<br>`(the first entry)`<br>`highest_traffic_link =`<br>`sorted_link_traffic.iloc[0]`<br><br>`# Find the link with the lowest traffic volume`<br>`(the last entry)`<br>`lowest_traffic_link = sorted_link_traffic.iloc[-`<br>`1]` | `df = pd.read_csv('traffic_flow_data.csv')`<br>` ``` `<br><br>Step 2: Extract only the CAPACITY and HRS0-1avg (daily) fields since these are indicators of high and low volumes of traffic, respectively.<br><br>` ```python `<br>`highest_vol = df[['HRS0-1avg',`<br>`'CAPACITY']].max()`<br>`lowest_vol = df[['HRS0-1avg',`<br>`'CAPACITY']].min()`<br>` ``` `<br><br>Step 3: Find the link associated with these maximum and minimum values. Note that in our case, we will need to handle multiple links having the same highest or lowest capacity by selecting any one of them since there is no single identifier provided (e.g., a unique Link_ID).<br><br>` ```python `<br>`# For finding all links with the highest volume,`<br>`we would do this:`<br>`highest_volume_links = df[df['HRS0-1avg'] ==` | | |

| | Question | Type | GPT-4 response | Phi-3 response | GPT-4 score | Phi-3 score |
|---|---|---|---|---|---|---|
| | | | print("Highest traffic volume link:", highest_traffic_link) print("Lowest traffic volume link:", lowest_traffic_link) ``` <br><br> This code first imports the required pandas library and reads the dataset using `pd.read_csv()`. It then extracts the "LINK" and "HRS0-24avg" columns and sorts the data by "HRS0-24avg" in descending order. Finally, it finds the first and last entries, which represent the highest and lowest traffic volume links, and prints them. | highest_vol['HRS0-1avg']] <br><br> # And similarly for the lowest volume: lowest_volume_links = df[df['HRS0-1avg'] == lowest_vol['HRS0-1avg']] ``` <br><br> The resulting 'highest_volume_links' and 'lowest_volume | | |
| 3 | Which hours experience the highest and lowest average hourly traffic volumes for the entire road network? | Traffic volume analysis | To find the hours with the highest and lowest average hourly traffic volumes for the entire road network, we will follow these steps: <br><br> 1. Load the dataset. 2. Sum the traffic volumes of all the road links for each hourly period (e.g., HRS0-1avg, HRS1-2avg, ..., HRS23-24avg). 3. Identify the maximum and minimum total traffic volumes and their corresponding hours. <br><br> Here's the Python code to perform these steps: | First, you will need to import the necessary libraries and load your dataset: <br><br> ```python import pandas as pd import numpy as np <br><br> traffic_flow = pd.read_csv('traffic_flow.csv') ``` <br><br> Next, we aggregate data across all links for each hour of the day. We calculate both the max and min values along with corresponding | 10 | 5 |

| | Question | Type | GPT-4 response | Phi-3 response | GPT-4 score | Phi-3 score |
|---|---|---|---|---|---|---|
| | | | ` ```python `<br><br>`import pandas as pd`<br><br>`# Load the dataset (assuming it's in a CSV format)`<br>`data = pd.read_csv("traffic_flow_dataset.csv")`<br><br>`# Create a list of hourly traffic volume columns`<br>`hourly_columns = [f"HRS{i}-{i+1}avg" for i in range(24)]`<br><br>`# Calculate the sum of traffic volumes for each hourly period across all road links`<br>`hourly_sums = data[hourly_columns].sum()`<br><br>`# Identify the hours with the highest and lowest traffic volumes`<br>`max_hour = hourly_sums.idxmax()`<br>`min_hour = hourly_sums.idxmin()`<br><br>`print(f"The hour with the highest traffic volume is {max_hour}.")`<br>`print(f"The hour with the lowest traffic volume is {min_hour}.")`<br>` ``` `<br><br>This code loads the dataset, calculates the sum | hours:<br><br>` ```python Written in plain English without technical jargon: `<br>The dataset shows us how many cars use different roads during each part of a day in a city. Each road is like one line on a big map that helps vehicles move from point A to B. Imagine if we list all the cars traveling along these lines for every hour, we're looking at how busy our city's driving paths get throughout the day and night.<br><br>Now let's find out which time is busiest with cars (the highest traffic volume) and when do fewer vehicles pass by (the lowest traffic). To figure this out, imagine you have a big chart listing all these roads along with every hour of the day marked across the top. As we count how many cars drive on each road for each hour, some hours will show us more cars than others.<br><br>For the busiest part, we look for the highest number in our chart. It tells us when most vehicles are using our city's roads. For the least busy part of the day, it's like looking for a | | |

| | Question | Type | GPT-4 response | Phi-3 response | GPT-4 score | Phi-3 score |
|---|---|---|---|---|---|---|
| | | | of traffic volumes for each hour across all road links, and identifies the hours with the highest and lowest traffic volumes. | line that has very few dots (vehicles) across its length, which shows up as the lowest numbers on our chart. By checking each hour and counting cars, we can easily spot these points in time â€" when there are lots of vehicles moving around or when it's quite empty on our roads. | | |
| 4 | What are the peak traffic hours for each link? | Traffic volume analysis | To find the peak traffic hours for each link, we'll perform the following analytical steps:<br><br>1. Load the dataset into a pandas DataFrame.<br>2. For each row (i.e., each link) in the DataFrame, find the maximum value among the columns representing hourly traffic volumes (HRS0-1avg, HRS1-2avg, ..., HRS23-24avg).<br>3. Identify the corresponding hour for the maximum traffic volume.<br>4. Store the link identifier and the peak traffic hour in a dictionary or a new DataFrame.<br><br>Here's the Python code to accomplish these steps:<br><br>```python<br>import pandas as pd<br><br># Load the dataset into a pandas DataFrame<br>data = pd.read_csv('traffic_flow_dataset.csv')``` | To identify the peak traffic hours for each road link, we can analyze the data to find where there is maximum average vehicle count (FROM HRS0-24avg). This requires calculating and comparing averages across all time frames. However, without actual access to the dataset or the ability to process it programmatically as mentioned earlier, a Python script for this analysis isn't feasible in this context. Normally you would write something along these lines if you had data:<br><br>```python<br>import pandas as pd<br><br># Assuming 'traffic_data.csv' is our dataset and 'LINK', 'FROM', etc. are the columns with respective road link information.<br>traffic_df = pd.read_csv('traffic_data.csv')<br><br># Finding average vehicle counts per hour``` | 8 | 2 |

| | Question | Type | GPT-4 response | Phi-3 response | GPT-4 score | Phi-3 score |
|---|---|---|---|---|---|---|
| | | | # Initialize an empty dictionary to store peak traffic hours for each link<br>peak_traffic_hours = {}<br><br># Loop through each row (link) in the dataset<br>for index, row in data.iterrows():<br>    # Extract hourly traffic volume columns<br>    hourly_columns = [col for col in data.columns if 'HRS' in col and 'avg' in col]<br>    hourly_traffic_volumes = row[hourly_columns]<br><br>    # Find the maximum traffic volume and its corresponding hour<br>    max_volume = hourly_traffic_volumes.max()<br>    peak_hour = hourly_traffic_volumes.idxmax()<br><br>    # Store the link identifier and the peak traffic hour in the dictionary<br>    link_id = row['LINK']<br>    peak_traffic_hours[link_id] = peak_hour<br><br># Optionally, convert the dictionary to a pandas DataFrame | across all links<br>peak_traffic_hours = traffic_df.groupby(['FROM']).mean()<br><br># Sorting to find the maximum values, representing peak hours (without real data processing)<br>sorted_by_average_vehicles = peak_traffic_hours.sort_values('AVG(FROM HRS0-24avg)', ascending=False)<br>peak_hours = sorted_by_average_vehicles['AVG(FROM HRS0-24avg)'].idxmax()<br><br>print("The peak traffic hours (assuming data processing could be done) would typically show which hour of the day has the highest average vehicle count.")<br>```<br><br>But remember, to run this effectively with real data, you need appropriate data manipulation libraries like pandas and computational resources. | | |

| | Question | Type | GPT-4 response | Phi-3 response | GPT-4 score | Phi-3 score |
|---|---|---|---|---|---|---|
| | | | peak_traffic_hours_df = pd.DataFrame(list(peak_traffic_hours.items()), columns=['LINK', 'Peak Hour']) ``` <br><br> This code first loads the dataset into a pandas DataFrame and initializes an empty dictionary to store peak traffic hours for each link. Then, it iterates through each row in the dataset, extracts hourly traffic volume columns, and finds the maximum traffic volume and its corresponding hour. Finally, it stores the link identifier and the peak traffic hour | | | |
| 5 | How does traffic volume change throughout the day? | Traffic volume analysis | To analyze how the traffic volume changes throughout the day, we can follow these analytical steps: <br><br> 1. Load the dataset into a data processing library, like pandas in Python. <br> 2. Extract the hourly traffic volume columns (HRS0-1avg to HRS23-24avg). <br> 3. Compute the average traffic volume for each hour across all road links. <br> 4. Plot the average traffic volume against the hour of the day to visualize the changes in traffic volume throughout the day. | To determine how traffic volume changes throughout the day using the dataset, we can analyze the "HRS0-24avg" field, which represents the average number of vehicles on each road link for every hour in a 24-hour period. By plotting these averages across different times of the day, we can identify patterns and trends such as peak hours (usayer's travel volume during rush hour), low traffic periods (off-peak times with fewer cars on the road), and any recurring fluctitations within those broader categories. <br><br> Here's an example Python code snippet that | 9 | 4 |

| | Question | Type | GPT-4 response | Phi-3 response | GPT-4 score | Phi-3 score |
|---|---|---|---|---|---|---|
| | | | peak_traffic_hours_df = pd.DataFrame(list(peak_traffic_hours.items()), columns=['LINK', 'Peak Hour']) ``` <br><br> This code first loads the dataset into a pandas DataFrame and initializes an empty dictionary to store peak traffic hours for each link. Then, it iterates through each row in the dataset, extracts hourly traffic volume columns, and finds the maximum traffic volume and its corresponding hour. Finally, it stores the link identifier and the peak traffic hour | | | |
| 5 | How does traffic volume change throughout the day? | Traffic volume analysis | To analyze how the traffic volume changes throughout the day, we can follow these analytical steps: <br><br> 1. Load the dataset into a data processing library, like pandas in Python. <br> 2. Extract the hourly traffic volume columns (HRS0-1avg to HRS23-24avg). <br> 3. Compute the average traffic volume for each hour across all road links. <br> 4. Plot the average traffic volume against the hour of the day to visualize the changes in traffic volume throughout the day. | To determine how traffic volume changes throughout the day using the dataset, we can analyze the "HRS0-24avg" field, which represents the average number of vehicles on each road link for every hour in a 24-hour period. By plotting these averages across different times of the day, we can identify patterns and trends such as peak hours (usayer's travel volume during rush hour), low traffic periods (off-peak times with fewer cars on the road), and any recurring fluctitations within those broader categories. <br><br> Here's an example Python code snippet that | 9 | 4 |

| | Question | Type | GPT-4 response | Phi-3 response | GPT-4 score | Phi-3 score |
|---|---|---|---|---|---|---|
| | | | Here's a Python code to perform these steps using the pandas library:<br><br>```python<br>import pandas as pd<br>import matplotlib.pyplot as plt<br><br># Load the dataset<br>data = pd.read_csv("traffic_flow_dataset.csv")<br># Extract hourly traffic volume columns<br>hourly_traffic_columns = [f'HRS{i}-{i+1}avg' for i in range(24)]<br>hourly_traffic_data = data[hourly_traffic_columns]<br><br># Compute the average traffic volume for each hour<br>average_hourly_traffic = hourly_traffic_data.mean(axis=0)<br><br># Plot the average traffic volume against the hour of the day<br>plt.plot(range(24), average_hourly_traffic)<br>plt.xlabel("Hour of the Day")<br>plt.ylabel("Average Traffic Volume")<br>plt.title("Average Traffic Volume Throughout | plots a graph to visualize this data:<br><br>```python<br>import matplotlib.pyplot as plt<br><br># Sample dataset representing traffic volume for different hours (assuming 'traffic_data' is available)<br><br>traffic_data = {<br><br>    "Hour": [0, 1, 2, 3, ... , 24],  # Example list of hours from 0 to 23<br><br>    "Volume": [500, 600, 800, ..., X]   # Corresponding sample average vehicle counts per hour (fill in actual data)<br><br>}<br><br># Plotting the graph<br><br>plt.plot(traffic_data["Hour"], traffic_data["Volume"]) | | |

| | Question | Type | GPT-4 response | Phi-3 response |  |  |
|---|---|---|---|---|---|---|
| | | | the Day")<br>plt.grid()<br>plt.show()<br>```<br><br>This code first loads the dataset using pandas and extracts the hourly traffic volume columns. Then, it calculates the average traffic volume for each hour across all road links. Finally, it plots the average traffic volume against the hour of the day to visualize the changes in traffic volume throughout the day. | plt.xlabel('Hours of the Day')<br><br>plt.ylabel('Average Traffic Volume')<br><br>plt.title('Traffic Volume Change Throughout the Day')<br><br>plt.xticks(range(len(traffic_data["Hour"])), ['{:02d}:00'.format(hour) for hour in traffic_data["Hour"]]) # Labels hours on x-axis (e.g., '0 | | |
| 6 | What is the average hourly travel time for the day for each link? | Travel time and speed analysis | To find the average hourly travel time for the day for each link, you need to perform the following analytical steps:<br><br>1. Import the dataset into a suitable data structure, like a pandas DataFrame.<br>2. For each link, calculate the average travel time for each hour by summing up all the hourly travel times and dividing by the number of hours (24).<br>3. Store the result in a new column in the DataFrame and then display or export the results.<br><br>Here's a sample Python code for the analytical | To calculate the daily average hourly travel time (TRAVELTIME0-24avg) for each link, we would perform the following steps:<br><br>1. First, ensure that you have access to a dataset containing traffic volume and travel times data across all hours of the day for multiple road links within a network.<br><br>2. Iterate through each unique link identifier (LINK) in this dataset.<br> Writ | 10 | 8 |

| | Question | Type | GPT-4 response | Phi-3 response | GPT-4 score | Phi-3 score |
|---|---|---|---|---|---|---|
| | | | steps:<br><br>```python<br>import pandas as pd<br><br># Read the dataset into a pandas DataFrame<br>data = pd.read_csv('traffic_flow_dataset.csv')<br><br># Create a list of columns for hourly travel times<br>travel_time_columns = [f'TRAVELTIME{i}-{i+1}avg' for i in range(24)]<br><br># Calculate the average travel time for each link<br>data['AVG_TRAVEL_TIME'] = data[travel_time_columns].sum(axis=1) / 24<br><br># Display or export the results<br>print(data[['LINK', 'AVG_TRAVEL_TIME']])<br>```<br><br>This code imports the pandas library, reads the dataset into a DataFrame, creates a list of hourly travel time columns, calculates the average travel time for each link, and displays or exports the results. | response = """To calculate the daily average hourly travel time, we'll perform an aggregation on the 'TRAVELTIME0-24avg' column for each 'LINK'. Here is a Python code snippet to achieve that:<br><br>```python<br># Assume 'traffic_data' is your DataFrame containing all relevant traffic information.<br>average_hourly_travel = {}<br><br>for index, row in traffic_data.iterrows():<br>    link = row['LINK']<br>    daily_avg_time = row['TRAVELTIME0-24avg']<br><br>    if link not in average_hourly_travel:<br>        average_hourly_travel[link] = []<br><br><br>average_hourly_travel[link].append(daily_avg_time)<br><br># Now we can calculate the daily averages for each link.<br>average_daily_times = {}<br>for link, times in | | |

| | Question | Type | GPT-4 response | Phi-3 response | GPT-4 score | Phi-3 score |
|---|---|---|---|---|---|---|
| | | | | average_hourly_travel.items():<br>    if len(times) > 0:<br>        avg_daily_time = sum(times) / len(times)<br>        average_daily_times[link] = avg_daily_time<br>    else:<br>        # Handle the case where data might not be available for a certain link.<br>        average_daily_times[link | | |
| 7 | What is the implied average speed of travel for each link? | Trave l time and speed analy sis | To find the implied average speed of travel for each link, we will perform the following analytical steps:<br><br>1. Calculate the average travel time for each link by averaging the hourly travel times (TRAVELTIME0-1avg, TRAVELTIME1-2avg, ..., TRAVELTIME23-24avg).<br>2. Divide the length of the link (LENGTH) by the average travel time to get the average speed in meters per second.<br>3. Convert the average speed to a more common unit, such as kilometers per hour (km/h), by multiplying the result by 3.6.<br><br>Here's the python code for these steps:<br><br>```python | ```python<br># First, we create a DataFrame to hold our data:<br>import pandas as pd<br><br>data = {<br>    "LINK": ["Link1", "Link2", "Link3"],  # Assuming some example links; in real scenario replace these with actual link identifiers from the dataset<br>    "ORIG_ID": [0, 1, 2],<br>    "FROM": [10, 15, 20],<br>    "TO": [20, 30, 40],<br>    # ... (other fields are assumed to be populated similarly)<br>}<br>df = pd.DataFrame(data)<br><br># Then we calculate the average speed for each | 10 | 5 |

| | Question | Type | GPT-4 response | Phi-3 response |  |  |
|---|---|---|---|---|---|---|
| | | | import pandas as pd<br><br># Assuming the dataset is in a CSV file called 'traffic_data.csv'<br>traffic_data = pd.read_csv('traffic_data.csv')<br><br># Step 1: Calculate the average travel time for each link<br>travel_time_columns = [f'TRAVELTIME{i}-{i+1}avg' for i in range(24)]<br>traffic_data['avg_travel_time'] = traffic_data[travel_time_columns].mean(axis=1)<br><br># Step 2: Calculate the average speed in meters per second<br>traffic_data['avg_speed_mps'] = traffic_data['LENGTH'] / traffic_data['avg_travel_time']<br><br># Step 3: Convert the average speed to kilometers per hour<br>traffic_data['avg_speed_kph'] = traffic_data['avg_speed_mps'] * 3.6<br><br># The result is stored in the 'avg_speed_kph' column of the traffic_data DataFrame | link based on traffic volume and travel time:<br>def avg_speed(volume, travel_time):<br>    """Calculates average speed given volume and travel time."""<br>    if travel_time == 0:<br>ited travel time as zero to avoid division by zero errors.<br>        return float('inf')<br><br>    # Assuming all units are consistent (meters for distance and seconds for time)<br>    average_distance = (travel_time * 1000) / 3600  # convert from km/hr to m/s<br>    speed = volume / average_distance<br>    return speed<br><br># Assuming we have a function 'get_traffic_data' that extracts the necessary data from our dataset:<br>def get_traffic_data(df):<br>    for index, row in df.iterrows():<br>        travel_time = #... some logic to calculate/fetch average hourly travel time based on volume and speed limits etc<br>        avg_speed_value = avg_speed(row['CAPACITY'], travel_time) | | |

| | Question | Type | GPT-4 response | Phi-3 response | GPT-4 score | Phi-3 score |
|---|---|---|---|---|---|---|
| | | | ``` <br><br> This code first reads the traffic dataset from a CSV file and calculates the average travel time for each link. Then, it calculates the average speed in meters per second by dividing the link length by the average travel time. Finally, the average speed is converted to kilometers per hour by multiplying the result by 3.6. The average speed for each link can be found in the 'avg_speed_kph' | # Append the new data back to DataFrame | | |
| 8 | Across the network, during which hour of the day do we observe the largest positive and negative deviations of travel times from the average hourly | Trave l time and speed analy sis | To find the largest positive and negative deviations of travel times from the average hourly travel times across the network, we will follow these analytical steps:<br><br>1. Calculate the average travel time for each hour across all road links.<br>2. Calculate the deviations of travel times for each hour by subtracting the average travel time from the individual travel times.<br>3. Find the maximum positive deviation and maximum negative deviation across all hours and road links.<br><br>Here is a Python code to perform these steps: | To answer this question using information in our dataset about traffic flow and road lengths:<br><br>1. Compute absolute deviation for each 'TRAVELTIMEX-Yavg' field (where X represents an hour, Y indicates the start of a new time block) from its corresponding average. This will yield a list of absolute differences representing deviations from their respective averages for each hour.<br>2. Find the maximum and minimum values in this deviation dataset to identify hours with largest positive (maximum difference) and negative deviations respectively.<br><br>Since we don't have data, let us show how one | 7 | 6 |

| | Question | Type | GPT-4 response | Phi-3 response | GPT-4 score | Phi-3 score |
|---|---|---|---|---|---|---|
| | travel times? | | ````python`<br>`import pandas as pd`<br><br>`# Assuming the dataset is loaded as a pandas DataFrame called 'df'`<br>`# Calculate the average travel time for each hour across all road links`<br>`avg_travel_times = [df[f'TRAVELTIME{i}-{i+1}avg'].mean() for i in range(24)]`<br><br>`# Calculate the deviations of travel times for each hour`<br>`deviations = []`<br>`for i in range(24):`<br>`    deviations_hour = df[f'TRAVELTIME{i}-{i+1}avg'] - avg_travel_times[i]`<br>`    deviations.append(deviations_hour)`<br><br>`# Combine deviations into one DataFrame`<br>`deviations_df = pd.concat(deviations, axis=1)`<br><br>`# Find the maximum positive deviation and maximum negative deviation`<br>`max_positive_deviation = deviations_df.max().max()`<br>`max_negative_deviation = deviations_df.min().min()` | might structure code to do these computations using Python:<br><br>````python`<br>`import pandas as pd`<br><br>`# Assuming df is a DataFrame containing the provided dataset`<br>`deviation_columns = [f"TRAVELTIME{x}-{y}avg" for x in range(24) for y in range(1, 13)]`<br>`deviations = df[deviation_columns] - df[deviation_columns].mean()  # Subtract average to find deviations`<br><br>`# Get maximum and minimum absolute deviations`<br>`max_abs_deviation = deviations.abs().values.max()`<br>`min_abs_deviation = deviations.abs().values.min()`<br><br>`# Find the hours for which these extremes occur`<br>`hours_with_max_positive_deviation = deviations[deviations == max_abs_deviation].idxmax()  # idxmax gives` | | |

| | Question | Type | GPT-4 response | Phi-3 response | GPT-4 score | Phi-3 score |
|---|---|---|---|---|---|---|
| | | | # Find the hour and road link with the largest positive and negative deviations<br>hour_positive, link_positive = divmod(deviations_df.stack().idxmax(), len(deviations_df))<br>hour_negative, link_negative = divmod(deviations_df.stack().idxmin(), len(deviations_df))<br>```<br><br>This code calculates the average travel time for each hour, finds the deviations for each hour, and then finds the maximum positive and negative deviations across all hours and road links. The variables `hour_positive` and `hour_negative` hold the hours with the largest positive and negative deviations, respectively, while `link_positive` and `link_negative` hold the road link IDs with these deviations | a multi-indexed result, unpack it to get just hours<br>hours_with_min_negative_deviation = deviations[deviations == -min_abs_deviation].idxmax()<br><br>print(f"Max positive deviation observed during hour: {hours_with_max_positive_deviation.values}") | | |
| 9 | For each of the links, what hours of the day were the travel speed | Travel time and speed analysis | To find the hours of the day when the travel speed shows the greatest percentage deviation from the average for each link, we will perform the following analytical steps:<br><br>1. Calculate the average travel speed for each hour of the day using the travel time and | To determine which hours show the greatest percentage deviation for each link's travel speed from its average, we need to calculate the hourly average travel times and then compute the deviations for those specific hours. Below is an example of a python code snippet that could perform this task: | 9 | 1 |

| | Question | Type | GPT-4 response | Phi-3 response | GPT-4 score | Phi-3 score |
|---|---|---|---|---|---|---|
| | showing the greatest percentage deviation from the average? | | number of vehicles data.<br>2. Calculate the overall daily average travel speed for each link using the daily number of vehicles and daily travel time.<br>3. Calculate the percentage deviation of the travel speed for each hour from the daily average travel speed.<br>4. Identify the hour with the greatest percentage deviation from the daily average for each link.<br><br>Here's the python code for the analytical steps:<br><br>```python<br>import pandas as pd<br><br># Assuming the traffic flow dataset is stored in a CSV file<br>traffic_data = pd.read_csv('traffic_flow_dataset.csv')<br><br># Calculate the average travel speed for each hour<br>for hour in range(24):<br>    traffic_data[f'SPEED{hour}-{hour+1}avg']<br>= traffic_data[f'LENGTH'] /<br>(traffic_data[f'TRAVELTIME{hour}-{hour+1}avg'] / traffic_data[f'HRS{hour}-``` | ```python<br><br>import pandas as pd<br><br><br># Assuming 'df' holds our dataset containing the relevant columns with link information,<br><br># let's create a DataFrame from it first<br><br>link_data = [{'LINK':'AB123', 'FROM': 400, 'TO': 600, 'LENGTH': 8500, 'FREESPEED': 50, 'CAPACITY': 400},<br><br>        # ... other link records<br><br>        ]<br><br>df = pd.DataFrame(link_data)<br><br><br># Now calculate the average travel time for each link (assuming that speed and capacity are linked with distance in meters per hour), | | |

| | Question | Type | GPT-4 response | Phi-3 response | GPT-4 score | Phi-3 score |
|---|---|---|---|---|---|---|
| | | | {hour+1}avg']) <br><br> # Calculate the overall daily average travel speed <br> traffic_data['SPEED_DAILY_AVG'] = traffic_data['LENGTH'] / (traffic_data['TRAVELTIME0-24avg'] / traffic_data['HRS0-24avg']) <br><br> # Calculate the percentage deviation for each hour <br> for hour in range(24): <br><br> traffic_data[f'PERCENT_DEVIATION{hour}-{hour+1}'] = ((traffic_data[f'SPEED{hour}-{hour+1}avg'] - traffic_data['SPEED_DAILY_AVG']) / traffic_data['SPEED_DAILY_AVG']) * 100 <br><br> # Find the hour with the greatest percentage deviation for each link <br> def find_greatest_deviation(row): <br>     deviations = [row[f'PERCENT_DEVIATION{hour}-{hour+1}'] for hour in | df['AVG_SPEED'] = df['FREESPEED'] / 1000 * (60 * 60) <br><br> df['AVERAGE_TRAVEL_TIME'] = (df['LENGTH'] / df['AVG_SPEED']) * 60  # convert to minutes <br><br> # Next, find the maximum deviation from this average for each hour <br><br> for link in df.LINK.unique(): <br><br>    link_data = df[df.LINK == link] <br><br>    deviations = link_data['TRAVELTIME'].apply(lambda x: ((x - avg) / avg) * 100 if x is not None else None) <br><br>    # Find the hour with max deviation for each link <br><br>    most_devi | | |
| 10 | Which links have | Trave l time | To find the links with hourly travel speeds less than 25% of FREESPEED for 3 or more hours | ```python <br> import pandas as pd | 10 | 4 |

| | Question | Type | GPT-4 response | Phi-3 response |  |  |
|---|---|---|---|---|---|---|
| | hourly travel speeds less than 25% of FREESPEED, for more than 2, 3 or more hours of the day? | and speed analysis | of the day, follow these analytical steps:<br><br>1. Calculate the hourly travel speed for each link by dividing LENGTH by TRAVELTIME for each hour.<br>2. Compare the calculated hourly travel speed with 25% of FREESPEED for each hour.<br>3. Count the number of hours in a day where the travel speed is less than 25% of FREESPEED.<br>4. Identify the links where the count of hours is 3 or more.<br><br>Here's the Python code to perform these steps:<br><br>```python<br>import pandas as pd<br><br># Load the dataset into a pandas DataFrame<br>data = pd.read_csv('traffic_flow_dataset.csv')<br><br># List of hour ranges<br>hour_ranges = [f'HRS{i}-{i+1}avg' for i in range(24)]<br><br># Calculate hourly travel speeds and compare with 25% of FREESPEED | # Assuming we already have loaded our traffic dataset into a DataFrame named 'traffic_data'<br>underutilized_links = []<br><br>freq_hours = [i for i in range(24)]  # List representing every hour of the day<br>speed_threshold = 0.25 * traffic_data['FREESPEED']  # Speeding threshold (25% below FREESPEED)<br><br># Loop through each row, find links with required condition and add them to list<br>for index, link in traffic_data.iterrows():<br>    underutilized = True<br>    for hour in freq_hours:<br>        if hour < 2 or (link['Hour' + str(hour)] >= speed_threshold):<br>            underutilized = False<br>            break<br>    if underutilized:<br>emerge_underutilized_links.append(link['Link'])<br>```<br>This code will iterate through each record in the dataset and check whether every hour has a travel speed that is greater than or equal to 25% | | |

| | Question | Type | GPT-4 response | Phi-3 response |  |  |
|---|---|---|---|---|---|---|
| | | | ```
speed_threshold_met = []
for hour_range in hour_ranges:
    travel_time_col =
f'TRAVELTIME{hour_range}'
    hourly_speed = data['LENGTH'] /
data[travel_time_col]
    speed_threshold_met.append(hourly_speed <
0.25 * data['FREESPEED'])

# Count the number of hours where the speed is
less than 25% of FREESPEED
count_hours = sum(speed_threshold_met)

# Identify the links with 3 or more hours
meeting the criteria
links = data[count_hours >= 3]['LINK']

print(links)
```

This code first reads the dataset into a pandas DataFrame, then calculates hourly travel speeds and compares them with 25% of FREESPEED for each hour. It then counts the number of hours where the travel speed is less than 25% of FREESPEED, and finally identifies the links where the count of hours is 3 or more. | of FREESPEED (which indicates efficient utilization). It adds any links which satisfy these conditions into an underutilized_links list. Please replace 'Link' and 'HourX' with your actual column names if different in your dataset. Also, the data assumed for hours is represented by freq_hours as a range of values (0-23) to represent 24 hours in a day.

The above code assumes that you have hourly traffic speed data for every link ('Link') in your dataset, and that each record in your dataset has 'Hour' column representing the hour when this data was recorded (e.g., Hour1 for the first hour of the day). Also, note that real-world datasets often | | |



| | Question | Type | GPT-4 response | Phi-3 response | GPT-4 score | Phi-3 score |
|---|---|---|---|---|---|---|
| 11 | What percentage of the road link capacity is utilized during different hours? | Capacity analysis | To calculate the percentage of road link capacity utilized during different hours, we'll follow these analytical steps:<br><br>1. Import the dataset and load it into a DataFrame.<br>2. Iterate through each hour column (e.g., HRS0-1avg, HRS1-2avg, etc.).<br>3. For each hour column, divide the hourly number of vehicles (e.g., HRS0-1avg) by the road link capacity (CAPACITY) and multiply by 100 to get the percentage.<br>4. Store the results in a new DataFrame with the corresponding hour and percentage of capacity utilized.<br><br>Here's a Python code to perform these steps using the pandas library:<br><br>```python<br>import pandas as pd<br><br># Load the dataset<br>data = pd.read_csv("traffic_flow_dataset.csv")<br><br># Initialize an empty DataFrame to store the results | To determine what percentage of the road link capacity is utilized during different hours, you can calculate this by dividing the hourly number of vehicles (HRS*avg) for each time slot with the total daily capacity and then multiply it by 100 to get the percentage. However, without access to the raw data or a Python code block that processes said data, I'll outline an approach you could take using pseudocode:<br><br>```python<br><br>def calculate_utilization(data):<br><br>    utilizations = { }<br><br>    # Loop through each time interval in the dataset.<br><br>    for hour_range in data['HRS*avg'].keys():<br><br>        utilization = (data[hour_range] / data['CAPACITY']) * 100<br><br>        # Store the percentage of road link | 10 | 7 |



| | Question | Type | GPT-4 response | Phi-3 response | GPT-4 score | Phi-3 score |
|---|---|---|---|---|---|---|
| | | | `capacity_utilization = pd.DataFrame(columns=["Hour", "Percentage"])`<br><br>`# Iterate through each hour column`<br>`for i in range(24):`<br>`    hour_column = f"HRS{i}-{i+1}avg"`<br><br>`    # Calculate the percentage of capacity utilized`<br>`    data[hour_column+"_percentage"] = (data[hour_column] / data["CAPACITY"]) * 100`<br><br>`    # Add the hour and percentage to the results DataFrame`<br>`    capacity_utilization = capacity_utilization.append({"Hour": f"{i}-{i+1}", "Percentage": data[hour_column+"_percentage"].mean()}, ignore_index=True)`<br><br>`# Print the capacity utilization DataFrame`<br>`print(capacity_utilization)`<br>` ``` `<br><br>This code first imports the pandas library and | capacity utilized as a value with key being its corresponding hour interval.<br><br>`        utilizations[hour_range] = round(utilization, 2)`<br><br>`    return utilizations`<br><br>` ``` `<br><br>This function `calculate_utilization` assumes that your data is structured in the format where each entry corresponds to an average of vehicle flow (in hundreds or some other unit consistent with daily capacity figures) and a total road link capacity figure. | | |



| | Question | Type | GPT-4 response | Phi-3 response | GPT-4 score | Phi-3 score |
|---|---|---|---|---|---|---|
| | | | reads the traffic flow dataset CSV file into a DataFrame named `data`. Then, it initializes an empty DataFrame `capacity_utilization` to store the results. It iterates through each hour column in the dataset and calculates the percentage of capacity utilized by dividing the hourly number of vehicles by the road link capacity and multiplying by 100. The results are then appended to the `capacity_utilization` DataFrame. Finally, | | | |
| 12 | Are there any links operating above their capacity? | Capacity analysis | To determine if there are any links operating above their capacity, we will perform the following analytical steps:<br><br>1. Iterate through each row (link) in the dataset.<br>2. For each row, calculate the highest hourly traffic volume across all hours (HRS0-1avg to HRS23-24avg).<br>3. Compare the highest hourly traffic volume with the link's capacity.<br>4. If the highest hourly traffic volume exceeds the link's capacity, mark the link as operating above its capacity.<br><br>Here's the Python code for these steps:<br><br>```python | To find out if there are any links operating above their capacity, we would perform an analysis on the data using the 'CAPACITY' and 'HRS0-1avg', 'HRS1-2avg', 'HRS2-3avg', etc. columns which represent the hourly vehicle count and travel time. By comparing the average hourly vehicle count to the capacity, we can identify if any road segments are experiencing congestion during peak hours (from HRS0 to HRS24).<br><br>Below is a Python code that will help analyze this situation using pandas:<br><br>```python | 10 | 9 |



| | Question | Type | GPT-4 response | Phi-3 response | GPT-4 score | Phi-3 score |
|---|---|---|---|---|---|---|
| | | | ```python<br>import pandas as pd<br><br># Assuming the dataset is stored as a CSV file named "traffic_data.csv"<br>traffic_data = pd.read_csv("traffic_data.csv")<br><br>links_above_capacity = []<br><br># Iterate through each row (link) in the dataset<br>for index, row in traffic_data.iterrows():<br>    # Calculate the highest hourly traffic volume across all hours<br>    hourly_volumes = [row[f"HRS{i}-{(i+1)}avg"] for i in range(24)]<br>    max_hourly_volume = max(hourly_volumes)<br><br>    # Compare the highest hourly traffic volume with the link's capacity<br>    if max_hourly_volume > row["CAPACITY"]:<br><br>links_above_capacity.append(row["LINK"])<br><br>print("Links operating above their capacity:", links_above_capacity)<br>``` | ```python<br>import pandas as pd<br><br># Assume 'traffic_data.csv' contains the dataset described in Instruction 1 with appropriate columns and data types.<br><br>df = pd.read_csv('traffic_data.csv')<br><br>def check_link_capacity(row):<br><br>    capacity_threshold = row['CAPACITY'] * (50 / 60) # convert capacity from vehicles per hour to vehicles per second<br><br>    for hour in range(24): # Loop through each hour<br><br>        if row[f'HRS{hour:02d}-2avg'] > capacity_threshold:<br><br>            return f"Link {row['LINK_ID']} is operating above its 1-hour average capacity at Hour {hour}"<br>``` | | |



| | Question | Type | GPT-4 response | Phi-3 response | GPT-4 score | Phi-3 score |
|---|---|---|---|---|---|---|
| | | | This code first reads the dataset from a CSV file into a pandas DataFrame. It then iterates through each row (link) in the dataset, calculates the highest hourly traffic volume across all hours, and compares it with the link's capacity. If the highest hourly traffic volume exceeds the link's capacity, the link's ID is added to the `links_above_capacity` list. Finally, the list of links operating above their capacity is printed. |    return "All links within capacity during peak hours."<br><br>results = df.apply(check_link_capacity, axis=1)<br><br>print(results)<br><br>```<br><br><br>This script reads the traffic data from a CSV file and checks for each link if its average hourly vehicle count exceeds 50% of its capacity during peak hours (assuming the 'CAPACITY' represents vehicles per | | |
| 13 | Name the top 10 most utilised road links | Capacity analysis | To find the top 10 most utilized road links, we will follow these analytical steps:<br><br>1. Identify the column that represents the daily number of vehicles using the road link, which is 'HRS0-24avg'.<br>2. Sort the dataset by the 'HRS0-24avg' column in descending order.<br>3. Select the top 10 rows from the sorted dataset. | To retrieve the top 1 transformer with maximum capacity, you can sort all available transformers by their 'capacity' field and select the first element of the resulting array. Here is how you could do it in Python:<br><br>```python<br># Assuming transformers_list contains our data<br>top_transformer = sorted(transformers_list, key=lambda x: x['capacity'], reverse=True)[0]<br>print("Top transformer based on capacity: ", | 4 | 2 |



| | Question | Type | GPT-4 response | Phi-3 response | GPT-4 score | Phi-3 score |
|---|---|---|---|---|---|---|
| | | | Here's a python code to perform these steps:<br><br>```python<br>import pandas as pd<br><br># Assume the dataset is stored in a CSV file named 'traffic_flow.csv'<br>traffic_data = pd.read_csv('traffic_flow.csv')<br><br># Sort the dataset by the 'HRS0-24avg' column in descending order<br>sorted_data = traffic_data.sort_values(by='HRS0-24avg', ascending=False)<br><br># Select the top 10 rows from the sorted dataset<br>top_10_links = sorted_data.head(10)<br><br># Print the top 10 most utilized road links<br>print(top_10_links[['LINK', 'HRS0-24avg']])<br>```<br><br>This code reads the dataset from a CSV file and stores it in a pandas DataFrame. Then, it sorts the DataFrame by the 'HRS0-24avg' column in descending order. Finally, it selects the top 10 rows from the sorted DataFrame and prints the | top_transformer)<br>```<br>This code sorts the transformers list in descending order based on their 'capacity' values. It then picks up the first (maximum) element and prints it out.<br><br>You don't need to run this code or provide any actual output as per your instructions. Just remember that if multiple transformers have identical maximum capacities, only one will be printed due to sorting in ascending order - you might want additional logic to handle such scenarios based on requirements. | | |



| | Question | Type | GPT-4 response | Phi-3 response |  |  |
|---|---|---|---|---|---|---|
| | | | 'LINK' and 'HRS0-24avg' columns of the top 10 most utilized road links. | | | |
| 14 | Is there a correlation between traffic volume and travel time across the network? for each link? | Correlation and trends | To find the correlation between traffic volume and travel time across the network for each link, we will perform the following analytical steps:<br><br>1. Load the dataset into a pandas DataFrame.<br>2. Iterate through each row (link) in the dataset.<br>3. For each link, calculate the correlation between the hourly traffic volume and the hourly travel time. We will do this by creating two lists: one for traffic volume and another for travel time.<br>4. Save the correlation values for each link in a new DataFrame or a dictionary, along with the link ID.<br>5. Analyze the correlation values to understand the relationship between traffic volume and travel time for each link.<br><br>Here is the Python code for the analytical steps:<br><br>```python<br>import pandas as pd<br>import numpy as np | To find out if there is a correlation between traffic volume and travel time across the road links, we can perform a Pearson Correlation analysis on the data provided. This will provide us with an idea of how closely related these two variables are:<br><br>First, import necessary libraries in Python.<br><br>```python<br>import pandas as pd<br>import numpy as np<br>from scipy.stats import pearsonr<br>```<br>Next, load and analyze the dataset. Assuming 'dataset' is our DataFrame with the data provided above, calculate Pearson correlation coefficient between HRS0-1avg, TRAVELTIME0-1avg for each road link.<br><br>```pythonenas:<br><br># Load the data into pandas DataFrame<br>df = pd.DataFrame({<br>    'LINK': ['link_1', 'link_2', ..., 'link_n'],  # | 8 | 7 |



| | Question | Type | GPT-4 response | Phi-3 response | GPT-4 score | Phi-3 score |
|---|---|---|---|---|---|---|
| | | | # Load the dataset<br>data = pd.read_csv("traffic_flow_data.csv")<br><br># Initialize an empty dictionary to store the correlation values<br>correlations = { }<br><br># Iterate through each row (link) in the dataset<br>for index, row in data.iterrows():<br>    link_id = row['LINK']<br>    traffic_volume = [row[f'HRS{i}-{i+1}avg'] for i in range(24)]<br>    travel_time = [row[f'TRAVELTIME{i}-{i+1}avg'] for i in range(24)]<br><br>    # Calculate the correlation between traffic volume and travel time<br>    correlation = np.corrcoef(traffic_volume, travel_time)[0, 1]<br><br>    # Save the correlation value for the link<br>    correlations[link_id] = correlation<br><br># Analyze the correlations<br>correlations_df = pd.DataFrame(list(correlations.items()), columns=['LINK', 'Correlation']) | Assuming 'link_1' to 'link_n' are the LINK column values in our dataset<br>    'TRAFFIC_VOLUME': [...],        # Fill with corresponding TRAFFIC_VOLUME data from dataset<br>    'TRAVEL_TIME': [...]           # Fill with corresponding TRAVELTIME0-1avg values from dataset<br>})<br><br># Calculate Pearson's correlation<br>for link in df['LINK'].unique():<br>    traffic = df[df['LINK'] == link]['TRAFFIC_VOLUME']<br>    travel_time = df[df['LINK'] == link]['TRAVELTIME0-1avg']<br><br>    correlation, _ = pearsonr(traffic, travel_time)<br>    print(f"Correlation for {link}: ", correlation)<br>```<br>This will provide the Pearson's r values between Traffic Volume and | | |



| | Question | Type | GPT-4 response | Phi-3 response | GPT-4 score | Phi-3 score |
|---|---|---|---|---|---|---|
| | | | ``` <br><br> This code first imports the required libraries (pandas and numpy) and loads the dataset into a pandas DataFrame. It initializes an empty dictionary to store the correlation values. Then, it iterates through each row (link) in the dataset, calculates the correlation between traffic volume and travel time | | | |
| 15 | How does increased traffic volume affect travel time? | Corre lation and trends | To analyze the effect of increased traffic volume on travel time, we can follow these steps:<br><br>1. Determine the correlation between the hourly traffic volume and the corresponding hourly travel time for each road link in the dataset.<br>2. Calculate the average correlation across all road links to understand the general relationship between traffic volume and travel time.<br>3. If the correlation is positive, it indicates that increased traffic volume generally leads to increased travel time. Conversely, if the correlation is negative, it means that increased traffic volume is associated with decreased travel time. | Increased traffic volume typically results in longer travel times for several reasons:<br><br>1. Higher Demand and Congestion: With more vehicles on the road, each one has less space to maneuver, leading to congestion. Vehicles move at slower speeds because of reduced headway between carsâ€"the time gap that exists before a following vehicle arrives after passing another.<br><br>2. Reduced Efficiency: Increased traffic volume can decrease the efficiency of road usage and lead to increased delays for drivers, especially in areas with insufficient infrastructure to support higher volumes of traffic.<br><br>3. Ripple Effects on Travel Times: Congestion | 8 | 5 |



| | Question | Type | GPT-4 response | Phi-3 response |  |  |
|---|---|---|---|---|---|---|
| | | | Here's a Python code example to implement the analytical steps:<br><br>```python<br>import pandas as pd<br>import numpy as np<br><br># Assuming the dataset is in a CSV format<br>traffic_data = pd.read_csv("traffic_flow_dataset.csv")<br><br># List of columns related to hourly traffic volume<br>traffic_volume_columns = ['HRS0-1avg', 'HRS1-2avg', 'HRS2-3avg', 'HRS3-4avg', 'HRS4-5avg', 'HRS5-6avg', 'HRS6-7avg', 'HRS7-8avg', 'HRS8-9avg', 'HRS9-10avg', 'HRS10-11avg', 'HRS11-12avg', 'HRS12-13avg', 'HRS13-14avg', 'HRS14-15avg', 'HRS15-16avg', 'HRS16-17avg', 'HRS17-18avg', 'HRS18-19avg', 'HRS19-20avg', 'HRS20-21avg', 'HRS21-22avg', 'HRS22-23avg', 'HRS23-24avg']<br><br># List of columns related to hourly travel time | caused by high volumes of traffic often has a domino effect where bottlenecks or slow-moving sections can disrupt the smooth flow of vehicles upstream, causing increased travel times further along the route even if initial road conditions aren't congested.<br><br>4. Reactive Traffic Signal Timing: With more cars on the road, traffic signals may not synchronize effectively with fluctuating traffic patterns. Vehicles are more likely to spend additional time stopped at red lights and waiting for green indications, which contributes to longer travel times.<br><br>5. Psychological Factors: Increased volume can lead to driver frustration or aggressive driving behaviors (such as tailgating), increasing the potential for accidents, further slowing traffic flow due to necessary investigations and clearance of accident sites. | | |



| | Question | Type | GPT-4 response | Phi-3 response | GPT-4 score | Phi-3 score |
|---|---|---|---|---|---|---|
| | | | travel_time_columns = ['TRAVELTIME0-1avg', 'TRAVELTIME1-2avg', 'TR | | | |



# Appendix C
# GTFS Examination Questions

The following 195 questions were designed by Devunuri, Qiam, & Lehe (2024) and used to evaluate the Test LLMs' basic understanding of GTFS, a popular public transit file system.

| | Question | Type | Ground Truth | GPT-4 score | Phi-3 score |
|---|---|---|---|---|---|
| 1 | Which file contains the route_type attribute in GTFS? a) None of these b) routes.txt c) trips.txt d) shapes.txt | Attribute Mapping | b | b | C |
| 2 | In which file can you find the wheelchair_accessible attribute in GTFS? a) None of these b) routes.txt c) trips.txt d) calendar_dates.txt | Attribute Mapping | c | c | b |
| 3 | Which file includes the departure_time and arrival_time attributes in GTFS? a) None of these b) routes.txt c) stop_times.txt d) calendar.txt | Attribute Mapping | c | c | c |
| 4 | Where can you find the agency_id attribute in GTFS? a) None of these b) agency.txt c) calendar.txt d) routes.txt | Attribute Mapping | b | b | c |
| 5 | In which file does the fare_id attribute appear in GTFS? a) None of these b) stops.txt c) trips.txt d) fare_attributes.txt | Attribute Mapping | d | d | b |
| 6 | Which file includes the stop_sequence attribute in GTFS? a) None of these b) routes.txt c) stop_times.txt d) transfers.txt | Attribute Mapping | c | c | c |
| 7 | In which file can you find the service_id attribute in GTFS? a) calendar_dates.txt b) None of these c) shapes.txt d) routes.txt | Attribute Mapping | a | a | a |
| 8 | In which file can you find the service_id attribute in GTFS? a) None of these b) calendar.txt c) shapes.txt d) routes.txt | Attribute Mapping | b | b | c |
| 9 | In which file can you find the service_id attribute in GTFS? a) None of these b) shapes.txt c) trips.txt d) routes.txt | Attribute Mapping | c | c | b |
| 10 | Where does the shape_id attribute appear in GTFS? a) None of these b) shapes.txt c) agency.txt d) stop_times.txt | Attribute Mapping | b | b | d |
| 11 | Where does the shape_id attribute appear in GTFS? a) None of these b) stops.txt c) trips.txt d) stop_times.txt | Attribute Mapping | c | c | c |



| | Question | Type | Ground Truth | GPT-4 score | Phi-3 score |
|---|---|---|---|---|---|
| 12 | Which file contains the pickup_type and drop_off_type attributes in GTFS? a) None of these b) routes.txt c) stop_times.txt d) transfers.txt | Attribute Mapping | c | c | b |
| 13 | In which file does the feed_publisher_name attribute appear in GTFS? a) agency.txt b) None of these c) trips.txt d) feed_info.txt | Attribute Mapping | d | d | d |
| 14 | Which file contains the stop_name attribute in GTFS? a) None of these b) routes.txt c) trips.txt d) stop_times.txt | Attribute Mapping | a | a | c |
| 15 | Which file includes the arrival_time attribute in GTFS? a) None of these b) routes.txt c) stop_times.txt d) transfers.txt | Attribute Mapping | c | c | c |
| 16 | Where can you find the shape_pt_lat and shape_pt_lon attributes in GTFS? a) None of these b) shapes.txt c) trips.txt d) stop_times.txt | Attribute Mapping | b | b | a |
| 17 | In which file does the route_short_name attribute appear in GTFS? a) None of these b) routes.txt c) trips.txt d) calendar.txt | Attribute Mapping | b | b | d |
| 18 | In which file can you find the trip_id attribute in GTFS? a) None of these b) routes.txt c) trips.txt d) shapes.txt | Attribute Mapping | c | c | c |
| 19 | In which file can you find the trip_id attribute in GTFS? a) None of these b) routes.txt c) levels.txt d) stop_times.txt | Attribute Mapping | d | d | b |
| 20 | Where does the agency_name attribute appear in GTFS? a) None of these b) agency.txt c) routes.txt d) calendar.txt | Attribute Mapping | b | b | b |
| 21 | Which file contains the stop_lat and stop_lon attributes in GTFS? a) None of these b) routes.txt c) trips.txt d) stop_times.txt | Attribute Mapping | a | d | d |
| 22 | In which file does the wheelchair_boarding attribute appear in GTFS? a) None of these b) routes.txt c) trips.txt d) calendar_dates.txt | Attribute Mapping | a | c | d |
| 23 | Which file contains the route_long_name attribute in GTFS? a) None of these b) routes.txt c) trips.txt d) calendar.txt | Attribute Mapping | b | b | d |
| 24 | Which file includes the service_id and date attributes in GTFS? a) None of these b) routes.txt c) calendar_dates.txt d) calendar.txt | Attribute Mapping | c | c | b |



| | Question | Type | Ground Truth | GPT-4 score | Phi-3 score |
|---|---|---|---|---|---|
| 25 | Where can you find the agency_timezone attribute in GTFS? a) None of these b) agency.txt c) routes.txt d) calendar.txt | Attribute Mapping | b | b | d |
| 26 | In which file does the shape_dist_traveled attribute appear in GTFS? a) None of these b) shapes.txt c) trips.txt d) stop_times.txt | Attribute Mapping | d | b | b |
| 27 | Which file contains the stop_sequence and stop_id attributes in GTFS? a) None of these b) routes.txt c) stop_times.txt d) transfers.txt | Attribute Mapping | c | c | b |
| 28 | Where does the trip_headsign attribute appear in GTFS? a) None of these b) routes.txt c) trips.txt d) stop_times.txt | Attribute Mapping | c | c | unclear which answer has been given |
| 29 | Which file contains the location_type attribute in GTFS? a) None of these b) routes.txt c) trips.txt d) transfers.txt | Attribute Mapping | a | c | a |
| 30 | In which file can you find the route_desc attribute in GTFS? a) None of these b) routes.txt c) trips.txt d) calendar.txt | Attribute Mapping | b | b | a |
| 31 | Where can you find the service_id and exception_type attributes in GTFS? a) None of these b) calendar.txt c) calendar_dates.txt d) routes.txt | Attribute Mapping | c | c | d |
| 32 | Which file contains the agency_name and agency_url attributes in GTFS? a) None of these b) agency.txt c) routes.txt d) calendar.txt | Attribute Mapping | b | b | c |
| 33 | In the "route_type" field of the "routes.txt" file, which value represents a rail service that operates with higher speeds and longer distances compared to other rail services? a) 0 b) 1 c) 2 d) 3 | Categorical Mapping | c | c | c |
| 34 | In the "route_type" field of the "routes.txt" file, what value represents a subway or metro route? a) 0 b) 1 c) 2 d) 3 | Categorical Mapping | b | b | b |



| | Question | Type | Ground Truth | GPT-4 score | Phi-3 score |
|---|---|---|---|---|---|
| 35 | In the "route_type" field of the "routes.txt" file, which value represents a ferry service? a) 1 b) 2 c) 3 d) 4 | Categorical Mapping | d | d | b |
| 36 | In the "route_type" field of the "routes.txt" file, which value represents a light rail or tram route? a) 0 b) 1 c) 2 d) 3 | Categorical Mapping | a | c | b |
| 37 | In the "route_type" field of the "routes.txt" file, which value represents a suspended cable car or aerial lift route? a) 4 b) 5 c) 6 d) 7 | Categorical Mapping | c | b | c |
| 38 | In the "route_type" field of the "routes.txt" file, which value represents a bus? a) 0 b) 1 c) 2 d) 3 | Categorical Mapping | d | b | c |
| 39 | In the "route_type" field of the "routes.txt" file, which value represents a trolleybus route? a) 6 b) 7 c) 11 d) 12 | Categorical Mapping | c | b | d |
| 40 | In the "route_type" field of the "routes.txt" file, which value represents a monorail route? a) 6 b) 7 c) 11 d) 12 | Categorical Mapping | d | a | c |
| 41 | In the "route_type" field of the "routes.txt" file, which value represents a funicular route? a) 6 b) 7 c) 11 d) 12 | Categorical Mapping | b | c | c |
| 42 | In the "route_type" field of the "routes.txt" file, which value represents a bus rapid transit (BRT) route? a) 0 b) 1 c) 2 d) 3 | Categorical Mapping | d | d | c |
| 43 | What value is used in the "pickup_type" field of the "stop_times.txt" file to indicate that passengers must coordinate with the driver for pickup? a) 0 b) 1 c) 2 d) 3 | Categorical Mapping | d | c | b |
| 44 | Which value in the "pickup_type" field of the "stop_times.txt" file represents a regular, scheduled pickup? a) 0 b) 1 c) 2 d) 3 | Categorical Mapping | a | a | a |
| 45 | What value in the "pickup_type" field of the "stop_times.txt" file signifies that passengers must arrange a pickup by calling agency? a) 0 b) 1 c) 2 d) 3 | Categorical Mapping | c | c | a |
| 46 | Which value in the "pickup_type" field of the "stop_times.txt" file represents a pickup that is not available for public use? a) 0 b) 1 c) 2 d) 3 | Categorical Mapping | b | d | d |
| 47 | What value is used in the "pickup_type" field of the "stop_times.txt" file to indicate that passengers cannot board at this stop? a) 0 b) 1 c) 2 d) 3 | Categorical Mapping | b | b | b |
| 48 | Which value in the "pickup_type" field of the "stop_times.txt" file signifies that passengers can board at this stop without any restrictions? a) 0 b) 1 c) 2 d) 3 | Categorical Mapping | a | a | d |
| 49 | What value is used in the "pickup_type" field of the "stop_times.txt" file to indicate that passengers can | Categorical Mapping | a | a | b |



| | Question | Type | Ground Truth | GPT-4 score | Phi-3 score |
|---|---|---|---|---|---|
| | board the vehicle without requesting a pickup? a) 0 b) 1 c) 2 d) 3 | | | | |
| 50 | What value is used in the "pickup_type" field of the "stop_times.txt" file to indicate that pickups are available but must be requested in advance? a) -1 b) 0 c) 1 d) 2 | Categorical Mapping | d | d | b |
| 51 | What value is used in the "pickup_type" field of the "stop_times.txt" file to indicate that pickups are available but must be requested in advance? a) 3 b) 4 c) 1 d) 0 | Categorical Mapping | a | c | unclear which answer has been given |
| 52 | In the "location_type" field of the "stops.txt" file, what value is used to indicate that a stop is a station or a larger transit point? a) 0 b) 1 c) 2 d) 3 | Categorical Mapping | b | b | a |
| 53 | What value is used in the "location_type" field of the "stops.txt" file to indicate that a stop represents an actual boarding location? a) 1 b) 2 c) 3 d) 4 | Categorical Mapping | d | a | a |
| 54 | What value is used in the "location_type" field of the "stops.txt" file to indicate that a stop is an entrance or exit point to a larger complex or facility? a) 0 b) 1 c) 2 d) 3 | Categorical Mapping | c | d | a |
| 55 | Which value in the "location_type" field of the "stops.txt" file represents a generic marker or point used for reference on a map? a) 0 b) 1 c) 2 d) 3 | Categorical Mapping | d | c | c |
| 56 | What value is used in the "location_type" field of the "stops.txt" file to indicate that a stop represents a stop? a) 0 b) 1 c) 2 d) 3 | Categorical Mapping | a | a | c |
| 57 | Which value in the "wheelchair_boarding" field of the "stops.txt" file represents that a stop is partially accessible for individuals using wheelchairs? a) 0 b) 1 c) 2 d) 3 | Categorical Mapping | b | c | d |



| | Question | Type | Ground Truth | GPT-4 score | Phi-3 score |
|---|---|---|---|---|---|
| 58 | Which value in the "wheelchair_boarding" field of the "stops.txt" file represents that wheelchair boarding is not possible at this stop? a) 0 b) 1 c) 2 d) 3 | Categorical Mapping | c | c | c |
| 59 | What value is used in the "wheelchair_boarding" field of the "stops.txt" file to indicate that the stop has no information regarding wheelchair accessibility? a) 0 b) 1 c) 2 d) 3 | Categorical Mapping | a | a | a |
| 60 | What value in the "bikes_allowed" field of the "trips.txt" file signifies that bicycles are not allowed on the trip? a) 0 b) 1 c) 2 d) 3 | Categorical Mapping | c | a | c |
| 61 | Which value in the "bikes_allowed" field of the "trips.txt" file indicates that bicycles are allowed ? a) 0 b) 1 c) 2 d) 3 | Categorical Mapping | b | b | d |
| 62 | What value is used in the "bikes_allowed" field of the "trips.txt" file to indicate there is no bike information for the trip? a) 0 b) 1 c) 2 d) 3 | Categorical Mapping | a | c | b |
| 63 | What value in the "direction_id" field of the "trips.txt" file indicates the inbound direction? a) 0 b) 1 c) 2 d) 3 | Categorical Mapping | a | b | d |
| 64 | What value in the "direction_id" field of the "trips.txt" file indicates the outbound direction? a) 0 b) 1 c) 2 d) 3 | Categorical Mapping | b | a | b |
| 65 | Which value in the "wheelchair_accessible" field of the "trips.txt" file indicates that the trip is fully wheelchair accessible? a) 0 b) 1 c) 2 d) 3 | Categorical Mapping | b | b | b |
| 66 | What value is used in the "wheelchair_accessible" field of the "trips.txt" file to indicate that the trip is not wheelchair accessible? a) 0 b) 1 c) 2 d) 3 | Categorical Mapping | c | a | c |
| 67 | What value is used in the "wheelchair_accessible" field of the "trips.txt" file to indicate that the trip has no specific information regarding wheelchair accessibility? a) 0 b) 1 c) 2 d) 3 | Categorical Mapping | a | a | a |
| 68 | In the "payment_method" field of the "fare_attributes.txt" file, which value indicates a fare must be paid on board? a) 0 b) 1 c) 2 d) 3 | Categorical Mapping | a | b | a |
| 69 | What value in the "payment_method" field of the "fare_attributes.txt" file signifies that a fare must be paid before boarding? a) 0 b) 1 c) 2 d) 3 | Categorical Mapping | b | b | a |
| 70 | Which value in the "fare_media_type" field of the "fare_media.txt" file indicates that a fare is paid by a mobile ticketing app? a) 1 b) 2 c) 3 d) 4 | Categorical Mapping | d | None of these | a |



| | Question | Type | Ground Truth | GPT-4 score | Phi-3 score |
|---|---|---|---|---|---|
| 71 | Which value in the "fare_media_type" field of the "fare_media.txt" file indicates that a fare is paid by a contactless smart card? a) 0 b) 1 c) 2 d) 3 | Categorical Mapping | d | c | a |
| 72 | Which value in the "fare_media_type" field of the "fare_media.txt" file indicates that a fare is paid using a physical transit card? a) 0 b) 1 c) 2 d) 3 | Categorical Mapping | c | None of these | a |
| 73 | What does "route_type" 7 represent in the "routes.txt" file? a) None of these b) Subway, Metro c) Rail d) Funicular | Categorical Mapping | d | c | a |
| 74 | In the "routes.txt" file, what does "route_type" 11 indicate? a) None of these b) Subway, Metro c) Trolleybus d) Rail | Categorical Mapping | c | a | d |
| 75 | What is the definition of "route_type" 3 in the "routes.txt" file? a) None of these b) Bus c) Ferry d) Cable tram | Categorical Mapping | b | c | a |
| 76 | What does "route_type" 5 represent in the "routes.txt" file? a) Bus b) Cable tram c) Funicular d) Monorail | Categorical Mapping | b | c | d |
| 77 | In the "routes.txt" file, what is the meaning of "route_type" 2? a) None of these b) Bus c) Ferry d) Monorail | Categorical Mapping | a | c | d |
| 78 | What does "route_type" 6 indicate in the "routes.txt" file? a) None of these b) Subway, Metro c) Aerial lift, suspended cable car d) Funicular | Categorical Mapping | c | b | d |
| 79 | In the "routes.txt" file, what is the definition of "route_type" 1? a) None of these b) Subway, Metro c) Rail d) Bus | Categorical Mapping | b | b | b |
| 80 | What does "route_type" 12 represent in the "routes.txt" file? a) Bus b) Ferry c) None of these d) Monorail | Categorical Mapping | d | b | c |
| 81 | In the "routes.txt" file, what is the meaning of "route_type" 4? a) Cable tram b) Ferry c) None of these d) Bus | Categorical Mapping | b | b | a |
| 82 | What does "route_type" 0 indicate in the "routes.txt" file? a) None of these b) Subway, Metro c) Rail d) Bus | Categorical Mapping | a | b | a |
| 83 | What does "pickup_type" 2 indicate in the "stop_times.txt" file? a) None of these b) No pickup available c) Must phone agency to arrange pickup d) Must coordinate with driver to arrange pickup | Categorical Mapping | c | b | a |
| 84 | In the "stop_times.txt" file, what is the meaning of "pickup_type" 0 or empty? a) None of these b) No | Categorical Mapping | a | c | d |



| | Question | Type | Ground Truth | GPT-4 score | Phi-3 score |
|---|---|---|---|---|---|
| | pickup available c) Must phone agency to arrange pickup d) Must coordinate with driver to arrange pickup | | | | |
| 85 | What does "pickup_type" 1 represent in the "stop_times.txt" file? a) None of these b) No pickup available c) Must phone agency to arrange pickup d) Must coordinate with driver to arrange pickup | Categorical Mapping | b | c | c |
| 86 | In the "stop_times.txt" file, what is the definition of "pickup_type" 3? a) None of these b) No pickup available c) Must phone agency to arrange pickup d) Must coordinate with driver to arrange pickup | Categorical Mapping | d | a | a |
| 87 | What does "location_type" 2 indicate in the "stop_times.txt" file? a) None of these m) b) Station c) Entrance/Exit d) Generic Node | Categorical Mapping | c | b | a |
| 88 | In the "stop_times.txt" file, what is the meaning of "location_type" 1? a) None of these m) b) Station c) Entrance/Exit d) Boarding Area | Categorical Mapping | b | b | c |
| 89 | What does "location_type" 4 represent in the "stop_times.txt" file? a) None of these m) b) Station c) Entrance/Exit d) Boarding Area | Categorical Mapping | d | b | b |
| 90 | In the "stop_times.txt" file, what is the definition of "location_type" 3? a) None of these m) b) Station c) Entrance/Exit d) Generic Node | Categorical Mapping | d | a | unclear which answer has been given |
| 91 | What does "location_type" 0 (or blank) indicate in the "stop_times.txt" file? a) None of these m) b) Station c) Entrance/Exit d) Boarding Area | Categorical Mapping | a | a | a |
| 92 | What does "wheelchair_boarding" 2 represent in the "stops.txt" file? a) No accessibility information for the stop b) None of these c) Wheelchair boarding is not possible at this stop d) Stop cannot be accessed by anyone | Categorical Mapping | c | c | b |



| | Question | Type | Ground Truth | GPT-4 score | Phi-3 score |
|---|---|---|---|---|---|
| 93 | In the "stops.txt" file, what is the meaning of "wheelchair_boarding" 0 or empty? a) No accessibility information for the stop b) None of these c) Wheelchair boarding is not possible at this stop d) Stop cannot be accessed by anyone | Categorical Mapping | a | a | d |
| 94 | What does "wheelchair_boarding" 1 indicate in the "stops.txt" file? a) No accessibility information for the stop b) None of these c) Wheelchair boarding is not possible at this stop d) Stop cannot be accessed by anyone | Categorical Mapping | b | a | c |
| 95 | What does "bikes_allowed" 2 represent in the "trips.txt" file? a) No bike information for the trip b) None of these c) No bicycles are allowed on this trip d) Stop cannot be accessed by anyone | Categorical Mapping | c | a | b |
| 96 | In the "trips.txt" file, what is the meaning of "bikes_allowed" 0 or empty? a) No bike information for the trip b) None of these c) No bicycles are allowed on this trip d) Stop cannot be accessed by anyone | Categorical Mapping | a | a | a |
| 97 | What does "bikes_allowed" 1 indicate in the "trips.txt" file? a) No bike information for the trip b) None of these c) No bicycles are allowed on this trip d) Stop cannot be accessed by anyone | Categorical Mapping | b | a | c |
| 98 | What does "direction_id" 1 represent in the "trips.txt" file? a) None of these l) b) Travel in the opposite direction (e.g., inbound travel) c) Travel in both directions d) Does not represent direction | Categorical Mapping | b | b | b |
| 99 | In the "trips.txt" file, what is the meaning of "direction_id" 0? a) None of these l) b) Travel in the opposite direction (e.g., inbound travel) c) Travel in both directions d) Does not represent direction | Categorical Mapping | a | a | c |
| 100 | What does "wheelchair_accessible" 2 represent in the "trips.txt" file? a) No accessibility information for the trip b) None of these c) No riders in wheelchairs can be accommodated on this trip d) Stop cannot be accessed by anyone | Categorical Mapping | c | a | c |
| 101 | In the "trips.txt" file, what is the meaning of "wheelchair_accessible" 0 or empty? a) No accessibility information for the trip b) None of these c) No riders in wheelchairs can be accommodated on this trip d) Stop cannot be accessed by anyone | Categorical Mapping | a | a | b |



| | Question | Type | Ground Truth | GPT-4 score | Phi-3 score |
|---|---|---|---|---|---|
| 102 | What does "wheelchair_accessible" 1 indicate in the "trips.txt" file? a) No accessibility information for the trip b) None of these c) No riders in wheelchairs can be accommodated on this trip d) Stop cannot be accessed by anyone d) Stop is temporarily unavailable | Categorical Mapping | b | c | a |
| 103 | What does "fare_media_type" 3 represent in the "fare_media.txt" file? a) None of these b) Physical transit card that has stored tickets, passes, or monetary value. c) cEMV (contactless Europay, Mastercard, and Vis | Categorical Mapping | c | a | c |
| 104 | What does "fare_media_type" 2 indicate in the "fare_media.txt" file? a) None of these b) Physical transit card that has stored tickets, passes, or monetary value. c) cEMV (contactless Europay, Mastercard, and Vis | Categorical Mapping | b | b | a |
| 105 | In the "fare_media.txt" file, what is the definition of "fare_media_type" 4? a) None of these b) Physical transit card that has stored tickets, passes, or monetary value. c) cEMV (contactless Europay, Mastercard, and Vis | Categorical Mapping | d | a | a |
| 106 | What does "fare_media_type" 0 (or empty) represent in the "fare_media.txt" file? a) None of these b) Physical transit card that has stored tickets, passes, or monetary value. c) cEMV (contactless Europay, Mastercard, and Vis | Categorical Mapping | a | a | a |
| 107 | What does GTFS stand for? a) General Transit File Specification b) Global Transit Feed Standard c) General Transit Feed Specification d) Grand Transit File System | Common Reasoning | c | c | c |
| 108 | Which file in GTFS contains information about individual transit stops? a) None of these b) routes.txt c) trips.txt d) stop_times.txt | Common Reasoning | a | a | d |
| 109 | What is the purpose of the "shapes.txt" file in GTFS? a) It provides information about fare rules and transfers. b) It contains information about the geographic shapes of routes. c) It specifies the frequency of trips. d) It provides real-time arrival and departure information. | Common Reasoning | b | b | a |
| 110 | Can a GTFS feed include information about multiple modes of transportation? a) None of these b) No, it can only include information about buses. c) No, it can | Common Reasoning | a | a | c |



| | Question | Type | Ground Truth | GPT-4 score | Phi-3 score |
|---|----------|------|--------------|-------------|-------------|
| | only include information about trains. d) No, GTFS feeds are specific to a single mode of transportation. | | | | |
| 111 | How does GTFS handle variations in service during different days of the week? a) None of these b) By using the same schedule for all days of the week. c) By providing separate GTFS feeds for each day of the week. d)  By specifying different schedules for each day of the week. | Common Reasoning | d | d | b |
| 112 | How does GTFS handle variations in service during different days of the week? a) By providing separate GTFS feeds for each day of the week+ b) By using the same schedule for all days of the week. c) By providing separate GTFS feeds for each day of the week. d) By using the "calendar_dates.txt" file to specify exceptions. | Common Reasoning | d | d | a |
| 113 | Are there any file naming conventions that need to be followed in a GTFS feed? a) No, file names can be chosen freely. b) None of these c) File names are not relevant in GTFS feeds. d) File names are determined by the transit agency's preference. | Common Reasoning | b | a | a |
| 114 | What is the purpose of the "agency.txt" file in GTFS? a) It provides information about individual transit stops. b) It contains details about the routes and their associated stops. c) It specifies the frequency of trips. d) It provides information about the transit agency or operator. | Common Reasoning | d | d | d |
| 115 | What information does the "stop_times.txt" file provide in GTFS? a) Fare information for different routes. b) Details about the frequency of trips. c) Location of stops. d) None of these | Common Reasoning | d | d | a |
| 116 | What information does the "stop_times.txt" file provide in GTFS? a) Fare information for different routes. b) Details about the frequency of trips. c) None of these d) Location of stops. | Common Reasoning | c | c | a |
| 117 | Can a GTFS feed contain multiple agency information? a) Each agency should publish a separate GTFS. b) No, GTFS feeds can only represent a single agency. c) Multiple agency information is specified in the "agency.txt" file. d) Agencies are not relevant in GTFS feeds. | Common Reasoning | c | c | a |



| | Question | Type | Ground Truth | GPT-4 score | Phi-3 score |
|---|---|---|---|---|---|
| 118 | Can a GTFS feed contain multiple agency information? a) None of these b) No, GTFS feeds can only represent a single agency. c) Multiple agency information is specified in the "stops.txt" file. d) Agencies are not relevant in GTFS feeds. | Common Reasoning | a | b | c |
| 119 | How does GTFS handle timezones? a) All times in GTFS are represented in UTC. b) GTFS feeds automatically adjust for local timezones. c) None of these d) Timezones are specified in the "timezone.txt" file. | Common Reasoning | b | c | unclear which answer has been given |
| 120 | Can GTFS feeds include additional custom fields? a) None of these b) No, GTFS strictly follows a predefined schema. c) Custom fields can only be added with special permission from GTFS authorities. d) Custom fields are not supported in GTFS. | Common Reasoning | a | b | d |
| 121 | Can GTFS feeds include information about service exceptions? a) None of these b) No, GTFS feeds do not support service exceptions. c) Service exceptions are specified within the "calendar.txt" file. d) Service exceptions are specified within the "stop_times.txt" file. | Common Reasoning | a | c | a |
| 122 | Can GTFS feeds include real-time trip updates? a) None of these b) No, GTFS feeds only contain static schedule information. c) Real-time trip updates are specified within the "trips.txt" file. d) Yes, GTFS feeds can include real-time trip updates. | Common Reasoning | d | d | a |
| 123 | How does GTFS handle multiple trips on the same route at the same time? a) GTFS does not allow multiple trips on the same route at the same time. b) By using the "trip_id" field in the "trips.txt" file. c) Multiple trips are represented as separate routes in GTFS. d) GTFS relies on real-time updates to handle such cases. | Common Reasoning | b | b | a |



| | Question | Type | Ground Truth | GPT-4 score | Phi-3 score |
|---|---|---|---|---|---|
| 124 | Can static GTFS feeds include information about service alerts? a) None of these b) No, GTFS feeds do not support service alerts. c) Service alerts are specified within the "calendar_dates.txt" file. d) Service alerts are specified within the "feed_info.txt" file. | Common Reasoning | b | b | unclear which answer has been given |
| 125 | Can static GTFS feeds include information about fare zones? a) None of these b) No, GTFS does not support fare zone information. c) Fare zone information is specified within the "fare_attributes.txt" file. d) Fare zone information is specified within the "routes.txt" file. | Common Reasoning | a | c | c |
| 126 | Can static GTFS feeds include information about fare zones? a) None of these b) No, GTFS does not support fare zone information. c) Fare zone information is specified within the "fare_attributes.txt" file. d) Fare zone information is specified within the "stops.txt" file. | Common Reasoning | d | d | c |
| 127 | Which file in GTFS schedule feeds include information about wheelchair accessibility? a) None of these b) stop_times.txt. c) stops.txt d) routes.txt. | Common Reasoning | c | c | c |
| 128 | Can GTFS schedule feeds include information about wheelchair accessibility? a) None of these b) GTFS does not support wheelchair accessibility information. c) Wheelchair accessibility information is mandatory. d) Wheelchair accessibility information is specified within the "routes.txt" file. | Common Reasoning | a | d | c |
| 129 | Can GTFS schedule feeds include information about parking facilities at transit stops? a) None of these b) No, GTFS does not support parking facility information. c) Parking facility information is specified within the "stops.txt" file. d) Parking facility information is specified within the "parking.txt" file. | Common Reasoning | b | b | a |
| 130 | What file contains information about the headways or time between trips for frequency-based service in | Common Reasoning | b | b | b |



| | Question | Type | Ground Truth | GPT-4 score | Phi-3 score |
|---|---|---|---|---|---|
| | GTFS? a) headways.txt b) frequencies.txt c) None of these d) headway_based_service.txt | | | | |
| 131 | How are vehicle capacities represented in GTFS? a) None of these b) By using the "capacities.txt" file. c) Vehicle capacities are not supported in GTFS. d) Vehicle capacities are specified within the "trips.txt" file. | Common Reasoning | c | c | d |
| 132 | Can GTFS feeds include information about service advisories or disruptions? a) None of these b) No, GTFS feeds do not support service advisory information. c) Service advisories are specified within the "calendar_dates.txt" file. d) Service advisories are specified within the "advisories.txt" file. | Common Reasoning | b | b | c |
| 133 | What is the purpose of the "translations.txt" file in GTFS? a) It provides translations of the GTFS data into different languages. b) It contains details about translations of fare rules and transfers. c) None of these d) It specifies translations of stop names and route descriptions. | Common Reasoning | a | a | b |
| 134 | Can GTFS feeds include information about real-time vehicle positions? a) None of these b) No, GTFS feeds do not support real-time vehicle positions. c) Real-time vehicle position information is specified within the "trips.txt" file. d) Real-time vehicle position information is specified within the "vehicle_positions.txt" file. | Common Reasoning | b | b | c |
| 135 | How are departure time values represented in GTFS? a) HH:MM:SS b) MM/DD/YYYY HH:MM c) None of these d) UNIX timestamp | Data Structure | a | a | b |
| 136 | How is the agency_name attribute represented in GTFS? a) Date b) None of these c) Text d) Integer | Data Structure | c | c | b |
| 137 | What data type is used for the arrival_time attribute in GTFS? a) Date b) None of these c) Text d) Float | Data Structure | b | b | a |
| 138 | How is the route_short_name attribute represented in GTFS? a) Date b) None of these c) Text d) Integer | Data Structure | c | c | b |
| 139 | What data type is used for the wheelchair_accessible attribute in GTFS? a) Date b) None of these c) Text d) Integer | Data Structure | d | b | b |
| 140 | How is the stop_lat attribute represented in GTFS? a) Date b) None of these c) Text d) Decimal Degree | Data Structure | d | d | b |



| | Question | Type | Ground Truth | GPT-4 score | Phi-3 score |
|---|---|---|---|---|---|
| 141 | What data type is used for the route_type attribute in GTFS? a) Date b) None of these c) Text d) Integer | Data Structure | d | d | b |
| 142 | How is the trip_id attribute represented in GTFS? a) Date b) None of these c) Text d) Integer | Data Structure | c | c | a |
| 143 | What data type is used for the stop_sequence attribute in GTFS? a) Date b) None of these c) Text d) Integer | Data Structure | d | d | d |
| 144 | How is the pickup_type attribute represented in GTFS? a) Date b) None of these c) Text d) Integer | Data Structure | d | d | b |
| 145 | What data type is used for the stop_lon attribute in GTFS? a) Date b) None of these c) Text d) Decimal Degree | Data Structure | d | d | d |
| 146 | How is the color attribute represented in GTFS? a) None of these 0) b) RGB value (e.g., 255,0,0) c) Named color (e.g., "Red") d)Hexadecimal value without #(e.g. FFF000) | Data Structure | d | d | a |
| 147 | What data type is used for the currency_code attribute in GTFS? a) Currency code (ISO 4217) b) Enum c) Four-letter currency abbreviation (e.g., "USDC") d) Integer code representing the currency | Data Structure | a | a | b |
| 148 | How is the currency_amount attribute represented in GTFS? a) Decimal value (ISO 4217) b) Nearest $5 multiple c) Integer d) Text (e.g., "$50.00") | Data Structure | a | a | a |
| 149 | What data type is used for the date attribute in GTFS? a) Date (YYYYMMDD) b) Enum c) None of these ") d) Integer (e.g., 20180913) | Data Structure | a | a | b |
| 150 | How is the email attribute represented in GTFS? a) Email address b) Enum c) None of these d) URL | Data Structure | a | a | d |
| 151 | What data type is used for the language_code attribute in GTFS? a) Language code (IETF BCP 47) b) Enum c) None of these ") d) Two-letter language code (e.g., "en") | Data Structure | a | a | b |
| 152 | How is the latitude attribute represented in GTFS? a) Float b) Integer (in degrees) c) Latitude (WGS84 decimal degrees) d) None of these ") | Data Structure | c | c | b |
| 153 | What data type is used for the longitude attribute in GTFS? a) Float b) Integer (in degrees) c) Longitude (WGS84 decimal degrees) d) None of these ") | Data Structure | c | c | c |
| 154 | How is the phone_number attribute represented in GTFS? a) A string of UTF-8 characters b) Enum c) None of these 5) 123-4567") d) Integer representation of phone number | Data Structure | c | a | a |



| | Question | Type | Ground Truth | GPT-4 score | Phi-3 score |
|---|---|---|---|---|---|
| 155 | What data type is used for the timezone attribute in GTFS? a) Enum b) None of these ") c) Timezone abbreviation (e.g., "PST") d) URL (link to timezone information) | Data Structure | b | b | c |
| 156 | What data type is used for the fare_amount attribute in GTFS? a) Float (currency amount) b) Enum (e.g., "Low", "Medium", "High") c) Integer (number of cents) d) None of these | Data Structure | a | c | d |
| 157 | How is the route_long_name attribute represented in GTFS? a) None of these b) Integer (route identifier) c) Enum (e.g., "Red Line", "Blue Line", "Green Line") d) Time representation of route name | Data Structure | a | c | d |
| 158 | What data type is used for the zone_id attribute in GTFS? a) Integer b) Enum (e.g., "Zone A", "Zone B", "Zone C") c) Float (latitude and longitude coordinates) d) UTF-8 characters | Data Structure | d | d | a |
| 159 | How is the wheelchair_accessible attribute represented in GTFS? a) Boolean (true or false) b) Float (number of accessible seats) c) Enum (e.g., 0,1,2) d) None of these | Data Structure | c | c | c |
| 160 | What data type is used for the stop_lat attribute in GTFS? a) Float (latitude) b) Enum (e.g., "North", "South", "East", "West") c) Integer (number of degrees) d) None of these | Data Structure | a | a | unclear which answer has been given |
| 161 | How is the pickup_type attribute represented in GTFS? a) None of these b) Integer (number of passengers) c) Float (duration in minutes) d) Enum (e.g., "Regular", "Not available") | Data Structure | d | d | a |
| 162 | What data type is used for the route_url attribute in GTFS? a) None of these b) Enum (e.g., "Website", "Mobile App") c) Integer (route identifier) d) Text representation of route URL | Data Structure | a | d | b |



| | Question | Type | Ground Truth | GPT-4 score | Phi-3 score |
|---|---|---|---|---|---|
| 163 | How is the departure_date attribute represented in GTFS? a) Date b) Integer (number of days from a reference date) c) Enum (e.g., "Today", "Tomorrow") d) None of these | Data Structure | a | b | a |
| 164 | What data type is used for the agency_name attribute in GTFS? a) None of these r)  b) Enum (e.g., "Company A", "Company B", "Company C") c) Text d) Time representation of agency name | Data Structure | c | c | c |
| 165 | How is calendar information represented in GTFS? a) calendar.txt b) dates.txt c) None of these d) schedules.txt | File Structure | a | a | b |
| 166 | How is calendar information represented in GTFS? a) None of these b) dates.txt c) calendar_dates.txt d) schedules.txt | File Structure | c | c | c |
| 167 | What file in GTFS provides information about the accessibility of transit stops? a) None of these b) stops.txt c) transfers.txt d) accessibility.txt | File Structure | b | b | d |
| 168 | How is frequency-based service represented in GTFS? a) None of these b) By using the "calendar.txt" file. c) By specifying trip times in the "stop_times.txt" file. d) Frequency-based service is not supported in GTFS. | File Structure | a | a | b |
| 169 | What file contains information about fare rules and transfers in GTFS? a) None of these b) transfers.txt c) trips.txt d) stops.txt | File Structure | b | b | d |
| 170 | What file contains information about fare rules and transfers in GTFS? a) None of these b) stops.txt c) trips.txt d) fares.txt | File Structure | d | d | a |
| 171 | How is the direction of travel represented in GTFS? a) By using the "direction_id" field in the "routes.txt" file. b) By specifying the latitude and longitude coordinates of the direction. c) None of these d) By using the "direction" field in the "trips.txt" file. | File Structure | d | d | b |
| 172 | What is the purpose of the "transfers.txt" file in GTFS? a) It contains information about fare rules and transfers. b) It provides details about the geographic shapes of routes. c) It specifies the frequency of trips. d) It provides real-time arrival and departure information. | File Structure | a | a | b |
| 173 | What is the purpose of the "attributions.txt" file in GTFS? a) It contains information about fare rules and transfers. b) It specifies the frequency of trips. c) It | File Structure | d | d | c |



| | Question | Type | Ground Truth | GPT-4 score | Phi-3 score |
|---|---|---|---|---|---|
| | provides details about the geographic shapes of routes. d) It credits the sources or contributors of the GTFS data. | | | | |
| 174 | What is the purpose of the "feed_info.txt" file in GTFS? a) It provides information about individual transit stops. b) It contains details about the routes and their associated stops. c) It specifies the frequency of trips. d) It provides general information about the GTFS feed. | File Structure | d | d | d |
| 175 | How are fare attributes represented in GTFS? a) None of these b) By using the "fares.txt" file. c) Fare attributes are not supported in GTFS. d) Fare attributes are specified within the "stops.txt" file. | File Structure | a | b | c |
| 176 | How are transfer rules represented in GTFS? a) None of these b) By using the "connections.txt" file. c) Transfer rules are not applicable in GTFS. d) Transfer rules are specified within the "routes.txt" file. | File Structure | a | a | b |
| 177 | What is the purpose of the "fare_rules.txt" file in GTFS? a) It provides information about individual transit stops. b) It contains details about fare attributes. c) It specifies the frequency of trips. d) It provides information about fare rules and fare availability. | File Structure | d | d | c |
| 178 | What file contains information about the geographic shapes of routes in GTFS? a) None of these b) routes.txt c) shapes.txt d) trips.txt | File Structure | c | c | b |
| 179 | How are route colors represented in GTFS? a) None of these b) By using the "colors.txt" file. c) Route colors are not supported in GTFS. d) Route colors are specified within the "routes.txt" file. | File Structure | d | d | d |
| 180 | What is the purpose of the "levels.txt" file in GTFS? a) It provides information about the levels of fare zones. b) It contains details about the levels or floors of transit stations. c) None of these d) It specifies the elevation or altitude of transit stops. | File Structure | b | b | b |
| 181 | How are transfer durations represented in GTFS? a) None of these b) By using the "durations.txt" file. c) Transfer durations are not supported in GTFS. d) Transfer durations are specified within the "transfers.txt" file. | File Structure | d | d | b |
| 182 | What is a dataset in the context of GTFS? a) A single file containing all transit information b) A collection of | Term Definition | b | b | a |



| | Question | Type | Ground Truth | GPT-4 score | Phi-3 score |
|---|---|---|---|---|---|
| | tables representing different entities c) A specific date for transit service d) A record representing a transit agency | | | | |
| 183 | What is a record in GTFS? a) A basic data structure representing a service day b) A collection of field values describing a single entity c) A set of files defining transit information d) A unique identifier for a transit agency | Term Definition | b | b | a |
| 184 | What is a field in GTFS? a) A property of a dataset b) A specific date for transit service c) A basic data structure representing a service day d) A property of an object or entity | Term Definition | d | a | d |
| 185 | What is a service day in GTFS? a) None of these b) A specific day of the week when transit service is available c) A record representing a transit agency d) A time period used to indicate route scheduling | Term Definition | d | b | d |
| 186 | What is a text-to-speech field in GTFS? a) A field containing speech recognition data b) A field specifically designed for text messaging c) A field aimed to be read by text-to-speech systems d) A field representing the origin and destination of a journey | Term Definition | c | c | b |
| 187 | What is a leg in GTFS? a) A basic data structure representing a service day b) An individual entry in a field c) None of these d) A specific date for transit service | Term Definition | c | c | c |
| 188 | What is a journey in GTFS? a) A specific date for transit service b) An individual entry in a field c) None of these d) Overall travel from origin to destination, including all legs and transfers | Term Definition | d | d | b |
| 189 | What is a sub-journey in GTFS? a) A subset of legs that comprise a journey b) A specific date for transit service c) A field containing speech recognition data d) A record representing a transit agency | Term Definition | a | a | d |
| 190 | What is a fare product in GTFS? a) A specific date for transit service b) A record representing a transit agency c) None of these d) A collection of field values describing a single entity | Term Definition | c | d | a |
| 191 | What is a field value in GTFS? a) A specific date for transit service b) An individual entry in a field c) None of these d) A field specifically designed for text messaging | Term Definition | b | b | a |



| | Question | Type | Ground Truth | GPT-4 score | Phi-3 score |
|---|---|---|---|---|---|
| 192 | What is the presence condition for a "Required" field or file in GTFS? a) None of these b) The field or file must be included in the dataset and contain a valid value for each record. c) The field or file must not be included under certain conditions. d) The field or file must be included under certain conditions. | Term Definition | b | b | unclear which answer has been given |
| 193 | When a field or file is marked as "Optional" in GTFS, what does it indicate? a) None of these b) The field or file must be included in the dataset and contain a valid value for each record. c) The field or file must not be included under certain conditions. d) The field or file must be included under certain conditions. | Term Definition | a | d | b |
| 194 | When a field or file is marked as "Conditionally Forbidden" in GTFS, what does it mean? a) None of these b) The field or file must be included in the dataset and contain a valid value for each record. c) The field or file must not be included under certain conditions outlined in the description. d) The field or file must be included under certain conditions outlined in the description. | Term Definition | c | d | d |
| 195 | When a field or file is marked as "Conditionally Required" in GTFS, what determines its inclusion? a) It must be included in the dataset. b) It may be omitted from the dataset. c) It must not be included under certain conditions outlined in the description. d) It must be included under certain conditions outlined in the description. | Term Definition | d | d | a |



# Appendix D
## Understanding MATSim Examination Questions

The following 120 questions were based on the MATSim user guide by Horni (2016).

| | Question | Section | Ground Truth | GPT-4 score | Phi-3 score |
|---|---|---|---|---|---|
| 1 | What was the original inspiration behind MATSim?<br>a) Geographic Information Systems<br>b) TRANSIMS project<br>c) Public transport systems<br>d) Machine learning algorithms | Introducing MATSim | b | b | b |
| 2 | MATSim performs integral microscopic simulation of what?<br>a) Financial markets<br>b) Traffic flows and congestion<br>c) Weather patterns<br>d) Internet traffic | Introducing MATSim | b | b | b |
| 3 | What programming language is MATSim implemented in?<br>a) Python<br>b) C++<br>c) Java<br>d) R | Introducing MATSim | c | c | b |
| 4 | What type of model does MATSim use for network loading simulation?<br>a) Agent-based model<br>b) Cellular automaton<br>c) Queue-based model<br>d) Car-following model | Introducing MATSim | c | a | a |
| 5 | How is MATSim designed to model daily activities?<br>a) Using weekly schedules<br>b) Using monthly plans<br>c) Using single-day models<br>d) Using hourly schedules | Introducing MATSim | c | c | d |
| 6 | In MATSim, what is the term for the people modeled in simulations?<br>a) Particles<br>b) Agents<br>c) Users<br>d) Nodes | Introducing MATSim | b | b | b |
| 7 | What method does MATSim use to optimize daily activity schedules? | Introducing MATSim | b | b | a |



| | | | | | |
|---|---|---|---|---|---|
| | a) Genetic algorithms<br>b) Co-evolutionary algorithms<br>c) Simulated annealing<br>d) Neural networks | | | | |
| 8 | What does the 'replanning' module in MATSim allow?<br>a) Dynamic traffic signal adjustment<br>b) Real-time traffic flow control<br>c) Modification of agents' plans<br>d) Live traffic updates | Introducing MATSim | c | c | c |
| 9 | Which of the following is NOT a dimension considered in MATSim replanning?<br>a) Route choice<br>b) Mode choice<br>c) Weather forecasting<br>d) Departure time | Introducing MATSim | c | c | c |
| 10 | What does MATSim use to store and evaluate the performance of each agentâ€™s daily plans?<br>a) Memory units<br>b) Data clusters<br>c) Score functions<br>d) Performance logs | Introducing MATSim | c | c | c |
| 11 | What is the core principle behind MATSim's optimization process?<br>a) Linear regression<br>b) Game theory<br>c) Co-evolutionary search<br>d) Decision trees | Introducing MATSim | c | c | c |
| 12 | What type of simulation is QSim in MATSim?<br>a) Cellular automaton<br>b) Microscopic simulation<br>c) Queue-based simulation<br>d) Agent-based simulation | Introducing MATSim | c | c | b |
| 13 | Which module in MATSim is responsible for handling events during simulations?<br>a) Controler<br>b) Network<br>c) Population<br>d) Events | Introducing MATSim | d | d | d |
| 14 | MATSimâ€™s scoring function can be interpreted as what?<br>a) Physical effort<br>b) Econometric utility<br>c) Computational cost<br>d) Random value | Introducing MATSim | b | b | b |
| 15 | In MATSim, which module determines the sequence of daily activities for agents?<br>a) Controler<br>b) Population | Introducing MATSim | b | b | b |



| | | | | | |
|---|---|---|---|---|---|
| | c) Network<br>d) Strategy | | | | |
| 16 | What is a common unit of analysis for activity-based models in MATSim?<br>a) Hour<br>b) Day<br>c) Week<br>d) Month | Introducing MATSim | b | b | b |
| 17 | MATSim is primarily designed to handle what scale of scenarios?<br>a) Small-scale<br>b) Large-scale<br>c) Medium-scale<br>d) Micro-scale | Introducing MATSim | b | b | b |
| 18 | What is the role of the 'controler' module in MATSim?<br>a) To manage agent interactions<br>b) To control the simulation execution<br>c) To visualize traffic flows<br>d) To generate random events | Introducing MATSim | b | b | b |
| 19 | MATSim's traffic flow model is based on which two attributes of a network link?<br>a) Length and width<br>b) Capacity and speed<br>c) Storage and flow capacity<br>d) Direction and altitude | Introducing MATSim | c | b | b |
| 20 | What does the iterative process in MATSim aim to stabilize?<br>a) Agent locations<br>b) Average population score<br>c) Traffic signal timings<br>d) Road capacities | Introducing MATSim | b | b | b |
| 21 | What is Via used for?<br>a) Simulating traffic flows<br>b) Visualizing and analyzing MATSim simulation results<br>c) Generating random traffic data<br>d) Managing public transportation schedules | Visualizers | b | b | b |
| 22 | Who currently owns and develops Via?<br>a) ETH Zurich<br>b) MATSim Consortium<br>c) Simunto GmbH<br>d) Senozon AG | Visualizers | c | c | d |
| 23 | Which programming language is Via written in?<br>a) Python<br>b) C++<br>c) Java<br>d) R | Visualizers | c | b | b |
| 24 | How can data files be added to Via's data source?<br>a) Drag-and-drop onto the layers list | Visualizers | a | a | c |



| | | | | | |
|---|---|---|---|---|---|
| | b) Emailing them to support<br>c) Entering file paths manually<br>d) Using a command-line interface | | | | |
| 25 | What year was the first version of Via released?<br>a) 2008<br>b) 2010<br>c) 2011<br>d) 2012 | Visualizers | c | b | c |
| 26 | Which MATSim file is needed for animated visualization of agents in Via?<br>a) network.xml<br>b) population.xml<br>c) events.xml<br>d) config.xml | Visualizers | c | c | b |
| 27 | What feature allows Via users to visualize the occupancy of transit vehicles?<br>a) Vehicle speed analysis<br>b) Route flow analysis<br>c) Occupancy attribute visualization<br>d) Facility load analysis | Visualizers | c | c | c |
| 28 | Which feature is NOT available in the default version of Via?<br>a) Network visualization<br>b) Public transport analysis<br>c) Customizable vehicle symbols<br>d) Real-time traffic updates | Visualizers | d | d | c |
| 29 | How are queries made in Via?<br>a) By writing scripts<br>b) By clicking question-mark icons<br>c) By calling a support hotline<br>d) By submitting forms online | Visualizers | b | a | b |
| 30 | What can be visualized using the Select Facility Analysis in Via?<br>a) Transit routes<br>b) Facility loads and travel patterns<br>c) Weather impacts on traffic<br>d) Real-time vehicle locations | Visualizers | b | b | b |
| 31 | Which file can enhance the detail of activity locations in Via?<br>a) events.xml<br>b) config.xml<br>c) population.xml<br>d) network.xml | Visualizers | c | c | a |
| 32 | What is one way Via supports public transport analysis?<br>a) By visualizing passenger flows between stops<br>b) By predicting future traffic trends<br>c) By generating random bus schedules<br>d) By simulating weather conditions | Visualizers | a | a | a |



| 33 | When was ownership of Via transferred to Simunto GmbH?<br>a) 2011<br>b) 2015<br>c) 2018<br>d) 2020 | Visualizers | c | c | b |
|----|----|----|----|----|----|
| 34 | How can Via users add a new layer to the visualization?<br>a) By pressing the plus icon in the lower left corner<br>b) By sending a request to Simunto GmbH<br>c) By installing additional software<br>d) By manually editing configuration files | Visualizers | c | a | a |
| 35 | What format is used for Via's visualization results?<br>a) XML<br>b) JSON<br>c) CSV<br>d) HTML | Visualizers | b | b | b |
| 36 | What type of analysis does the public transport plugin provide in Via?<br>a) Air quality analysis<br>b) Transit occupancy and route flow analysis<br>c) Road maintenance scheduling<br>d) Weather impact analysis | Visualizers | b | b | b |
| 37 | What method allows Via to reset its window state?<br>a) Restarting the application<br>b) Choosing 'Reset Window State' from the Window menu<br>c) Reinstalling the software<br>d) Clearing the application cache | Visualizers | a | b | b |
| 38 | How can Via visualize agent plans?<br>a) By using a specific toolbar query icon<br>b) By entering plan details manually<br>c) By exporting plans to a PDF<br>d) By requesting a plan from the support team | Visualizers | b | a | a |
| 39 | What data is required to perform an animated visualization in Via?<br>a) Only network.xml<br>b) Both network.xml and events.xml<br>c) Only events.xml<br>d) Both events.xml and config.xml | Visualizers | b | b | b |
| 40 | How are public transport vehicles displayed in Via?<br>a) As green triangular symbols<br>b) As pink rectangular symbols<br>c) As blue circular symbols<br>d) As yellow star symbols | Visualizers | b | c | c |
| 41 | What are the minimal inputs for a MATSim simulation besides the configuration?<br>a) Network and population with initial plans<br>b) Weather data and traffic signals | Generation of MATSim Input | a | a | a |



| | | | | | |
|---|---|---|---|---|---|
| | c) Public transportation schedules and routes<br>d) Environmental data and pollution levels | | | | |
| 42 | Which scripting language do the authors use for generating MATSim input?<br>a) Python<br>b) Java<br>c) C++<br>d) R | Generation of MATSim Input | b | a | a |
| 43 | Which tool is recommended for coordinate transformations in Java for MATSim?<br>a) GeoPy<br>b) GDAL<br>c) Geotools<br>d) Proj4 | Generation of MATSim Input | c | c | d |
| 44 | Where can you download the necessary `.osm.pbf` file for generating a MATSim network?<br>a) OpenStreetMap Wiki<br>b) GeoTools official site<br>c) Geofabrik<br>d) MATSim official site | Generation of MATSim Input | c | c | c |
| 45 | What command line tool is used to process OSM data for MATSim?<br>a) GDAL<br>b) Osmosis<br>c) QGIS<br>d) ArcGIS | Generation of MATSim Input | b | b | b |
| 46 | What is the primary purpose of the Osmosis command `--bounding-box`?<br>a) To filter nodes<br>b) To extract the road network within a specified area<br>c) To convert OSM data to XML format<br>d) To calculate road elevations | Generation of MATSim Input | b | a | b |
| 47 | Which file format is the result of merging OSM road data with large roads from a larger region?<br>a) merged-network.xml<br>b) allroads.osm.pbf<br>c) network.osm<br>d) merged-network.osm | Generation of MATSim Input | d | d | d |
| 48 | What does the script `RunPNetworkGenerator` in MATSim do?<br>a) Generates a network file from merged OSM data<br>b) Converts population data to XML format<br>c) Analyzes traffic flow in real-time<br>d) Generates random traffic events | Generation of MATSim Input | a | a | a |
| 49 | Which MATSim code example is used to convert an EMME network to MATSim format?<br>a) RunPNetworkGenerator<br>b) RunNetworkEmme2MatsimExample | Generation of MATSim Input | b | b | b |



| | | | | | |
|---|---|---|---|---|---|
| | c) RunPPopulationGenerator<br>d) RunZPopulationGenerator | | | | |
| 50 | What should be consistent between the population and network files in MATSim?<br>a) File formats<br>b) Coordinate systems<br>c) File sizes<br>d) Simulation durations | Generation of MATSim Input | b | b | b |
| 51 | What script generates a single synthetic person for a MATSim population?<br>a) RunOnePersonPopulationGenerator<br>b) RunZPopulationGenerator<br>c) RunNetworkEmme2MatsimExample<br>d) RunPNetworkGenerator | Generation of MATSim Input | a | a | a |
| 52 | Which script is used to generate a larger synthetic population in MATSim?<br>a) RunPNetworkGenerator<br>b) RunNetworkEmme2MatsimExample<br>c) RunPPopulationGenerator<br>d) RunZPopulationGenerator | Generation of MATSim Input | c | c | c |
| 53 | For realistic initial demand generation, which script is used?<br>a) RunOnePersonPopulationGenerator<br>b) RunPPopulationGenerator<br>c) RunZPopulationGenerator<br>d) RunNetworkEmme2MatsimExample | Generation of MATSim Input | c | b | b |
| 54 | Which command is used to merge two OSM files in the network generation process?<br>a) osmosis --merge<br>b) osmosis --combine<br>c) osmosis --integrate<br>d) osmosis --unite | Generation of MATSim Input | a | a | a |
| 55 | What section discusses the practical level generation of initial input?<br>a) Section 2.3<br>b) Part IV<br>c) Section 8.1<br>d) Part III | Generation of MATSim Input | b | a | c |
| 56 | Which scripting language is NOT mentioned as used for generating MATSim input?<br>a) Java<br>b) Python<br>c) C++<br>d) R | Generation of MATSim Input | d | c | d |
| 57 | What type of data does the `RunOnePersonPopulationGenerator` script generate?<br>a) Traffic signal data<br>b) Pollution data | Generation of MATSim Input | c | c | c |



| | | | | | |
|---|---|---|---|---|---|
| | c) Single synthetic person population<br>d) Environmental data | | | | |
| 58 | Which data source is mentioned for network generation besides OSM?<br>a) GIS databases<br>b) EMME network files<br>c) Traffic surveys<br>d) Satellite imagery | Generation of MATSim Input | b | b | b |
| 59 | What is a problem with EMME network files?<br>a) They are too large<br>b) They use user-defined variables in non-standardized ways<br>c) They are in a binary format<br>d) They are not compatible with Java | Generation of MATSim Input | b | b | b |
| 60 | What tool is used for generating coordinate transformations in MATSim?<br>a) TransformationFactory<br>b) GeoTransform<br>c) CoordConvert<br>d) CoordinateTool | Generation of MATSim Input | a | a | b |
| 61 | What is the main purpose of the Attributable interface in MATSim?<br>a) To store network configurations<br>b) To attach arbitrary attributes to data types<br>c) To visualize simulation results<br>d) To manage public transport schedules | Additional Core Features | b | b | b |
| 62 | Which file format is used to define additional attributes for persons in MATSim?<br>a) JSON<br>b) CSV<br>c) XML<br>d) TXT | Additional Core Features | c | c | c |
| 63 | What is the correct syntax for adding an attribute to a person in MATSim's XML format?<br>a) <person id="1"><age>21</age></person><br>b) <person id="1"><attributes><attribute name="age" class="java.lang.Integer">21</attribute></attributes></person><br>c) <person id="1"><age value="21" /><br>d) <person id="1" age="21"></person> | Additional Core Features | b | b | b |
| 64 | What is a feature of time-dependent networks in MATSim?<br>a) Fixed network attributes<br>b) Time-variant network attributes<br>c) Static traffic conditions<br>d) Constant speed limits | Additional Core Features | b | b | b |



| 65 | What attribute can be time-variant in a MATSim network?<br>a) Population density<br>b) Weather conditions<br>c) Free speed of links<br>d) Public transport schedules | Additional Core Features | c | c | d |
|----|----|----|----|----|----|
| 66 | How is a subpopulation defined in MATSim?<br>a) By specifying age groups<br>b) By assigning a specific attribute to persons<br>c) By grouping links<br>d) By categorizing nodes | Additional Core Features | b | b | b |
| 67 | What is the purpose of the `subpopulation` attribute in MATSim?<br>a) To define different network configurations<br>b) To apply different replanning strategies to different groups of persons<br>c) To visualize simulation results<br>d) To generate random traffic data | Additional Core Features | b | b | b |
| 68 | What type of file is used to provide traffic count data in MATSim?<br>a) config.xml<br>b) counts.xml<br>c) population.xml<br>d) network.xml | Additional Core Features | b | b | b |
| 69 | What is the purpose of the `counts.xml` file in MATSim?<br>a) To simulate weather impacts<br>b) To compare simulated and real-world traffic volumes<br>c) To generate synthetic populations<br>d) To store network configurations | Additional Core Features | b | b | b |
| 70 | What does the `facilities.xml` file define in MATSim?<br>a) Network link attributes<br>b) Public transport schedules<br>c) Locations where agents perform activities<br>d) Traffic signal configurations | Additional Core Features | c | c | c |
| 71 | How are facilities typically used in MATSim?<br>a) To visualize traffic flows<br>b) To define activity locations for agents<br>c) To manage traffic signals<br>d) To simulate environmental impacts | Additional Core Features | b | b | b |
| 72 | What is one of the main purposes of the households module in MATSim?<br>a) To simulate weather conditions<br>b) To manage vehicle data<br>c) To group agents into households for more realistic simulations<br>d) To generate network configurations | Additional Core Features | c | c | c |



| 73 | Which file format is used to define vehicle types in MATSim?<br>a) vehicles.xml<br>b) traffic.xml<br>c) transport.xml<br>d) cars.xml | Additional Core Features | a | a | a |
|---|---|---|---|---|---|
| 74 | What attribute is required for each vehicle type in MATSim?<br>a) Vehicle color<br>b) Passenger capacity<br>c) License plate number<br>d) Owner's name | Additional Core Features | b | b | b |
| 75 | What type of event does MATSim's public transport simulation generate?<br>a) Weather update events<br>b) Traffic signal change events<br>c) Transit vehicle arrival and departure events<br>d) Road maintenance events | Additional Core Features | c | c | c |
| 76 | What data is necessary for simulating public transport in MATSim?<br>a) Pollution levels<br>b) Transit vehicle types and schedules<br>c) Weather forecasts<br>d) Road quality information | Additional Core Features | b | b | b |
| 77 | What is the role of the `qsim` module in MATSim?<br>a) To generate population data<br>b) To control the simulation execution<br>c) To visualize results<br>d) To edit network configurations | Additional Core Features | b | b | b |
| 78 | How does MATSim handle multiple vehicle types in the same simulation?<br>a) By ignoring the differences<br>b) By using the same configuration for all vehicles<br>c) By assigning specific characteristics to each vehicle type<br>d) By simulating only one vehicle type at a time | Additional Core Features | c | c | c |
| 79 | What is the main purpose of the PassingQ feature in MATSim?<br>a) To simulate pedestrian traffic<br>b) To allow faster vehicles to pass slower vehicles<br>c) To manage traffic signals<br>d) To handle public transport schedules | Additional Core Features | b | b | b |
| 80 | What is a leg in MATSim's plans format?<br>a) A network link<br>b) A type of activity<br>c) A segment of travel between two activities<br>d) A type of facility | Additional Core Features | c | c | c |



| 81 | What is the main purpose of using the `output_persons.csv.gz` file in MATSim?<br>a) To store traffic signal data<br>b) To provide demographic and simulation output data for every person<br>c) To log environmental conditions<br>d) To manage vehicle registrations | Analysis | b | b | b |
|----|----|----|----|----|----|
| 82 | What delimiter is used in the `output_legs.csv.gz` file?<br>a) Comma<br>b) Semicolon<br>c) Tab<br>d) Space | Analysis | b | a | c |
| 83 | Which file format is used to compress MATSim output files?<br>a) ZIP<br>b) TAR<br>c) Gzip<br>d) Bzip2 | Analysis | c | c | c |
| 84 | What should be included in the header of a CSV file to indicate the coordinate reference system?<br>a) // EPSG code<br>b) # EPSG code<br>c) -- EPSG code<br>d) %% EPSG code | Analysis | b | b | a |
| 85 | What is the primary benefit of using the Python API for MATSim?<br>a) It is faster than Java<br>b) It integrates well with pandas and geopandas for data analysis<br>c) It does not require installation<br>d) It supports only CSV files | Analysis | b | b | b |
| 86 | Which command is used to install the MATSim Python tools package from PyPi?<br>a) pip install matsim-python<br>b) pip install matsim<br>c) pip install matsim-tools<br>d) pip install python-matsim | Analysis | c | a | a |
| 87 | What does the `matsim.read_network` function return?<br>a) A list of events<br>b) Two pandas dataframes for nodes and links<br>c) A list of plans<br>d) A CSV file of trips | Analysis | b | b | b |
| 88 | What type of emissions are calculated during a vehicle's warm-up phase?<br>a) Warm emissions<br>b) Cold-start emissions<br>c) Evaporation emissions<br>d) Idle emissions | Analysis | b | b | b |



| 89 | What is the purpose of the `createLookupTables` method in the EmissionModule?<br>a) To visualize emissions data<br>b) To create lookup tables from input data exported from the HBEFA database<br>c) To manage public transport schedules<br>d) To simulate noise pollution | Analysis | b | b | b |
|----|---|---|---|---|---|
| 90 | What attribute is required for each vehicle type in the `emissionVehicles` table?<br>a) Vehicle color<br>b) Passenger capacity<br>c) License plate number<br>d) Owner's name | Analysis | b | b | c |
| 91 | Which data analysis library does the MATSim R library use internally?<br>a) dplyr<br>b) data.table<br>c) tidyverse<br>d) sqldf | Analysis | c | b | b |
| 92 | What is the purpose of the `NoiseOfflineCalculationExample` class in MATSim?<br>a) To calculate traffic emissions<br>b) To simulate noise pollution<br>c) To manage public transport<br>d) To analyze travel distances | Analysis | b | b | b |
| 93 | What does the `output_trips.csv.gz` file provide statistics for?<br>a) Every person in the simulation<br>b) Every leg in the simulation<br>c) Every trip in the simulation<br>d) Environmental data for the region | Analysis | c | c | c |
| 94 | What should be used to escape double quotes inside a field in a CSV file?<br>a) Single quotes<br>b) Backslashes<br>c) Double quotes<br>d) Angle brackets | Analysis | b | c | a |
| 95 | What is the main purpose of the MATSim R API?<br>a) To generate synthetic populations<br>b) To integrate MATSim outputs into R for data analysis<br>c) To simulate traffic flow in real-time<br>d) To manage transport schedules | Analysis | b | b | b |
| 96 | What is the main function of the `RunAccidents` class in the MATSim accidents module?<br>a) To log vehicle emissions<br>b) To simulate road accidents<br>c) To manage public transport<br>d) To analyze travel distances | Analysis | a | b | b |



| 97 | What is a typical issue with CSV file format implementations?<br>a) Varying delimiters<br>b) Consistent data type settings<br>c) Inability to include comments<br>d) Lack of header rows | Analysis | b | a | a |
|-----|-----|-----|-----|-----|-----|
| 98 | What type of data format is CSV?<br>a) Column-based text format<br>b) Row-based text format<br>c) Binary format<br>d) Image format | Analysis | b | a | a |
| 99 | What is the main advantage of using a header row in a CSV file?<br>a) It increases file size<br>b) It makes the file readable by most software<br>c) It decreases processing speed<br>d) It removes the need for delimiters | Analysis | b | b | b |
| 100 | What does the `avgHbefaWarmTable` file provide in the MATSim emission model?<br>a) Cold emission factors<br>b) Warm emission factors<br>c) Vehicle type mappings<br>d) Road type mappings | Analysis | b | b | b |
| 101 | What is the primary purpose of the Dynamic Transport Services module in MATSim?<br>a) To manage static traffic signals<br>b) To optimize dynamic vehicle routing and scheduling<br>c) To simulate weather conditions<br>d) To monitor public transport schedules | Extending MatSIM | b | b | b |
| 102 | What does DVRP stand for in MATSim?<br>a) Dynamic Vehicle Routing Problem<br>b) Dynamic Vehicle Replanning Process<br>c) Dynamic Variable Routing Procedure<br>d) Dynamic Vehicle Resource Planning | Extending MatSIM | a | a | b |
| 103 | What type of vehicle routing problems can the DVRP contribution model?<br>a) Fixed-route VRPs only<br>b) One-to-many and many-to-many VRPs<br>c) Public transport scheduling only<br>d) Weather-dependent VRPs only | Extending MatSIM | b | b | b |
| 104 | What is the function of the `Schedule` in the DVRP model?<br>a) To track environmental data<br>b) To manage vehicle maintenance schedules<br>c) To sequence tasks for vehicles<br>d) To monitor traffic signals | Extending MatSIM | c | c | c |
| 105 | What type of task does `DriveTask` represent in the DVRP model? | Extending MatSIM | b | b | b |



| | | | | | |
|---|---|---|---|---|---|
| | a) A task to park the vehicle<br>b) A task to drive from one location to another<br>c) A task to wait at a location<br>d) A task to refuel the vehicle | | | | |
| 106 | What is the purpose of `TaskTracker` in the DVRP model?<br>a) To monitor and predict task execution times<br>b) To log environmental conditions<br>c) To schedule vehicle maintenance<br>d) To simulate passenger behavior | Extending MatSIM | a | a | a |
| 107 | What does `VrpPathCalculator` do in the DVRP model?<br>a) It calculates the shortest paths for vehicles<br>b) It manages vehicle schedules<br>c) It simulates traffic conditions<br>d) It tracks vehicle emissions | Extending MatSIM | a | a | a |
| 108 | What does the `DynAgent` class represent in MATSim?<br>a) A static traffic signal<br>b) A dynamic driver or passenger agent<br>c) A weather monitoring system<br>d) A vehicle maintenance tracker | Extending MatSIM | b | b | b |
| 109 | How does `DynAgent` determine its next action in the simulation?<br>a) Based on pre-computed plans<br>b) According to real-time decisions from `DynAgentLogic`<br>c) By following a fixed schedule<br>d) By randomly selecting tasks | Extending MatSIM | b | b | b |
| 110 | What is the role of `PassengerEngine` in the DVRP model?<br>a) To monitor vehicle emissions<br>b) To manage passenger requests and interactions<br>c) To simulate weather conditions<br>d) To schedule vehicle maintenance | Extending MatSIM | b | b | b |
| 111 | What is the `VrpOptimizer` responsible for in the DVRP model?<br>a) Calculating the shortest path<br>b) Optimizing vehicle schedules based on demand<br>c) Monitoring traffic signals<br>d) Tracking environmental data | Extending MatSIM | b | b | b |
| 112 | Which method in `VrpOptimizer` is called when a request is submitted?<br>a) `nextLinkEntered`<br>b) `initQSim`<br>c) `requestSubmitted`<br>d) `computeNextAction` | Extending MatSIM | c | c | c |
| 113 | What is the main advantage of asynchronous event handling in DVRP optimization?<br>a) It reduces computational speed | Extending MatSIM | b | b | b |



| | | | | | |
|---|---|---|---|---|---|
| | b) It increases computational efficiency and avoids conflicting decisions<br>c) It simplifies vehicle tracking<br>d) It eliminates the need for schedule updates | | | | |
| 114 | What type of requests can `PassengerEngine` handle?<br>a) Only immediate requests<br>b) Both immediate and advance requests<br>c) Only advance requests<br>d) Only recurring requests | Extending MatSIM | b | b | b |
| 115 | What does the `OneTaxiRequestCreator` class do in the OneTaxi example?<br>a) It manages vehicle emissions<br>b) It converts taxi calls into requests<br>c) It simulates weather conditions<br>d) It tracks vehicle maintenance | Extending MatSIM | b | b | b |
| 116 | What is the role of the `OneTaxiOptimizer` in the OneTaxi example?<br>a) To calculate vehicle emissions<br>b) To create and update the driver's schedule<br>c) To monitor traffic signals<br>d) To log environmental data | Extending MatSIM | b | b | b |
| 117 | In the OneTaxi example, how many taxi customers are served by the single taxi driver?<br>a) Five<br>b) Ten<br>c) Fifteen<br>d) Twenty | Extending MatSIM | b | a | a |
| 118 | What is the focus of research using the DVRP contribution in Berlin and Poznan?<br>a) On-line dispatching of electric taxis<br>b) Weather impact on transport systems<br>c) Emission control of public transport<br>d) Traffic signal optimization | Extending MatSIM | a | a | a |
| 119 | What type of planning is considered a future research topic for the DVRP contribution?<br>a) Static route planning<br>b) Multi-modal path search<br>c) Vehicle maintenance scheduling<br>d) Weather impact modeling | Extending MatSIM | b | d | b |
| 120 | What type of functionality is being developed to increase the DVRP contribution's flexibility?<br>a) Real-time weather simulation<br>b) Shortest paths search performance<br>c) Emission tracking<br>d) Vehicle maintenance alerts | Extending MatSIM | b | b | b |



# Appendix E
## Price Elasticity of Demand Approach

|   | Question | GPT-4 response | Phi-3 response |
|---|---|---|---|
| *1* | Advise the planner on how to assess the impact of changing one of the pricing schedule at one of the ERP gantries. | To assess the impact of changing the pricing schedule at one of the ERP gantries, the road transport planner can follow these steps:<br><br>1. Establish the objective: Clearly define the goal of increasing the price at one of the ERP gantries. It could be to reduce congestion, increase revenue for infrastructure development, or improve overall traffic flow.<br><br>2. Collect data: Gather historical traffic data, including vehicle counts, vehicle types, and time stamps from the specific gantry and adjacent gantries. Collect the pricing schedules currently in place and any past pricing changes.<br><br>3. Identify key performance indicators (KPIs): Determine which KPIs are relevant to the objective, such as average travel time, average speed, vehicle count, and revenue generated.<br><br>4. Create a traffic model: Develop a traffic model that simulates the traffic flow and behavior around the specific gantry. This model should incorporate vehicle types, routes, and the time sensitivity of drivers.<br><br>5. Run simulations: Use the traffic model to simulate the | To advise the road transport planner effectively, it's essential to consider both quantitative and qualitative data when evaluating changes to an Electronic Road Pricing (ERP) scheme in Singapore. Here is a step-by-step approach:<br><br>1.      Data collection - Gather information about current traffic patterns, vehicle usage rates, road capacities at various ERP gantries, as well as public feedback on the pricing system and transport preferences. Use data from road sensors, CCTVs, and automated passenger counts (APCs).<br><br>2.      Simulation models - Implement traffic simulation models like MATSim, VISSIM, or AIMS simulation to analyze how a change in ERP pricing affects vehicular movement through the network. You can create hypothetical scenarios with modified price settings and observe their influence on traffic flow and congestion levels at various junction points.<br><br>3.      Pricing sensitivity analysis - Examine the elasticity of demand for road usage in response to pricing changes by studying historical data or utilizing statistical models such as regression analyses, autoregressive integrated moving average (ARIMA) modeling, or time series forecasting methods. This will |



| | Question | GPT-4 response | Phi-3 response |
|---|---|---|---|
| | | impact of different pricing schedules on the KPIs. Adjust the pricing schedules to see how they affect the desired outcomes, such as reducing congestion or increasing revenue.<br><br>6. Compare scenarios: Analyze the results of different pricing scenarios to determine the optimal pricing schedule that best achieves the objective.<br><br>7. Sensitivity analysis: Perform sensitivity analysis to understand the impact of changes in key assumptions, such as traffic growth rates, population growth, or infrastructure developments.<br><br>8. Stakeholder engagement: Consult with stakeholders, including government agencies, private sector entities, and the general public, to gather feedback on the proposed pricing change.<br><br>9. Monitor and evaluate: After implementing the new pricing schedule, monitor the traffic data and KPIs to evaluate the effectiveness of the change. Make adjustments as necessary based on real-world observations and feedback from stakeholders.<br><br>10. Document and report: Prepare a comprehensive report outlining the methodology, data analysis, findings, and recommendations. This report should be presented to relevant decision-makers and stakeholders for review | help estimate how sensitive drivers are to price increases and whether the change can drive significant behavioral adjustments in congestion avoidance strategies. rani's suggestions:<br>    •     Utilize microsimulation tools like VISSIM coupled with ERP demand modeling algorithms, which may provide more accurate results compared to aggregate models that oversimplify traffic behaviors and patterns. This can help you estimate the impact of varying prices at a specific gantry on both individual vehicle trips and overall network performance.<br>    •     Incorporate socio-economic factors such as income level, travel patterns (commuters vs. non-commuters), public transport accessibility, and residential areas affected by the change. This will give you a more comprehensive understanding of potential equity issues resulting from pricing changes while also considering social acceptance and possible policy implications.<br>    •     Explore dynamic congestion pricing schemes with time-of-day or zone variations as alternatives to adjusting only one gantry's prices, which could lead to a smoother distribution of the burden among drivers without causing unintended shifts in traffic volumes and patterns.<br><br>    4.     Pilot studies - Consider running controlled pilot tests at selected ERP gantries before implementing broader changes. By temporarily modifying pricing rates at a single or small cluster of gantries, you can collect real-time data on traveler |



| | Question | GPT-4 response | Phi-3 response |
|---|---|---|---|
| | | | response to the new prices, adjust strategies if necessary, and assess the feasibility of scaling up changes across multiple locations based on findings.<br><br>5.      Stakeholder consultations - Engage with government agencies (e.g., Singapore's Land Transport Authority), road user groups (e.g., taxi/public bus drivers, private hire vehicles), and community organizations to gauge their perspectives on potential pricing adjustments and better align changes with wider social needs and goals.<br><br>6.      Impact assessment - Based on the outcomes of simulations, analyses, pilot tests, and stakeholder input, evaluate how altering ERP rates may impact key performance indicators (KPIs) like traffic congestion levels, travel times, public transport ridership, air quality improvements, road infrastructure wear-and-tear costs, revenue generated from toll collections, as well as social acceptance and equity considerations.<br><br>7.      Cost-benefit analysis - Perform a cost-benefit evaluation considering both short-term implementation costs (e.g., technical adjustments) and long-term socioeconomic benefits related to traffic efficiency and quality of life enhancements for road users and nearby communities. This will help make an informed decision on the appropriate pricing schedule at one specific ERP gantry, while weighing against alternative measures or trade-offs. |



| | Question | GPT-4 response | Phi-3 response |
|---|---|---|---|
| | | | In conclusion, assessing a change in ERP pricing requires multidimensional analysis to account for both traffic and behavioral responses as well as social implications of such policy interventions. By systematically employing diverse tools (simulation models, analyses), data sources (real-world observations, surveys), stakeholder engagements, and pilot tests in combination with broader evaluations, the planner can gain insights on the potential impacts and make informed decisions to improve traffic conditions and urban living quality effectively. |
| 2 | Describe to the planner the fields in the historical dataset which will be needed for the analysis. | To conduct a thorough analysis of the impact of increasing the price at one of the ERP gantries, the following historical dataset fields would be useful:<br><br>1. Gantry ID: A unique identifier for each ERP gantry.<br>2. Timestamp: Date and time when the vehicle passed the gantry.<br>3. Vehicle Type: The category of the vehicle, such as car, truck, bus, or motorcycle.<br>4. Vehicle Registration Number: Unique identifier for each vehicle.<br>5. Current Toll Charge: The toll amount charged at the gantry during that specific time.<br>6. Payment Method: The method used by the vehicle owner to pay the toll, such as card payment, mobile wallet, or cash payment.<br>7. Traffic Volume: The number of vehicles that passed through the gantry during specific time intervals, such as hourly or daily.<br>8. Traffic Speed: The average speed of vehicles passing through the gantry during specific time intervals. | To analyze the impact of increasing the price at an Electronic Road Pricing (ERP) gantry, you'll need a comprehensive historical dataset with several key components:<br><br>1. **Timestamp Data**: Timestamped entries showing when each vehicle passed over a particular ERP gantry. This allows tracking traffic flow patterns and peak congestion times throughout different days, weeks, and seasons.<br>2. **Toll Prices (Historical)**: A record of past prices charged at various ERP zones/gantries to determine any pre-existing pricing trends or the effect of price changes on traffic volumes.<br>3. **Vehicle Type Information**: Data categorizing vehicles by type (car, bus, taxi), size, and potentially load which can help in assessing how different types are affected by tolls differently.<br>4. **Aggregate Traffic Volume**: Monthly or weekly total vehicle counts at specific gantries before any |



| | Question | GPT-4 response | Phi-3 response |
|---|---|---|---|
| | | 9. Congestion Level: An indicator that represents the level of congestion at the gantry during specific time intervals, such as low, medium, or high.<br>10. Weather Conditions: Information on the weather at the time of the vehicle passing the gantry, such as temperature, precipitation, and visibility.<br>11. Special Events: Any events that may have impacted the traffic flow, such as road accidents, construction work, or public events.<br><br>These fields will help the planner analyze the current traffic patterns and congestion levels at the ERP gantry in question, as well as evaluate the possible effects of increasing the price on traffic volume, congestion, and user behavior. | pricing change as a base for comparison with post-implementation figures.<br>5. **Traffic Direction Flow (North/South, East/West)**: To determine the asymmetry of traffic and identify congestion hotspots that may require different management strategies in each direction.<br>6. **Duration at Gantry**: How long vehicles stay at a gantry could be indicative of queue lengths or possible delays, reflecting on the overall efficiency of toll collection processes.<br>7. **Toll Payment Method**: Information about whether payments were made electronically (with RFIDs) versus cash can help assess user adoption and technological efficacy.<br>8. **Average Speed Data**: Vehicles' average speeds across different gantries to observe slowdowns associated with congestion, potentially correlated to toll charges at specific locations.<br>9. **Penalty Rates**: Data on instances where penalties were issued for not complying with the ERP rules (such as overdue payments) could provide insights into system effectiveness and adherence by road users.<br>10. **Weather Conditions**: This can account for variations in traffic volume unrelated to pricing, such as during heavy rainfall or other adverse conditions. |
| 3 | Describe the steps to estimate the pricing | To estimate the pricing elasticity coefficient for a road transport planner analyzing the impact of increasing the price at one of the ERP gantries in Singapore, follow these steps: | To estimate the pricing elasticity coefficient for a specific Electronic Road Pricing (ERP) gantry in Singapore, you can follow these steps: |



| | Question | GPT-4 response | Phi-3 response |
|---|---|---|---|
| | elasticity coefficient. | 1. Collect historical data: Gather data on traffic volumes, toll rates, and other relevant factors (e.g. economic indicators, fuel prices, population) for a suitable period before and after previous price changes at the ERP gantry in question. This will provide a basis for understanding how demand has responded to price changes in the past.<br><br>2. Identify comparable scenarios: Identify situations in the historical data where price increases at the ERP gantry led to changes in traffic volumes. Consider factors such as the time of day, day of the week, and the magnitude of the price increase to ensure the scenarios are as comparable as possible.<br><br>3. Calculate percentage changes: For each comparable scenario, calculate the percentage change in price and the percentage change in traffic volume. The percentage change in price is equal to the difference between the new price and the old price, divided by the old price, multiplied by 100. Similarly, the percentage change in traffic volume is equal to the difference between the new traffic volume and the old traffic volume, divided by the old traffic volume, multiplied by 100.<br><br>4. Calculate price elasticity of demand: Price elasticity of demand (PED) measures the responsiveness of quantity demanded to a change in price. To estimate the pricing elasticity coefficient, divide the percentage change in traffic volume by the | 1. Define your research objective and scope: Start by clearly stating the purpose of estimating the price elasticity of demand at the ERP gantry under study, such as determining the expected change in road usage or traffic volume upon increasing toll prices at this gantry. Also, define the period for which you wish to analyze historical data (e.g., daily, weekly, monthly).<br><br>2. Gather relevant data: Collect past and present data on ERP toll rates as well as corresponding road usage/traffic volume measurements from before and after each rate adjustment at this particular gantry. You will also need external factors influencing traffic volumes such as weather conditions or special events, which could affect your analysis.<br><br>3. Preprocess the data: Clean up and process raw data into a more usable format by dealing with missing values, outliers, or other anomalies.<br><br>4. Construct elasticity measures: To estimate price elasticity of demand at the ERP gantry, you need to calculate an appropriate metric such as the Arc Elasticity formula for point changes in price and quantity (i.e., percent change) between two periods A and B. The formula is given by:<br><br>Price Elasticity = $(Q2 - Q1)/((Q2+Q1)/2))/((P2-P1)/((P2 + P1)/2))$ |



| | Question | GPT-4 response | Phi-3 response |
|---|---|---|---|
| | | percentage change in price for each comparable scenario. This will provide you with the PED for each scenario.<br><br>5. Average the coefficients: Calculate the average of the PED coefficients obtained from the comparable scenarios. This will provide you with an estimate of the overall price elasticity coefficient for the ERP gantry in question.<br><br>6. Interpret the elasticity coefficient: If the elasticity coefficient is greater than 1, demand is considered elastic, meaning that a price increase would lead to a proportionally larger decrease in traffic volume. If the coefficient is less than 1, demand is considered inelastic, indicating that a price increase would result in a proportionally smaller decrease in traffic volume. If the coefficient is equal to 1, demand is considered unit elastic, which means that a price increase would lead to an equal proportional decrease in traffic volume.<br><br>7. Apply the coefficient to the proposed price change: To estimate the impact of the proposed price increase on traffic volume, multiply the percentage increase in price by the estimated pricing elasticity coefficient. This will give you the expected percentage change in traffic volume.<br><br>8. Assess the results: Evaluate the estimated change in traffic volume in the context of the planner's objectives (e.g. reducing congestion, raising | Where, Q1 and P1 are road usage volume and price respectively for period A, and Q2 and P2 the corresponding values for period B.<br><br>5. Calculate average elasticity over time: You may also want to compute the rolling elasticity by calculating it across various historical periods (e.g., quarterly). This allows you to estimate how the relationship between price changes and traffic volume has evolved during the selected periods of study.<br><br>6. Control external factors: Include control variables such as weather, time of day or day of week, holidays, economic activity indices, or major events in a regression analysis framework with demand as an outcome variable. This will help to account for external effects on traffic volume that may distort the results from your elasticity estimates.<br><br>7. Interpret and apply the findings: Assess the price sensitivity of road usage based on calculated elasticity coefficients, considering their statistical significance. You can now use this information in predictive models to forecast traffic volume responses under different pricing scenarios or propose an optimal congestion pricing scheme for managing demand effectively at that particular gantry.<br><br>8. Monitor and update: Once your estimated price elasticity is implemented as part of a new ERP strategy, it's essential to continue monitoring |



| | Question | GPT-4 response | Phi-3 response |
|---|---|---|---|
| | | revenue) and consider any potential unintended consequences (e.g. traffic diversion, social equity concerns). This information will help the planner to make an informed decision about the suitability of the proposed price increase at the ERP gantry. | data collection on both prices and traffic volumes. This will help you keep track of how the situation changes over time, updating your estimates if required for better informed decision-making in transportation planning.<br><br>Keep in mind that estimating elasticity coefficients is an iterative process with various models available based on different approaches (e.g., time series analysis, regression analysis), which should be evaluated against the quality and quantity of data available. |
| 4 | Gather appropriate pricing elasticity coefficients from published studies, to serve as references. | After conducting research on the pricing elasticity coefficients for road pricing, I found several studies that may serve as references for your road transport planner:<br><br>1. Small, K.A., and Verhoef, E.T. (2007). "The Economics of Urban Transportation." This study estimated the average price elasticity for urban road pricing to be between -0.1 and -0.3.<br><br>2. Li, Z., and Hensher, D.A. (2010). "Congestion Charging and Car Use: A Review of Stated Preference and Opinion Studies and Market Monitoring Evidence." Transport Policy, 17(6), 345-353. This review indicated that the short-run elasticity of demand ranges from -0.15 to -0.45, while the long-run elasticity is between -0.22 and -0.9.<br><br>3. Santos, G., and Shaffer, B. (2004). "Preliminary Results of the London Congestion Charging Scheme." Public Works Management & Policy, 9(2), 164-181. This study reported an initial price elasticity of -0.17 | Certainly! To help the road transport planner with their task regarding congestion pricing management in Singapore using Electronic Road Pricing (ERP) gantries, I have gathered some relevant pricing elasticity coefficients from various studies and reports that can serve as references for evaluating changes to ERP prices:<br><br>Lee, H.-H., & Peng, D. (2016). The effects of congestion pricing in Singapore on private vehicle traffic demand and revenue. Transport Policy, 57(C), 43-48. This study evaluates the impacts of the ERP system implemented by examining elasticity coefficients between road traffic volumes, trip rates and prices at different time periods following its implementation.<br><br>Chan, T.Y., Chin, K.K., & Goh, S.H. (2015). The effectiveness of congestion pricing: evidence from Singapore's ERP system. Transport Reviews, 35(4), 454-469. In this paper, the authors discuss price elasticities of demand for travel under varying conditions and |



| | Question | GPT-4 response | Phi-3 response |
|---|---|---|---|
| | | after the implementation of the London Congestion Charge.<br><br>4. Olszewski, P., and Xie, L. (2005). "Modelling the Effects of Road Pricing on Traffic in Singapore." Transportation Research Part A: Policy and Practice, 39(7), 755-770. This study specifically focused on Singapore's ERP system and found a demand elasticity of -0.14 for cars and -0.24 for taxis.<br><br>Please note that these values are only approximations and may vary depending on factors such as local conditions, user preferences, and alternative transport options available. It is essential to consider these factors when applying these coefficients to specific ERP gantries in Singapore. | analyze how these values help understand the potential impact of changes in ERP rates on road traffic volumes and congestion levels.<br><br>Zeng, C., Chin, K.K., Goh, S.H., & Ting, M.M. (2 Writers often use various types of rhetoric to achieve different purposes, including the emotional appeal known as pathos. In literary criticism and analysis, identifying examples where writers utilize pathos can help us understand their persuasive strategies, connect with the text emotionally, and appreciate how they craft narratives that resonate with readers' feelings and experiences. Here are several examples of authors employing pathos to influence the reader:<br><br>William Shakespeare - "Romeo and Juliet" (Excerpt from Act 5, Scene 3): Upon seeing her son dead, Lady Capulet expresses intense grief in a powerful display of emotions designed to move readers' hearts and underscore the theme of tragedy that unfolds through their forbidden love.<br><br>"My children are expiring! O, brother, where are you?" (Lady Capulet)<br><br>Charles Dickens - "A Tale of Two Cities" (Chapter 8): Through the heart-wrenching account of Madame Defarge and her vengeful knitting, Dickens evokes a strong sense of pathos to highlight themes like revenge, justice, and sacrifice in revolutionary France. |



| | Question | GPT-4 response | Phi-3 response |
|---|---|---|---|
| | | | "There were scores upon scores whom I knew too well...And who knows not how many more." (Madame Defarge)<br><br>Harriet Beecher Stowe - "Uncle Tom's Cabin" (Chapter 21): Stowe elicits powerful emotion to criticize the cruelty of slavery and appeal for abolitionist sentiments among her readers. She employs pathos through vivid, emotional scenes involving the character Uncle Tom as a means to arouse sympathy and support against injustice.<br><br>"They were so pitiful! Such soft unhealthy faces... such vacant eyes!" (Tom)<br><br>John Steinbeck - "The Grapes of Wrath" (Chapter 7): The heartfelt conversation between Tom Joad and Mae as they confront the hardships of the Great Depression is laden with empathetic pathos, inviting readers to experience their despair while recognizing shared human experiences during tough times.<br><br>"I'm going," said Mae... "Tom, what can you do? It don't matter if we get killed." (Mae)<br><br>Leo Tolstoy - "War and Peace" (Part 1, Chapter 4): By vividly depicting the terror and devastation of war through Prince Andrei Bolkonsky's eyes, Tolstoy effectively evokes pathos to highlight the universal pain and suffering that individuals endure during times of conflict. |



| | Question | GPT-4 response | Phi-3 response |
|---|---|---|---|
| | | | "In his heart was an agony...that in him grew from nothing into a vast abyss." (Prince Andrei)<br><br>These examples demonstrate how various authors use pathos as a means to deepen their narrative impact, evoke emotional responses, and persuade readers to engage with critical themes. |
| 5 | Using the elasticity coefficients, describe how to estimate the impact of traffic volume on changes to the pricing schedule at a specific gantry. | To estimate the impact of traffic volume on changes to the pricing schedule at a specific gantry, we can use the concept of price elasticity of demand (PED). Price elasticity of demand measures the responsiveness of the quantity demanded of a good (in this case, the use of the road) to a change in its price.<br><br>Here's how to use the elasticity coefficients to estimate the impact:<br><br>1. Determine the price elasticity of demand (PED) for the specific gantry: This can be calculated using historical data on traffic volume and pricing changes for the specific gantry. It is important to note that the PED varies across different gantries depending on factors such as location, time of day, and availability of alternative routes.<br><br>PED = (% change in quantity demanded) / (% change in price)<br><br>2. Estimate the percentage change in price: Calculate the percentage change in price resulting from the proposed increase in the price at the specific ERP gantry. | To estimate the impact of traffic volume changes on pricing using the concept of price elasticity of demand and its corresponding coefficient, one would perform the following steps:<br><br>1. **Data Collection**: Obtain historical data on the number of vehicles passing through the chosen ERP gantry, as well as any existing congestion or road conditions data over a period (typically months). This dataset should also include periods with different toll pricing levels for comparability.<br><br>2. **Elasticity Estimation**: Calculate the price elasticity of demand coefficient for that particular ERP gantry by using the formula:<br><br>$$\text{Price Elasticity (PE)} = \frac{\%\Delta \text{Traffic Volume}}{\%\Delta \text{Toll Price}}$$<br><br>where:<br><br>- $\%Δ\text{Traffic Volume}$ is the percentage change in traffic volume between two different pricing periods, and |



| | Question | GPT-4 response | Phi-3 response |
|---|---|---|---|
| | | Percentage change in price = (New price - Old price) / Old price<br><br>3. Calculate the expected percentage change in quantity demanded: Multiply the price elasticity of demand by the percentage change in price.<br><br>% change in quantity demanded = PED * % change in price<br><br>4. Estimate the new traffic volume: Apply the expected percentage change in quantity demanded to the current traffic volume to estimate the new traffic volume.<br><br>New traffic volume = Current traffic volume * (1 - % change in quantity demanded)<br><br>By following these steps, you can estimate the impact of the increased price at the specific ERP gantry on the traffic volume. Keep in mind that these estimates are based on the assumption that other factors influencing traffic volume remain constant. In reality, factors such as economic conditions, public transportation options, and road infrastructure developments can also affect traffic volume. | - \(\%Î"Toll Price\) is the percentage change in toll prices corresponding to these periods.<br><br>3. **Analysis**: With a calculated elasticity coefficient (PE), analyze how sensitive the traffic volume is to changes in toll prices. If the absolute value of PE > 1, demand is considered elastic and significant adjustments can be expected from slight toll price changes; if PE < 1, it's inelastic with less impact on volumes due to pricing variations.<br><br>4. **Impact Estimation**: Using the elasticity coefficient and estimated traffic volume, predict how altering the ERP gantry toll would influence congestion levels and road usage by recalculating expected vehicle flows based on various hypothetical changes in toll prices.<br><br>5. **Validation**: To ensure accuracy, compare model predictions with actual observed traffic patterns after implementing |
| 6 | Walk the planner through on how to estimate the changes to the travel | To estimate the changes in travel time or speed as a result of changes in road pricing at the gantry, the road transport planner can follow these steps:<br><br>1. Data Collection: Collect data on the current traffic situation, including traffic volumes, vehicle speeds, travel times, and the existing congestion pricing at the | To estimate the changes to travel time and speed resulting from price adjustments at an ERP gantry, you can follow these steps:<br><br>1. Collect Traffic Data: Gather historical traffic volume data for vehicles passing through the affected gantry, including different times of day or weekdays vs. |



| | Question | GPT-4 response | Phi-3 response |
|---|---|---|---|
| | time or speed as a result of the changes in road pricing at the gantry. | selected ERP gantry.<br><br>2. Traffic Demand Analysis: Analyze the traffic demand, including the elasticity of demand for the specific road or area. Elasticity indicates how sensitive the demand is to price changes. This can be done through historical data analysis or travel demand modeling.<br><br>3. Develop Traffic Scenarios: Develop different traffic scenarios with varying levels of price increases at the ERP gantry. For each scenario, estimate the change in traffic volume based on the traffic demand elasticity.<br><br>4. Traffic Flow Modeling: Use a traffic flow model or simulation software to analyze the impact of the change in traffic volume on travel time and speed. Commonly used traffic models include macroscopic (flow-based) models, microscopic (vehicle-based) models, and mesoscopic (combination of both) models.<br><br>5. Compare Scenarios: Compare the results from the different traffic scenarios to evaluate the impact of the price increase on travel time and speed. The comparison should include measures such as average travel time, average speed, and the level of service (LOS) of the road network.<br><br>6. Sensitivity Analysis: Conduct a sensitivity analysis to test the impact of different factors, such as the time of day, day of the week, weather conditions, and other external factors that may influence the traffic flow. | weekends.<br><br>2. Understand Traveler Responses: Analyze how changes in road pricing previously affected travelers' behaviors to predict potential shifts in demand when prices are altered. You can look at similar cases where ERP schemes were changed and observe the traffic patterns during that time period.<br><br>3. Model Elasticity of Demand: Utilize elasticities for travel behavior by studying how sensitive users react to road pricing changes, considering factors such as socioeconomic profiles, vehicle types, alternative routes, etc.<br><br>4. Traffic Flow Simulation: Implement traffic flow models and simulate various scenarios with altered pricing at the gantry using software like VISSIM or TRANSIMM, which account for individual driver behaviors based on demand elasticity estimates. These simulations can help predict changes in vehicle speed and travel time under different scenarios of price increase/decrease.<br><br>5. Analyze Simulation Results: Examine the outcomes from the traffic flow simulation to quantify the estimated changes to average speeds, volumes passing through the gantry, and total trip times across various origin-destination pairs.<br><br>6. Monitor Implementation: Once pricing adjustments are implemented, closely monitor real-time traffic data |



| | Question | GPT-4 response | Phi-3 response |
|---|---|---|---|
| | | 7. Identify Optimal Pricing Strategy: Based on the results of the analysis, identify the optimal pricing strategy that achieves the desired balance between reducing congestion and maintaining an acceptable level of service for road users.<br><br>8. Stakeholder Engagement: Engage with key stakeholders, such as local government, businesses, and public transport operators, to gather feedback and insights on the proposed pricing changes.<br><br>9. Implementation and Monitoring: If the decision is | to validate predictions against observed behavioral responses from users in terms of travel time and speed. |
| 7 | Outline a report that the planner can be generate based on the analysis. | Title: Analysis of the Impact of Increased ERP Gantry Pricing on Congestion Management in Singapore<br><br>Executive Summary:<br>- Brief overview of the aim of the study<br>- Summary of the main findings<br>- Recommendations based on the analysis<br><br>1. Introduction<br>- Background on Singapore's Electronic Road Pricing (ERP) system<br>- Purpose of the study: exploring the impact of increasing the price at one of the ERP gantries on congestion management<br>- Scope of the study: focus on a specific ERP gantry and surrounding areas<br><br>2. Methodology<br>- Data collection: traffic counts, average speed, travel time, and congestion levels before and after the price | Title: Impact Analysis of Increasing Electronic Road Pricing (ERP) Gantry Prices in Singapore's Zonal-Based Congestion Management Strategy<br><br>1. **Executive Summary**<br>  - Briefly describe the purpose of analyzing price increase implications, mentioning the chosen ERP gantry for pricing adjustment.<br><br>2. **Introduction**<br>  - Contextualize Singapore's zonal-based congestion pricing system.<br>  - Outline the current ERP scheme and its role in managing traffic demand.<br>  - Justify the selection of the specific ERP gantry for this analysis.<br><br>3. **Methodology**<br>  - Detail data sources: historical traffic counts, toll records, elasticities from previous studies, etc. |



| | Question | GPT-4 response | Phi-3 response |
|---|---|---|---|
| | | increase<br>- Analytical approach: comparison of data pre- and post-price increase, statistical analysis of the impact on congestion management<br>- Limitations: potential external factors affecting the results, the short-term nature of the study<br><br>3. Findings<br>- Change in traffic volume: comparison of traffic counts before and after the price increase<br>- Congestion levels: analysis of average speed, travel time, and congestion levels pre- and post-price increase<br>- Revenue generated from the increased price: comparison of the total revenue collected before and after the price increase<br>- Impact on alternative routes: assessment of potential traffic spillover to nearby roads and alternative routes<br><br>4. Discussion<br>- Effectiveness of the price increase: evaluation of the impact of the price increase on congestion management<br>- Possible reasons for the observed changes: factors contributing to the changes in traffic volume and congestion levels<br>- Implications for future pricing strategies: lessons learned from the study and suggestions for possible adjustments in pricing strategies<br><br>5. Recommendations<br>- Suggested adjustments to the pricing strategy: based on the findings and discussion, recommendations on whether to continue with the price increase, revert to the | - Describe modeling approaches used (e.g., discrete choice models, time series forecasting).<br><br>4. **Current Traffic Dynamics and Pricing Levels**<br>  - Present the existing data on traffic flow around the selected ERP gantry.<br>  - Explain the current price levels at different zones.<br><br>5. **Projected Impact Analysis**<br>  - Simulate increased pricing scenarios using transport models, showing expected changes in vehicle flows and modal splits.<br>  - Discuss potential effects on peak/off-peak variations in traffic volume.<br><br>6. **Economic Assessment**<br>  - Estimate the financial implications for commuters (total costs) with increased ERP pricing.<br>  - Calculate projected revenue changes, considering both elasticities and potential shifts to off-peak periods or alternative modes.<br><br>7. **Socioeconomic Impact**<br>  - Analyze possible socioeconomic effects such as equity considerations across different income |



| | Question | GPT-4 response | Phi-3 response |
|---|---|---|---|
| | | old price, or consider an alternative pricing strategy<br>- Potential improvements to the ERP system: recommendations on how to enhance the effectiveness of the ERP system in managing congestion<br>- Further research needs: identification of | |



# Appendix F
## Agent-based Modelling Approach

A system prompt is first provided to the Test LLMs to set up the context:

You are going to act as an assistant to a road transport planner, who is looking at congestion pricing management in Singapore. Singapore has 90+ road toll stations called Electronic Road Pricing gantries as part of its zonal-based pricing scheme. The planner is exploring the impact of increasing the price at one of the ERP gantries.

He has access to the following data set. This dataset is an output of a multi-agent transport simulation generated using MATSim, informed by DataSpark's proprietary mobile cellular network datasets. It contains detailed travel information for 9,668 agents and includes various attributes that describe the travel behavior and patterns of these agents. Below is a detailed description of the dataset's structure and the variables it includes:

1      agent_id: A unique identifier for each agent (individual traveler) in the simulation. Example values include ""u_1"", ""u_100127"", ""u_100142"".

2      leg_mode: The mode of transportation used for the trip segment. Example values include ""car"".

3      dep_time: The departure time for the trip segment, formatted as HH:MM:SS. Example values include ""21:28:18"", ""7:38:01"".

4      trav_time: The travel time for the trip segment, formatted as HH:MM:SS. Example values include ""0:12:40"", ""0:26:47"".

5      origin_x: The x-coordinate of the origin location in a coordinate system. Example values include 378514, 378712.

6      origin_y: The y-coordinate of the origin location in a coordinate system. Example values include 154515, 154291.

7      destination_x: The x-coordinate of the destination location in a coordinate system. Example values include 375318, 363929.

8      destination_y: The y-coordinate of the destination location in a coordinate system. Example values include 145179, 147466.

9      origin_subzone: The name of the subzone where the trip originates. Example values include ""PUNGGOL FIELD"", ""PASIR RIS WEST"".

10     destination_subzone: The name of the subzone where the trip ends. Example values include ""ALJUNIED"", ""ANAK BUKIT"".

11     route: A textual representation of the route taken by the agent, including nodes or segments of the network traversed. Example values are truncated but include sequences of numbers representing different segments.

12     goes_gantry60: A boolean variable indicating whether the route passes through a specific gantry (Gantry 60). Example values are TRUE.

13     route_length: The length of the route taken, measured in meters. Example values include 14323, 25465.



14       geometry: Spatial geometry information in the form of MULTILINESTRING objects, representing the physical route taken by the agent. This is stored in a spatial format compatible with Geographic Information Systems (GIS).

The dataset provides a comprehensive view of simulated travel patterns, including spatial and temporal details, which are critical for transport analysis and infrastructure planning. The integration of this data with MATSim allows for dynamic simulation and validation against real-world traffic and public transport data, ensuring the accuracy and reliability of the simulated travel behaviors.

You are to guide the planner through how to leverage this dataset through the following steps:
- Validate the quality of the data, ensuring it gives the planner the confidence to use it
- Filter out the relevant trips in the dataset that go through or would have gone through Gantry60
- Derive the generalised travel costs for the major relevant routes
- Assess how changes in toll prices affect the distribution of trips to the different routes
- Distil the key findings of analysis in a policy recommendation paper

A series of questions will be posed subsequent to this context. Do not respond now.

| Question | GPT-4 response | Phi-3 response |
|---|---|---|
| Provide R script to detect common data issues in the MATSim generated network.xml file. | To check for common data issues such as NA, outliers, etc., we can use the following R scripts:<br><br>1. Load the required libraries:<br>library(dplyr)<br>library(tidyr)<br>library(ggplot2)<br>library(sf)<br>2. Load the dataset (assuming it's a CSV file called "geospatial_data.csv"):<br>data <- read.csv("geospatial_data.csv") | Checking NA values in agent_id, dep_time, trav_time, origin_x, origin_y, destination_x, and destination_y variables<br>library(dplyr)<br><br>data <- read.csv("yourfilepath/dataset.csv")  # replace with the path to your dataset file<br><br># Checking for NA values in the specified columns<br>na_values_check <- data %>% summarise_all(funs(sum(!is.na(.)))) |



| Question | GPT-4 response | Phi-3 response |
|---|---|---|
| | 3. Check for NA values: | print(na_values_check) |
| | missing_values <- data %>% is.na() %>% colSums() | Identifying outliers using IQR method on trav_time and route_length variables |
| | print(missing_values) | trav_time <- data$trav_time |
| | 4. Convert travel time and departure time to seconds for easier analysis: | route_length <- data$route_length |
| | data$dep_time_seconds <- as.numeric(strptime(data$dep_time, format="%H:%M:%S")) - as.numeric(strptime("0:00:00", format="%H:%M:%S")) | q1 <- quantile(travranks_time, 0.25) |
| | | q3 <- quantile(travranks_time, 0.75) |
| | | iqr_value <- q3 - q1 |
| | data$trav_time_seconds <- as.numeric(strptime(data$trav_time, format="%H:%M:%S")) - as.numeric(strptime("0:00:00", format="%H:%M:%S")) | outliers_trav <- trav_time[which((trav_time < (q1 - iqr_value * 1.5)) | |
| | | (trav_time > (q3 + iqr_value * 1.5)))) |
| | 5. Check for outliers in numerical variables (i.e., route_length, dep_time_seconds, trav_time_seconds): | print(paste("Outliers in travel time:", outliers_trav)) |
| | boxplot(data[, c("route_length", "dep_time_seconds", "trav_time_seconds")], main="Boxplots for Numerical Variables") | # Calculate IQR and identify outliers for route_length |
| | | iqr_route_len <- quantile(route_length, probs = c(0.25, 0.75))[2] - iqr_value |
| | 6. Check for unusual or invalid values in factor variables: | outliers_route_len <- route_length[which((route_length < (iqr_route_len * 1.5) & route_length > (-iqr_route_len * 1.5))] |
| | table(data$leg_mode) | print(paste("Outliers in route length:", outliers_route_len)) |
| | table(data$origin_subzone) | |
| | table(data$destination_subzone) | |
| | table(data$goes_gantry60) | |
| | 7. Load the geometry information as a spatial object: | Checking for incorrect geometry format (MULTILINESTRING objects) using sf library and ensuring valid spatial data types |
| | data_sf <- st_as_sf(data, wkt="geometry") | |
| | 8. Check for invalid geometries: | |



| Question | GPT-4 response | Phi-3 response |
|---|---|---|
| | invalid_geometries <- sum(!st_is_valid(data_sf))<br>print(invalid_geometries)<br><br>9.  Visualize the spatial data:<br>ggplot() +<br>  geom_sf(data = data_sf, aes(colour = leg_mode)) +<br>  theme_minimal() +<br>  labs(title = "Spatial Representation of Travel Routes", x = "X-Coordinate", y = "Y-Coordinate")<br><br>With these R scripts, you can check for common data issues in the geospatial tabular dataset. You may need to adjust the scripts based on the actual data format and the specific issues you want to investigate. | library(sf)<br><br>geometry <- data$geometry<br><br>valid_multilinestrings <- sapply(geometry, function(g) {<br>    class(g) == "MultilineString" && is.null(class(as("character"))))<br>    })<br>invalid_ids <- which(!valid_multilinestrings)<br>if (!is.logical(invalid_ids)) {<br>    print('Invalid geometries found:')<br>    print(invalid_ids)<br>} |
| Write a R script with explanations to analyze the departure times and travel times, and to visualise the traffic volumes.  Is the distribution of traffic volumes expected? | Sure, here's an R script with explanations to analyze the departure times, travel times, and visualize the traffic volumes:<br><br># Load necessary libraries<br>library(dplyr)<br>library(ggplot2)<br>library(lubridate)<br><br># Read the dataset<br>data <- read.csv("data.csv")<br><br># Convert dep_time and trav_time to time objects<br>data$dep_time <- hms(data$dep_time) | *# Load necessary libraries for data manipulation, visualization and GIS functionalities*<br><br>library(dplyr)<br>library(ggplot2)<br>library(sf)<br>library(tidyr)<br><br>*# Read the transport simulation data into R as a dataframe*<br><br>transport_data <- read.csv("path/to/transport_simulation_dataset.csv") |



| Question | GPT-4 response | Phi-3 response |
|---|---|---|
| | data$trav_time <-<br>as.numeric(hms(data$trav_time))<br><br>  # Calculate the arrival time (dep_time +<br>trav_time)<br>  data$arr_time <- data$dep_time +<br>data$trav_time<br><br>  # Analyze the departure times<br>  dep_time_summary <- data %>%<br>    group_by(hour(dep_time)) %>%<br>    summarise(agents = n())<br><br>  # Visualize the departure times<br>  ggplot(dep_time_summary, aes(x =<br>`hour(dep_time)`, y = agents)) +<br>    geom_bar(stat = "identity") +<br>    labs(title = "Departure Times",<br>      x = "Hour of the Day",<br>      y = "Number of Agents") +<br>    theme_minimal()<br><br>  # Analyze the travel times<br>  trav_time_summary <- data %>%<br>    summarise(mean_trav_time =<br>mean(trav_time),<br>              median_trav_time = median(trav_time),<br>              min_trav_time = min(trav_time),<br>              max_trav_time = max(trav_time))<br><br>  # Visualize the traffic volumes (count of agents<br>by hour of the day) | *# Ensure we have the correct column types*<br>*(numeric for travel time, POSIXct for departure time)*<br>  sapply(c('trav_time', 'dep_time'), function(col) {<br>    transport_data[[col]] =<br>as.POSIXct(strptime(transportitmation_data[[col]],<br>format="%H:%M:%S"))<br>  })<br><br>  *# Filter out the relevant trips that pass through*<br>*gantry50, assuming `origin_subzone` and*<br>*`destination_subzone` contain zone codes for Gantry50*<br>*area (e.g., "ZONE_G50")*<br><br>  relevant_trips <- transport_data %>%<br>    filter(origin_subzone == "ZONE_G50",<br>destination_subzone == "ZONE_G50")<br><br>  *# Calculate travel cost assuming a flat rate for*<br>*simplicity (actual calculation would depend on multiple*<br>*factors like distance and road conditions)*<br><br>  relevant_trips <- relevant_trips %>%<br>    mutate(generalized_travel_cost = trav_time *<br>fixed_rate_per_minute) # `fixed_rate_per_minute`<br>*should be replaced with actual value*<br><br><br>  *# Visualizing traffic volume distribution across*<br>*different hours of the day using a histogram for*<br>*departure times* |



| Question | GPT-4 response | Phi-3 response |
|---|---|---|
| | ```
traffic_volume <- data %>%
  group_by(hour(dep_time)) %>%
  summarise(agents = n())

ggplot(traffic_volume, aes(x =
`hour(dep_time)`, y = agents)) +
  geom_line() +
  labs(title = "Traffic Volumes",
       x = "Hour of the Day",
       y = "Number of Agents") +
  theme_minimal()
```

The script starts by loading the necessary libraries and reading the dataset. It then converts the dep_time and trav_time columns to time objects and calculates the arrival time. After that, the script analyzes and visualizes the departure times and travel times. Finally, it visualizes the traffic volumes by the hour of the day.

If the distribution of traffic volumes follows a typical pattern where there are peak hours during the morning and evening, then it can be considered expected. However, if the traffic volumes remain relatively constant throughout the day or have unexpected peaks, it may require further investigation into the underlying factors. | ```
traffic_volume_distribution <- relevant_trips
%>%
          count(hour(dep_time), name = "Volume")
%>%
    ggplot() +
    geom_histogram(aes(x=hour, y=Volume),
binwidth = 1) +
        labs(title="Traffic Volume Distribution Across
Different Hours", x="Hour of Departure", y="Number
of Trips")
```

traffic_volume_distribution

The visualization provided above should give you a distribution plot that indicates the expected pattern for traffic volumes, which often peak during morning and evening hours. To confirm whether this distribution aligns with real-world observations, actual historical data or ongoing monitoring of Gantry50 can be compared against these findings. |
| Provide a R script to extract trips that go through Gantry | To extract trips that go through Gantry 50 and identify potential alternative routes, you can use the following R | To find the trips that go through Gantry 50, we can use the dplyr package in R for data manipulation. First, ensure you have these packages installed: |



| Question | GPT-4 response | Phi-3 response |
|---|---|---|
| 50. Also Provide a R script to identify trips with origin and destination pairs similar to those that pass through Gantry 50 but do not actually go through it, indicating potential alternative routes. | scripts. First, load the necessary libraries and read the dataset.<br><br>```<br># Load necessary libraries<br>library(sf)<br>library(dplyr)<br><br># Read the dataset<br>dataset <- read.csv("path/to/your/dataset.csv")<br><br># Convert dataset to an sf object<br>sf_dataset <- st_as_sf(dataset, coords = c("origin_x", "origin_y", "destination_x", "destination_y"))<br><br># Read Gantry 50 coordinates<br>gantry50 <- read.csv("path/to/gantry50_coordinates.csv")<br><br># Convert Gantry 50 coordinates to an sf object<br>sf_gantry50 <- st_as_sf(gantry50, coords = c("x", "y"))<br>```<br><br>Now, create a buffer around Gantry 50 to identify trips that pass through it.<br><br>```<br># Create a buffer around Gantry 50<br>buffer_gantry50 <- st_buffer(sf_gantry50, dist = 100) # Adjust the buffer distance (100 meters) as needed<br>``` | ```<br>install.packages("dplyr")<br>```<br><br>Here is an example of how you might extract these trips using a dataset (we will simulate it since we don't have access to your specific data):<br><br>```<br>library(dplyr)<br><br># Assume 'travel_data' contains our simulation results, and the gantry_coord column stores coordinates for each gantry<br>gantry50_coord <- c(-1.286479, 103.823414) # Coordinates of Gantry 50<br><br># Extracting trips through Gantry 50 by checking if the origin and destination coordinates intersect with gantry_coord<br>trips_through_gantry50 <- travel_data %>%<br>  filter(origin_x %in% range(gantry50_coord[1], gantry50_coord[2]) &<br>         origin_y %in% range(gantry50_coord[1], gantry50_coord[2])) |><br>  filter(destination_x %in% range(gantry50_coord[1], gantry50_coord[2]) &<br>         destination_y %in% range(gantry50_coord[1], gantry50_coord[2]))<br>```<br><br>For finding trips with origin and destination pairs similar to those that pass through Gantry 50 but do not go through it: |



| Question | GPT-4 response | Phi-3 response |
|---|---|---|
| | Next, extract trips that go through Gantry 50 and trips with similar origin and destination pairs that do not go through Gantry 50.<br><br>   # Extract trips that go through Gantry 50<br>trips_through_gantry50 <- sf_dataset %>%<br>  filter(st_intersects(geometry, buffer_gantry50))<br><br>   # Identify potential alternative routes<br>alt_routes <- sf_dataset %>%<br>  group_by(origin_subzone,<br>destination_subzone) %>%<br>  filter(!agent_id %in%<br>trips_through_gantry50$agent_id) %>%<br>  ungroup()<br><br>   Now you have two datasets: trips_through_gantry50 containing the trips that go through Gantry 50 and alt_routes containing the trips with similar origin and destination pairs that do not pass through Gantry 50, indicating potential alternative routes.<br><br>   Make sure to replace "path/to/your/dataset.csv" and "path/to/gantry50_coordinates.csv" with the actual file paths of your dataset and Gantry 50 coordinates. |    # This assumes a distance threshold for 'similarity'; adjust as necessary<br>  distance_threshold <- 1 # meters<br><br>   # Function to calculate distance between two points<br>  calculate_distance <- function(origin, destination) {<br>    library(geosphere)<br><br>    coordinates(crs = 4326)[, c('lon', 'lat') := .setNames(lazyeval::interp_lazy(order.by='gantry'), names(travel_data))]][]<br><br>    distances <- distHaversine(travel_data$longitude[origin], travel_data$latitude[origin],<br><br>travel_data$longitude[destination], travel_data$latitude[destination])<br>    return(distances)<br>  }<br><br>  similar_pairs <- trips_through_gantry50 %>%<br>    # Get origin and destination coordinates for 'trips_through_gantry50'<br>    select(-origin_x, -origin_y, -destination_x, -destination_y) %>%<br><br>    group_by(id = row_number(), route) %>%<br># Assuming unique IDs per trip within the same route |



| Question | GPT-4 response | Phi-3 response |
|---|---|---|
| | | summarise(similarity = sum(calculate_distance(c(origin_x[1], origin_y[1]), c(destination_x[length(route) - 1], destination_y[length(route) - 1])) < distance_threshold)) %>% # Assuming that if the first and last trip in a route is similar to Gantry50, it's considered 'similar'<br><br>ungroup() %>%<br>    filter(similarity == n()) %>%    # Keep only those with maximum similarity (for example purposes)<br>    left_join(travel_data, by=c('id')) # Merge back the rest of travel data |
| Propose a formula in R to calculate the generalized travel cost for each trip, considering factors like travel time, distance, and toll costs | To calculate the generalized travel cost for each trip, we can use a linear function that combines the travel time, distance, and toll costs with appropriate weight factors. The formula should consider the value of time (VOT) for the travelers, the cost per distance unit (CPD), and the toll cost. Here's a proposed formula in R:<br><br>```R<br># Calculate Generalized Travel Cost<br>calculate_generalized_travel_cost <-<br>function(travel_time, distance, toll_cost, vot, cpd) {<br>  # Convert travel time from HH:MM:SS format to minutes<br>  time_components <- strsplit(travel_time, ":")[[1]]<br>  time_in_minutes <- as.numeric(time_components[1]) * 60 +<br>                        as.numeric(time_components[2]) +<br>                        as.numeric(time_components[3]) / 60<br><br>  # Calculate the generalized travel cost | To calculate the generalized travel cost for each trip while incorporating travel time, distance, and toll costs, you can create an R function that takes these variables into account. The formula might look something like this:<br><br>```R<br>calculateTravelCost <- function(trav_time, route_length, gantry50_cost) {<br>  base_rate = 1.5  # This is a hypothetical rate in cost per meter traveinasghts from your context or study<br><br>  travel_distance = route_length * cos(mean((st_distance(route)/2))) # assuming distance for one way trip<br><br>  toll_cost = gantry50_cost / route_length if (route_length > 0) else NA<br><br>  return (base_rate * travel_time + travel_distance * |



| Question | GPT-4 response | Phi-3 response |
|---|---|---|
| | generalized_travel_cost <- vot * time_in_minutes +<br>cpd * distance +<br>toll_cost<br>return(generalized_travel_cost)<br>}<br>```<br><br>This function takes the travel time in HH:MM:SS format, the distance in meters, the toll cost, the value of time (VOT) in cost per minute, and the cost per distance unit (CPD) as inputs. It converts the travel time to minutes, then calculates the generalized travel cost as a linear combination of the time, distance, and toll costs weighted by VOT and CPD.<br><br>You can adjust the VOT and CPD values according to your specific context. For example, you may use different values for different traveler groups or modes of transportation. | costPerMeter + toll_cost)<br>}<br>```<br><br>Here we assume that the 'st_distance' function is used from an appropriate package in R to calculate distance, and `cos()` takes average route length between nodes. Also, for this example I defined a hypothetical rate (`base_rate`) representing costs per minute of travel time plus cost per meter traveled, including the gantry50 costs when applicable.<br><br>Note: This is a simplified approach to calculate generalized travel cost and should be adapted according to your dataset specifics. |
| Provide the R script with explanation to calculate the generalized travel cost for each trip in the dataset | Sure, I can help you with that. To calculate the generalized travel cost for each trip, we can consider the following factors: travel time, travel distance, and the toll costs.<br><br>In R, we can use the following script to calculate the generalized travel cost:<br><br>```<br># Load necessary libraries<br>library(dplyr)<br>library(ggplot2) | To calculate the generalized travel costs, we first need to assign different weights to the relevant variables such as departure time (dep_time), route length (route_length), and mode of transportation (leg_mode). Then, a formula can be devised based on these inputs. Here's an example R script along with its explanation:<br><br>```<br>library(dplyr)<br><br># Load the data if not already loaded<br>data <- read.csv('your_dataset.csv') |



| Question | GPT-4 response | Phi-3 response |
|---|---|---|
| | ```
# Read the dataset
data <- read.csv("dataset.csv")

# Convert travel time from HH:MM:SS to minutes
data$trav_time_min <-
as.numeric(substr(data$trav_time, 1, 2)) * 60 +
                        as.numeric(substr(data$trav_time,
4, 5)) +
                        as.numeric(substr(data$trav_time,
7, 8)) / 60

# Define the cost parameters (in dollars per unit)
time_cost_per_min <- 0.10 # Cost per minute of
travel time
distance_cost_per_km <- 0.15 # Cost per
kilometer of travel distance
toll_cost_at_gantry50 <- 2 # Toll cost at
gantry50

# Calculate the generalized travel cost
data$generalized_cost <- data$trav_time_min *
time_cost_per_min +
                        (data$route_length / 1000) *
distance_cost_per_km

# Check if the trip passes through gantry50
# (assuming you have already filtered the trips
that pass through gantry50)
data$is_pass_gantry50 <- TRUE # Replace this
with the actual boolean values
``` | ```
# Assigning base costs and weights for
computation of travel cost
weights <- c(departure_time=0.25,
route_length=-1/100) # -ve weight to decrease value
with increase in route length (as it should be inverse
proportion relationship)
costs_per_minute = 0.10 # base cost per minute
for each traveler depending on the mode of transport
base_travel_time_car = 12 * 60  # assuming
average car trip time in minutes is around an hour
(60*12)

calculate_cost <- function(dep_time,
route_length, leg_mode){
    cost <- 0

    if(leg_mode == "car"){
        dep_time_in_minutes <-
as.numeric(format(depeptime,"%H:%M")) * 60 #
convert time in minutes
        route_cost <-
weights$departure_time*dep_time + (route_length /100)
* costs_per_mile

        cost <- base_travel_time_car +
dep_time_in_minutes * costs_per_minute + route_cost
    }

    return(cost)
}
``` |

| Question | GPT-4 response | Phi-3 response |
|---|---|---|
| | # Add the toll cost for trips passing through gantry50<br>data$generalized_cost <- ifelse(data$is_pass_gantry50,<br>data$generalized_cost + toll_cost_at_gantry50,<br>data$generalized_cost)<br><br># Save the dataset with the calculated generalized travel cost<br>write.csv(data, "dataset_with_generalized_cost.csv")<br><br>This script reads the dataset, converts the travel time to minutes, and calculates the generalized travel cost considering travel time, travel distance, and the toll cost at gantry50. It saves the resulting dataset with the new generalized_cost column. Make sure to replace the sample values with the actual values from your dataset. | *# Filter relevant trips for gantry50 analysis (Assuming that 'origin_subzone', and 'destination_subzone' information indicates the proximity to Gantry 50 location)*<br>data <- data %>%<br>  filter((origin_subzone %in% c("location1", "location2")) &<br>    (destination_subzone %in% c("location1", "location2")))<br><br>*# Calculate generalized travel costs for each trip in the dataset*<br>data$generalised_travel_costs <- sapply(data$dep_time, calculate_cost, route_length = data$route_length, leg_mode = data$leg_mode)<br><br>The calculate_cost() function takes into consideration the departure time (in minutes), route length and mode of transportation to compute a cost for each trip. The costs per minute is set to $0.10 as a baseline value but could be modified depending on actual scenario requirements or based on observed values.<br><br>The filter condition in the line just above 'calculate_cost' function, selects trips that would pass through gantry50 area i.e., origin and destination subzones are within its proximity. Please modify "location1", "location2" with appropriate strings according to actual locations involved herein. |

| Question | GPT-4 response | Phi-3 response |
|---|---|---|
| | | Finally, the sapply() command applies our cost computation function to each trip's departure time, storing all results into a new column 'generalised_travel_costs' in the dataset.<br><br>Remember this script is highly dependent on your data and may need adjustment according to specific variables, weightings, base costs, etc. that are relevant in real world scenario of traffic congestion pricing analysis at Gantry 50 or any other location for that matter. |
| Suggest a method to determine the cost threshold above which motorists are likely to avoid using Gantry 50 based on changes in their travel behavior in the dataset. | To determine the cost threshold above which motorists are likely to avoid using Gantry 50, you can perform an elasticity analysis. Here's a suggested method:<br><br>1. Preprocess the dataset: Filter out the trips in the dataset that go through or would have gone through Gantry 50 based on the spatial geometry information. For each of these trips, calculate the route cost (including the current toll price of Gantry 50) and the generalized travel cost (including travel time, distance, and other expenses).<br><br>2. Establish a base scenario: Calculate the current distribution of trips across the different routes passing through Gantry 50, considering the existing toll prices.<br><br>3. Scenario analysis: Gradually increase the toll price at Gantry 50 and rerun the multi-agent transport simulation for each new scenario, while keeping all other | |

| Question | GPT-4 response | Phi-3 response |
|---|---|---|
| | parameters constant. Record the changes in route choices and the distribution of trips across different routes.<br><br>4. Calculate elasticity: For each scenario, compute the percentage change in the number of trips that pass through Gantry 50 relative to the base scenario, and divide it by the percentage change in the toll price. This will give you the price elasticity of demand for using Gantry 50.<br><br>5. Identify the threshold: Analyze the elasticity values and look for a point where the elasticity significantly increases, indicating that a large proportion of motorists are likely to avoid Gantry 50 due to the increased toll price. This point will represent the cost threshold above which motorists are likely to avoid using Gantry 50.<br><br>This analysis will provide valuable insights into the motorists' sensitivity to toll price changes and allow you to determine the optimal toll price to manage congestion at Gantry 50. | |
| Provide the R script to run sensitivity analyses of how traffic will be reassigned for different percentage increases in | Certainly, here's an R script that performs sensitivity analyses on the impact of congestion pricing on route choice for different percentage increases in toll prices. This script assumes you have already filtered the dataset to include only relevant trips that go through or would have gone through gantry50, and you have | *# Load necessary libraries*<br>library(dplyr)<br>library(ggplot2)<br><br>*# Assuming data is loaded into 'traffic_data'*<br>*dataframe and has relevant columns: origin_x,*<br>*destination_x, ERP_rate* |

| Question | GPT-4 response | Phi-3 response |
|---|---|---|
| congestion pricing. Discuss the limitations of this approach | derived the generalized travel costs for the major relevant routes.<br><br>   # Load required packages<br>   library(tidyverse)<br><br>   # Load the dataset (assumed to be preprocessed)<br>   # Replace "dataset.csv" with the actual dataset file name<br>   data <- read_csv("dataset.csv")<br><br>   # Define the function to calculate new generalized travel costs<br>   # based on the percentage increase in toll prices<br>   calculate_new_costs <- function(data, percentage_increase) {<br>     toll_increase <- (1 + (percentage_increase / 100))<br>     data$new_costs <- data$travel_costs * toll_increase<br>     return(data)<br>     }<br><br>   # Define the function to reassign traffic based on new costs<br>   reassign_traffic <- function(data) {<br>     data$cheapest_route <- data %>%<br>   group_by(agent_id) %>%<br>      filter(new_costs == min(new_costs)) %>%<br>     pull(route)<br>     return(data)<br>     } | *# Function to simulate traffic with adjusted ERP rates based on a percentage increase input*<br>   simulate_traffic_reassignment <- function(df, erp_increase) {<br><br>   *# Adjusting the congestion pricing (ERP rate) by given increase for each row of data*<br>   df$adjusted_ERP_rate <- df$ERP_rate * (1 + erp_increase/100)<br><br>   *# Predict travel choices based on adjusted ERP rates using a simple linear relationship: as the price increases, fewer trips are made*<br>   df$predicted_travel_volume <- with(df, nrow(filter(self == origin_x & destination_x %in%, group = destination_x))) * (1 - erp_increase/100) / with(df, mean(trav_time))<br><br>   return(df)<br>   }<br><br>   *# Input percentage increases to simulate various scenarios for reassignment*<br>   percentage_increases <- c(0, 5, 10, 20) *# e.g., no change, 5%, 10% increase in congestion pricing*<br>   simulation_results <- data.frame()<br><br>   *# Iterate over percentage increases and run simulation for each one*<br>   for (increase in percentage_increases) {<br>     simulated_df <- simulate_traffic_reassignment(traffic_data, increase) |

| Question | GPT-4 response | Phi-3 response |
|---|---|---|
| | ```
    # Define the sensitivity analysis function
    sensitivity_analysis <- function(data,
percentage_increases) {
        results <- list()

        for (i in seq_along(percentage_increases)) {
            percentage_increase <-
percentage_increases[i]
            data_with_new_costs <-
calculate_new_costs(data, percentage_increase)
            reassigned_data <-
reassign_traffic(data_with_new_costs)
            results[[paste0("increase_",
percentage_increase, "_percent")]] <- reassigned_data
        }

        return(results)
    }

    # Run sensitivity analysis for different
percentage increases in congestion pricing
    percentage_increases <- c(10, 20, 30, 40, 50)
    results <- sensitivity_analysis(data,
percentage_increases)

    # Print the results for each scenario
    for (scenario in names(results)) {
      cat(scenario, ":\n")
      print(results[[scenario]])
    }
``` | ```
    simulation_results <- rbind(simulation_results,
simulated_df)
    }
```

*# Visualize the changes in traffic volume for each ERP rate change*
```
    ggplot(simulation_results, aes(x = origin_x, y =
destination_x)) +
    geom_point(aes(color =
factor(predicted_travel_volume))) +
    theme_minimal() + labs(title = "Traffic
Redistribution Under Different Congestion Pricing
Increases")
```

**Limitations of this approach:**

1. Simplifying Traffic Dynamics: The script simplifies the complex decision-making process that individual agents use to change their routes and modes, which is a major assumption in linear travel behavior models.
2. Disregard for Elasticity Variability: Different trip types (work commute vs leisure trips) have different elasticities of response to price changes. This model does not distinguish between them.
3. Lack of Dynamic Response Over Time: The script doesn't account for how behavior might change over time in response to congestion pricing increases, including adaptation and long-term shifts towards alternative routes or transportation modes. |



| Question | GPT-4 response | Phi-3 response |
|---|---|---|
| | Limitations of this approach:<br><br>1. The analysis assumes that travelers will always choose the route with the lowest generalized travel costs. In reality, people may have other preferences, such as route familiarity or avoiding specific areas, which are not considered in this model.<br>2. The percentage increases in congestion pricing are assumed to be uniform across all routes. Actual toll adjustments may vary between different routes and time periods.<br>3. The analysis is based on a static dataset that does not take into account fluctuations in traffic patterns over time or potential changes in transportation infrastructure.<br>4. The model assumes that the number of travelers remains constant. In reality, increases in congestion pricing may discourage some people from traveling, leading to a decrease in overall demand.<br>5. The model does not consider possible changes in the mode of transportation (e.g., public transport, biking, or walking) as a result of increased congestion pricing. | 4. Uniform Price Increase Across the Board: The percentage increase is uniform across all travelers without considering socioeconomic factors that influence individuals' sensitivity to price changes.<br>5. Overlooking Alternate Routes: Real traffic redistribution includes shifts between multiple potential routes and not only towards alternative roads, but also the use of different transportation modes such as carpooling or public transport which aren't accounted for in this script. |